\documentclass{article}

\usepackage[preprint]{neurips_2026}

\usepackage[utf8]{inputenc}
\usepackage[T1]{fontenc}
\usepackage{booktabs}
\usepackage{amsfonts}
\usepackage{amsmath}
\usepackage{amssymb}
\usepackage{latexsym}
\usepackage{nicefrac}
\usepackage{microtype}
\usepackage[table]{xcolor}
\usepackage{graphicx}
\usepackage{float}
\usepackage{inconsolata}
\usepackage{hyperref}
\hypersetup{hypertexnames=false}

\graphicspath{{./}{./images/}}

\title{Disentangling Optimization Scale from Preference Scale in DPO}

\author{Ivan Kruzhilov}

\begin{document}

\maketitle

\begin{abstract}
Direct Preference Optimization (DPO) is a widely used objective for aligning language models from preference data, with the coefficient $\beta$ commonly interpreted as controlling the KL constraint to a reference policy. We show that $\beta$ entangles two distinct roles: it governs the effective inverse preference-noise scale and simultaneously rescales the optimization dynamics, coupling this scale with the effective step size. As a consequence, at a fixed learning rate the achieved policy deviation is non-monotone in $\beta$: it vanishes in a dead zone at small $\beta$, reaches a peak at an intermediate value, and decreases again for larger $\beta$. Moreover, standard DPO loss values are not comparable across $\beta$: runs with nearly identical loss curves can differ several-fold in KL divergence from the reference model. This entanglement obscures the role of $\beta$, increases sensitivity to hyperparameter choices, and complicates learning-rate scheduling. We propose a centered-softplus reformulation that is argmin-equivalent to DPO for $\beta>0$, while making the inverse preference-noise-scale and learning-rate effects explicit and independently tunable. The normalized centered-softplus objective also admits a continuous $\beta\to0$ endpoint that reduces to a linear preference-margin objective.
\end{abstract}

\section{Introduction}

Direct Preference Optimization (DPO) is a post-training alignment method for large language models that directly optimizes pairwise human preferences without explicit reward-model training, and it is widely used in instruction tuning and RLHF-style pipelines \citep{Mitchell2023DPO}.
The closest lines of work to our question study the sensitivity and dynamics of DPO itself: $\beta$-DPO dynamically adapts $\beta$ to data quality and instance difficulty \citep{Wu2024BetaDPO}, Balanced-DPO reweights winner/loser gradients to mitigate gradient imbalance \citep{GradientImbalance2025}, and length-focused analyses separate preference quality from verbosity effects in DPO-trained policies \citep{Eldan2024LengthDPO}.

Across this literature, $\beta$ is predominantly treated as a control parameter for the preference--regularization trade-off.
Under this common KL-control interpretation, smaller $\beta$ values are often expected to permit larger policy deviation from the reference model \citep{Mitchell2023DPO,Wu2024BetaDPO}.
DPO performance is also empirically known to be highly sensitive to the choice of $\beta$ \citep{Wu2024BetaDPO}.
Prior work has not isolated the direct optimizer-scale role of $\beta$: in standard DPO-style objectives, $\beta$ not only changes the effective inverse preference-noise scale or regularization behavior, but also multiplies the marginal gradient and therefore changes the effective update scale.

Decreasing $\beta$ reduces gradient magnitude proportionally. Under SGD this is equivalent to lowering the learning rate and complicates hyperparameter tuning. We first demonstrate the effect in standard DPO training: with very small $\beta$, gradients stay weak, KL changes little, and the policy remains close to the reference model rather than moving farther toward the preferred responses.
The same entanglement also undermines the loss as a monitoring signal: the standard DPO loss depends on $\beta$ and the preference margin only through their product, so its value can remain nearly flat during training, or coincide across runs whose policies differ several-fold in KL.

Our main contributions are as follows:
\begin{itemize}
    \item We show that the DPO parameter $\beta$ directly scales the gradient magnitude, so changing $\beta$ affects not only the preference--KL trade-off but also the effective learning speed. As a consequence, $\beta$ acts non-monotonically on training: with a fixed learning rate, the achieved policy deviation is suppressed in a dead zone at small $\beta$, peaks near $\beta_{\mathrm{peak}}\approx 4.6\,\tau$, where $\tau$ is the effective stopping tolerance of the optimization (Section~\ref{sec:loss-geometry}), and decreases again for larger $\beta$.
    \item We propose an alternative centered-softplus loss (Section~\ref{sec:normalized}) that has the same optimum as the original DPO objective for $\beta>0$ while removing the direct dependence of update magnitude on $\beta$. The proposed loss also has a well-defined $\beta=0$ limiting case, where optimization reduces to a linear preference-margin objective; we discuss this endpoint as a useful but less regularized regime rather than as a default replacement for finite-$\beta$ DPO.
    \item We show that standard DPO loss values are not comparable across different $\beta$ (Section~\ref{sec:loss-incomparability}): as an illustration, we exhibit runs whose training or validation loss curves nearly coincide while their KL divergence from the reference model, validation NLL, and downstream behavior differ substantially (Appendices~\ref{app:saturation},~\ref{app:loss-invariance}, and~\ref{app:normalized-trajectories}). The proposed normalized loss instead tracks the physical preference margin and separates such runs.
\end{itemize}

We emphasize that we do not claim a different optimum: the entanglement of inverse preference-noise scale and update scale can be partly compensated by retuning the learning rate, and is often partially masked by adaptive optimizers, but such compensation is external to the objective parameterization itself. Our goal is to expose and decouple these effects in the parameterization.
Finally, the gradient-vanishing issues identified above, and the remedies we propose, are not specific to classical DPO: the same coupling arises for the DPO-family modifications surveyed in Section~\ref{sec:background}.

\section{Background: KL-Regularized RLHF and DPO}
\label{sec:background}

Consider a reference policy $\pi_{\mathrm{ref}}$ and a reward function $r(x,y)$.
The standard KL-regularized RLHF objective is
\begin{equation}
\label{eq:rlhf}
\max_{\pi_{\theta}}\;
\mathbb{E}_{x\sim\mathcal{D},\,y\sim\pi_{\theta}(\cdot\mid x)}
\bigl[r(x,y)\bigr]
- \lambda\,\mathbb{E}_{x\sim\mathcal{D}}\!
\mathrm{KL}\!\bigl(\pi_{\theta}(\cdot\mid x)\,\|\,\pi_{\mathrm{ref}}(\cdot\mid x)\bigr),
\end{equation}
where $\lambda>0$ is the KL-regularization coefficient controlling the trade-off between reward maximization and staying close to the reference model.
A common pipeline first fits a reward model on human preference data under the Bradley--Terry model \citep{BradleyTerry1952} and then optimizes Equation~\eqref{eq:rlhf} with PPO or a related policy-gradient method.
\citet{Mitchell2023DPO} show that this two-stage procedure can be collapsed into a single objective over policy log-probabilities. The optimal policy of Equation~\eqref{eq:rlhf} induces an implicit reward $r(x,y)=\lambda\log\frac{\pi(y\mid x)}{\pi_{\mathrm{ref}}(y\mid x)}+\mathrm{const}$. If the Bradley--Terry likelihood has pairwise annotator-noise scale $s_a$, substitution cancels the normalizing constant and yields the effective DPO coefficient $\beta=\lambda/s_a$ in Equation~\eqref{eq:dpo-loss}; the conventional unit-noise formulation sets $s_a=1$. Thus no separate reward model needs to be trained.
The closed-form optimal policy and the full derivation are recalled in Appendix~\ref{app:bt-temperature}.

The original DPO formulation connects preference optimization to KL-constrained RLHF and establishes the basic log-ratio objective \citep{Mitchell2023DPO}.
Subsequent work has refined this framework in several directions.
Objective-level variants such as IPO \citep{Azar2024IPO}, CPO \citep{Xu2024CPO}, KTO \citep{Ethayarajh2024KTO}, and EXPO \citep{EXPO2025} reformulate preference learning with alternative optimization principles, while related extensions handle noisy labels, soft preferences, adaptive margins, entropy control, and richer comparison structures \citep{Chowdhury2024rDPO,Furuta2024GAPO,Wu2025AlphaDPO,Omura2024HDPO,ADPO2025,RatingDPO2026}.

Following the standard DPO setup, let $(x,y_w,y_l)$ denote a prompt, a preferred response, and a rejected response.
Define the policy log-probability gap
\begin{equation}
\Delta_{\theta} = \log \pi_{\theta}(y_w \mid x) - \log \pi_{\theta}(y_l \mid x),
\end{equation}
and analogously for the reference model,
\begin{equation}
\Delta_{\mathrm{ref}} = \log \pi_{\mathrm{ref}}(y_w \mid x) - \log \pi_{\mathrm{ref}}(y_l \mid x).
\end{equation}
We use the margin variable
\begin{equation}
\Delta = \Delta_{\theta} - \Delta_{\mathrm{ref}}.
\end{equation}
The classical DPO objective is
\begin{equation}
\label{eq:dpo-loss}
\mathcal{L}_{\mathrm{DPO}}(\Delta; \beta) = -\log \sigma(\beta \Delta),
\end{equation}
equivalently $\mathcal{L}_{\mathrm{DPO}}(\Delta; \beta) = \log(1 + \exp(-\beta \Delta))$, where $\sigma(\cdot)$ denotes the sigmoid function.

Most DPO variants share an explicit multiplicative $\beta$ factor in the marginal gradient, so reducing $\beta$ similarly suppresses update magnitude; we analyze this coupling in detail in Section~\ref{sec:cause}.
Table~\ref{tab:dpo-losses} summarizes pairwise objectives in the margin $\Delta$; Appendix~\ref{app:dpo-variant-gradients} discusses how the same coupling appears in unpaired KTO, reference-free SimPO, and token-level TDPO, and how IPO, ROPO, and AOT depart from the standard DPO pattern.

\begin{table}[t]
\caption{Comparison of objectives and marginal gradients across DPO-family methods.}
\label{tab:dpo-losses}
\centering
\footnotesize
\setlength{\tabcolsep}{1pt}
\begin{tabular}{p{0.13\linewidth} p{0.17\linewidth} p{0.33\linewidth} p{0.31\linewidth}}
\toprule
\textbf{Method} & \textbf{\shortstack{Hyper-param /\\ Labels}} & \textbf{Loss} & \textbf{\shortstack{Marginal gradient\\$\partial L/\partial\Delta$}} \\
\midrule
DPO \citep{Mitchell2023DPO} & $\beta$ & $-\log \sigma(\beta \Delta)$ & $-\beta\,\sigma(-\beta \Delta)$ \\
IPO \citep{Azar2024IPO} & $\beta$ & $\left(\Delta-\frac{1}{2\beta}\right)^2$ & $2\left(\Delta-\frac{1}{2\beta}\right)$ \\
cDPO \citep{Mitchell2023cDPO} & $\beta,\varepsilon$ & $-(1-\varepsilon)\log \sigma(\beta \Delta)-\varepsilon\log \sigma(-\beta \Delta)$ & $\beta\bigl(\sigma(\beta \Delta)-(1-\varepsilon)\bigr)$ \\
$\alpha$-DPO \citep{Wu2025AlphaDPO} & $\beta,\alpha,\gamma$ & $-\log \sigma(u-c_{\alpha})$ & $\partial L/\partial m_{\theta}=-\beta\,\sigma(c_{\alpha}-u)$ \\
GAPO \citep{Furuta2024GAPO} & $\beta,\hat{p}\in(0,1)$ & $-\log \sigma\!\left(\beta(2\hat{p}-1)\Delta\right)$ & $-\beta(2\hat{p}-1)\sigma\!\left(-\beta(2\hat{p}-1)\Delta\right)$ \\
$\beta$-DPO \citep{Wu2024BetaDPO} & $\beta_{\mathrm{dyn}}(\cdot)$ & $-\log \sigma(\beta_{\mathrm{dyn}}\Delta)$ & $-\beta_{\mathrm{dyn}}\,\sigma(-\beta_{\mathrm{dyn}}\Delta)$ \\
SimPO \citep{Meng2024SimPO} & $\beta,\gamma$ & $-\log \sigma(\beta m-\gamma)$ & $\partial L/\partial m=-\beta\,\sigma(\gamma-\beta m)$ \\
TDPO \citep{Zeng2024TDPO} & $\beta,\alpha$ & $-\log \sigma(\beta M)$, $M=\Delta-\delta_{\mathrm{SeqKL}}$ & $\partial L/\partial M=-\beta\,\sigma(-\beta M)$ \\
ROPO \citep{Liang2025ROPO} & $\beta,\alpha$ & robust loss inducing gradient weight $w_{\alpha}(\Delta)$ & $\propto -\beta\,w_{\alpha}(\Delta)$ \\
AOT \citep{Melnyk2024AOT} & $\beta,\{\Delta_i\}_{i=1}^{B}$ & distributional loss $D_{\mathrm{dist}}\!\left(\{\beta \Delta_i\}_{i=1}^{B}\right)$ & $\partial L/\partial\Delta_i=\beta\,w_i^{\mathrm{dist}}$ \\
\begin{minipage}[t]{\linewidth}\raggedright
KTO\\
\citeauthor{Ethayarajh2024KTO}\\
(\citeyear{Ethayarajh2024KTO})%
\end{minipage} &
\begin{minipage}[t]{\linewidth}\raggedright
$\beta,\lambda_D,\lambda_U,$\\
$z_0,b$%
\end{minipage} &
\begin{tabular}[t]{@{}l@{}}
$b{=}1$: $-\lambda_D\log\sigma(\beta(v{-}z_0))$\\
$b{=}0$: $-\lambda_U\log\sigma(-\beta(v{-}z_0))$
\end{tabular} &
\begin{tabular}[t]{@{}l@{}}
$\partial L/\partial v=-\lambda_D\beta\bigl(1{-}\sigma(\beta(v{-}z_0))\bigr)$\\
$\partial L/\partial v=+\lambda_U\beta\bigl(1{-}\sigma(-\beta(v{-}z_0))\bigr)$
\end{tabular} \\
\bottomrule
\end{tabular}
\end{table}

\section{Gradient-Scale Coupling and Loss Incomparability in Standard DPO}
\label{sec:pathologies}

We isolate two related scale effects in the standard DPO parameterization.
First, $\beta$ multiplies the marginal gradient, so it rescales the effective update and makes the achieved policy deviation non-monotone.
Second, the pairwise loss depends on $\beta$ and the preference margin only through their product, so loss values are not comparable across $\beta$.

\subsection{When Smaller \texorpdfstring{$\beta$}{beta} Weakens DPO Updates}
\label{sec:beta-weakens-updates}

\begin{figure}[t]
  \centering
  \includegraphics[width=0.92\linewidth]{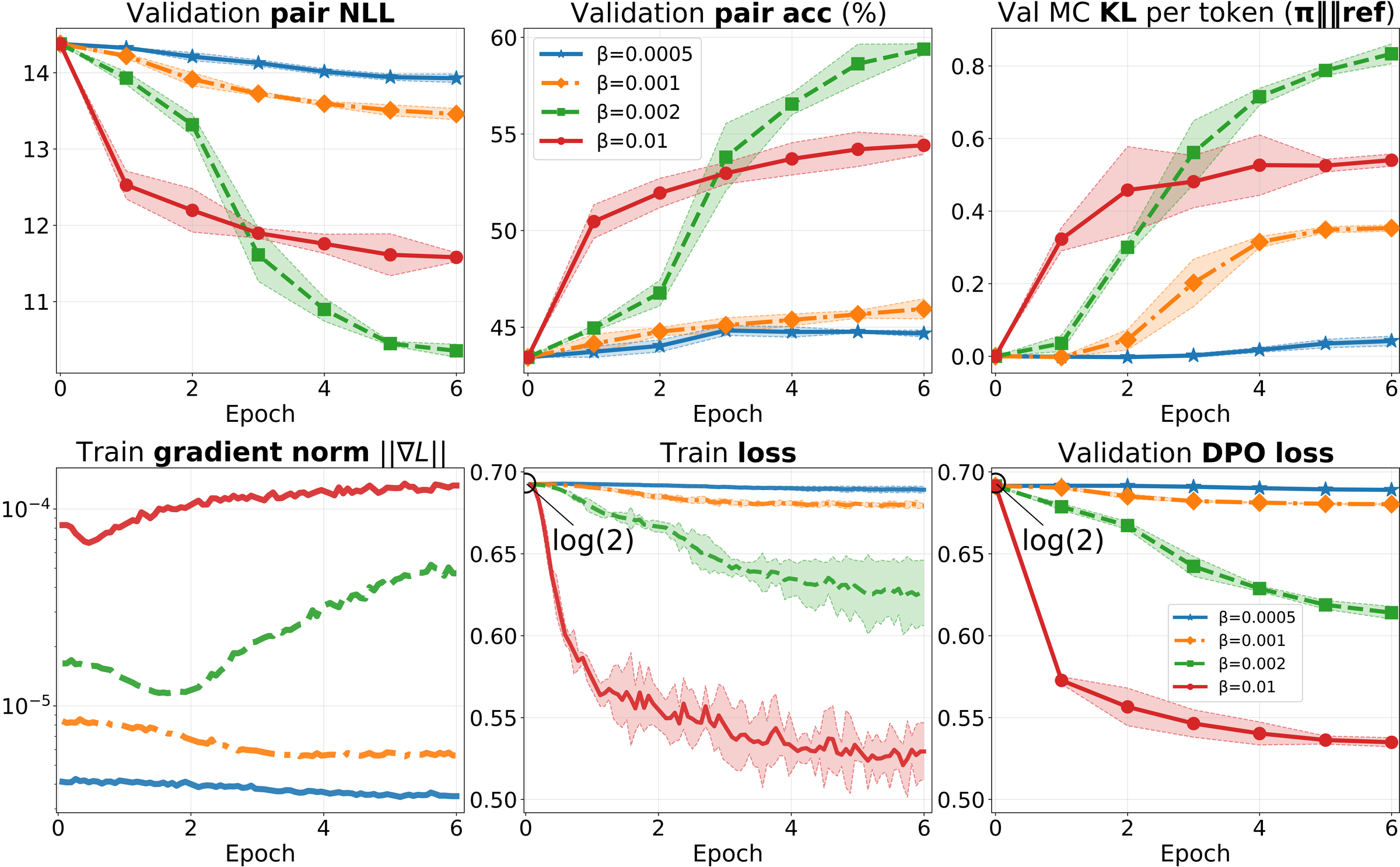}
  \caption{Training and validation dynamics of our DPO implementation on HelpSteer3 with Qwen3-4B-Instruct-2507 (SGD, $\mathrm{lr}=2\times10^{-4}$, batch size 24). Curves show the mean over five seeds; shaded bands denote one standard deviation. Final KL is non-monotone in $\beta$; see Section~\ref{sec:loss-geometry} and Appendix~\ref{app:saturation}.}
  \label{fig:cause-drift}
\end{figure}

Figure~\ref{fig:cause-drift} is the empirical starting point of our analysis.
It shows a standard-DPO $\beta$ sweep on HelpSteer3 with Qwen3-4B-Instruct-2507 (SGD, learning rate $2\times10^{-4}$, batch size~24), comparing $\beta\in\{5\times10^{-4}, 10^{-3}, 2\times10^{-3}, 10^{-2}\}$.
Two observations follow from this sweep.
First, when $\beta$ is made very small, optimization degrades rather than producing a larger policy deviation: the observed gradients are weak, the validation KL remains almost unchanged, and the trained policy stays close to the reference model.
This conflicts with the usual role assigned to $\beta$ in DPO, where it is interpreted as a control parameter for the strength of the preference update relative to the reference-policy constraint.
Second, the sweep violates a monotone KL-control reading of $\beta$: final validation KL peaks at the intermediate value $\beta=2\times10^{-3}$ (0.83 per token) while $\beta=10^{-2}$ reaches only 0.54, and after epoch~3 the $\beta=0.002$ run overtakes $\beta=0.01$ in both gradient norm and KL.
The sweep therefore breaks a simple $\beta\leftrightarrow\mathrm{KL}$ reading: $\beta\cdot\mathrm{KL}$ grows by more than two orders of magnitude (Appendix~\ref{app:saturation}).
Section~\ref{sec:cause} identifies the cause---an explicit multiplicative $\beta$ factor in the marginal gradient---Section~\ref{sec:loss-geometry} turns it into a stopping-condition argument that explains both observations, and the normalized objective in Section~\ref{sec:normalized} restores a much more predictable scaling (for $\beta>0$).
Table~\ref{tab:init-grad-linearity} already makes the linear suppression visible at the very first optimization step of the same sweep: the measured initial gradient magnitudes $G_0$ scale linearly with $\beta$, with $G_0/\beta$ constant to within roughly $2\%$ across a $20\times$ range of $\beta$.
At initialization $\Delta=0$, so the marginal gradient is exactly $-\beta/2$ (Section~\ref{sec:cause}) and $G_0/\beta$ is predicted to be constant across $\beta$.
Appendix~\ref{app:grad-decomposition} relates $G_0$ to the Euclidean loss gradient and explains why a minibatch log identifies an effective batch-averaged Jacobian scale rather than a typical per-pair norm.

\begin{table}[h]
\centering
\small
\caption{Initial gradient magnitudes $G_0$ (mean absolute per-parameter gradient at the first logged step of epoch~1, averaged over five seeds) from the HelpSteer3 SGD runs in Figure~\ref{fig:cause-drift}.}
\label{tab:init-grad-linearity}
\begin{tabular}{lcccc}
\toprule
$\beta$ & $5\times10^{-4}$ & $10^{-3}$ & $2\times10^{-3}$ & $10^{-2}$ \\
\midrule
$G_0$ & $4.12\times10^{-6}$ & $8.37\times10^{-6}$ & $1.64\times10^{-5}$ & $8.28\times10^{-5}$ \\
$G_0/\beta$ & $8.23\times10^{-3}$ & $8.37\times10^{-3}$ & $8.22\times10^{-3}$ & $8.28\times10^{-3}$ \\
\bottomrule
\end{tabular}
\end{table}

Neither effect is specific to this dataset or optimizer.
Adaptive optimizers such as AdamW can partly obscure the $\beta$-dependent scaling of gradient magnitude (Appendix~\ref{app:adam-threshold} estimates $\beta_{\mathrm{crit}}$ such that for $\beta\lesssim\beta_{\mathrm{crit}}$ the linear-in-$\beta$ update suppression is again visible even under AdamW).
Moment estimates effectively stabilize step sizes and smooth training dynamics, so the coupling between $\beta$ and update strength is most apparent when $\beta$ is already small, exactly where the explicit $\beta$ prefactor in $\partial\mathcal{L}/\partial\Delta$ suppresses gradients most strongly.
Appendix~\ref{app:hh-rlhf-dynamics} (Figure~\ref{fig:intro-overview}) documents the same degradation on UltraFeedback with TRL \texttt{DPOTrainer} under AdamW at very small $\beta$; the same pattern appears on PKU-processed HH-RLHF with Mamba-2 (Figure~\ref{fig:hh-rlhf-mamba2-trl}).
Compared with those AdamW runs, plain SGD makes the effect easier to see, and it already appears at larger $\beta$.
An alternative HelpSteer3 SGD configuration at learning rate $5\times10^{-4}$ (Figure~\ref{fig:hsteer-sgd-lr5e4}) and HH-RLHF trajectories are reported in Appendix~\ref{app:hh-rlhf-dynamics}.

\subsection{Mechanism of Gradient-Scale Coupling}
\label{sec:cause}

Using the margin $\Delta$ and classical DPO loss from Section~\ref{sec:background} (Equation~\eqref{eq:dpo-loss}), the gradient with respect to $\Delta$ is
\begin{equation}
\label{eq:dpo-marginal-grad}
\frac{\partial \mathcal{L}_{\mathrm{DPO}}}{\partial \Delta}
= -\beta\frac{\exp(-\beta \Delta)}{1+\exp(-\beta \Delta)}
= -\beta\,\sigma(-\beta \Delta).
\end{equation}
Figures~\ref{fig:intro-overview} and~\ref{fig:cause-drift} mirror the linear $\beta$ prefactor in Equation~(\ref{eq:dpo-marginal-grad}): on HelpSteer3 (Figure~\ref{fig:cause-drift}), gradient norms rank-order with $\beta$, with the larger-$\beta$ curve lying above the smaller-$\beta$ curve, while the UltraFeedback TRL run in Appendix~\ref{app:hh-rlhf-dynamics} (Figure~\ref{fig:intro-overview}) uses a very small $\beta$ under AdamW and maintains comparatively small gradient norms throughout training.
This linear scaling is exact at initialization: the policy coincides with the reference model, so $\Delta=0$, $\partial\mathcal{L}_{\mathrm{DPO}}/\partial\Delta=-\beta/2$, and the full parameter gradient $-(\beta/2)\,\nabla_{\theta}\Delta_{\theta}$ has norm proportional to $\beta$ for a fixed model state, i.e., $G_0/\beta$ should be constant across $\beta$.
Table~\ref{tab:init-grad-linearity} (Section~\ref{sec:beta-weakens-updates}) quantifies this prediction on the HelpSteer3 SGD runs of Figure~\ref{fig:cause-drift}: the measured ratios $G_0/\beta$ agree within roughly $2\%$ across a $20\times$ range of $\beta$.
Measurement and analysis of the logged gradient magnitudes are discussed in Appendix~\ref{app:grad-decomposition}.
Thus, the classical form contains an explicit multiplicative $\beta$ factor in the gradient magnitude.
As $\beta$ decreases, gradient magnitude is linearly suppressed by this front multiplier.
This coupling motivates a normalized reformulation: if $\beta$ is intended to control the preference--regularization trade-off, it should not also directly rescale the effective update size.
As summarized in Table~\ref{tab:dpo-losses}, most DPO-family objectives share this explicit $\beta$ prefactor in the marginal gradient.

The same linear $\beta$-dependent gradient-scale coupling is not specific to logistic Bradley--Terry \citep{BradleyTerry1952}: it holds for any pairwise loss $-\log F(\beta\Delta)$ with a full-line noise CDF $F$ (Appendix~\ref{app:thurstone-grad}, Eq.~\eqref{eq:general-link-loss}), including the Thurstone case $F=\Phi$ \citep{Sun2025RethinkingBT}; the corresponding marginal gradient always contains an explicit multiplicative $\beta$ prefactor (Eq.~\eqref{eq:general-link-grad}).

\subsection{Dead Zone, Peak, and Saturation}
\label{sec:loss-geometry}

The explicit $\beta$ prefactor in Equation~\eqref{eq:dpo-marginal-grad} does more than rescale updates: it enters the condition under which training effectively stops, and thereby makes the achieved policy deviation non-monotone in $\beta$, as observed in Figure~\ref{fig:cause-drift}.
We model effective convergence with a stopping tolerance $\tau>0$: updates on a preference pair become negligible once the magnitude of the marginal gradient drops below $\tau$, which lumps together the SGD noise floor, learning-rate decay, and the finite training horizon.
By Equation~\eqref{eq:dpo-marginal-grad}, $|\partial\mathcal{L}_{\mathrm{DPO}}/\partial\Delta|=\beta\,\sigma(-\beta\Delta)$, and the stopping condition $\beta\,\sigma(-\beta\Delta)=\tau$ yields
\begin{equation}
\label{eq:sat-margin-dpo}
\Delta^{*}_{\mathrm{DPO}}=\frac{1}{\beta}\log\!\left(\frac{\beta}{\tau}-1\right),
\qquad \beta>2\tau.
\end{equation}
Three regimes follow (Appendix~\ref{app:saturation} gives the full derivation and the normalized-objective counterpart).
\begin{itemize}
\item \emph{Dead zone} ($\beta\le 2\tau$): the marginal gradient is largest at initialization, where $\Delta=0$ and $|\partial\mathcal{L}_{\mathrm{DPO}}/\partial\Delta|=\beta/2$. If $\beta/2\le\tau$, the gradient is below tolerance from the start and training never escapes initialization.
\item \emph{Non-monotone peak} ($\beta>2\tau$): maximizing Equation~\eqref{eq:sat-margin-dpo} over $\beta$ gives
\begin{equation}
\label{eq:sat-peak}
\beta_{\mathrm{peak}}\approx 4.6\,\tau,
\end{equation}
after which $\Delta^{*}_{\mathrm{DPO}}$ decays as $\log(\beta/\tau)/\beta$. Unlike the normalized objective, the product $\beta\,\Delta^{*}_{\mathrm{DPO}}=\log(\beta/\tau-1)$ grows only logarithmically, so $\beta$ is a poor knob for the achieved margin.
\item \emph{Right-hand saturation}: the same $\log(\beta/\tau)/\beta$ decay drives the policy back toward the reference. Appendix~\ref{app:right-dead-zone} uses the scalar closure $\beta_{\mathrm{r}}\approx1/(g_B^{2}\tau H)$ to obtain only an order-of-magnitude finite-horizon scale, where $g_B$ is the initial batch-averaged Jacobian proxy and $H$ is the learning-rate budget. The estimate is approximate because $g_B$ changes during training and differs from a typical per-pair Jacobian; exact minibatch dynamics require cross-pair Gram products.
\end{itemize}

\begin{figure}[t]
  \centering
  \includegraphics[width=\linewidth]{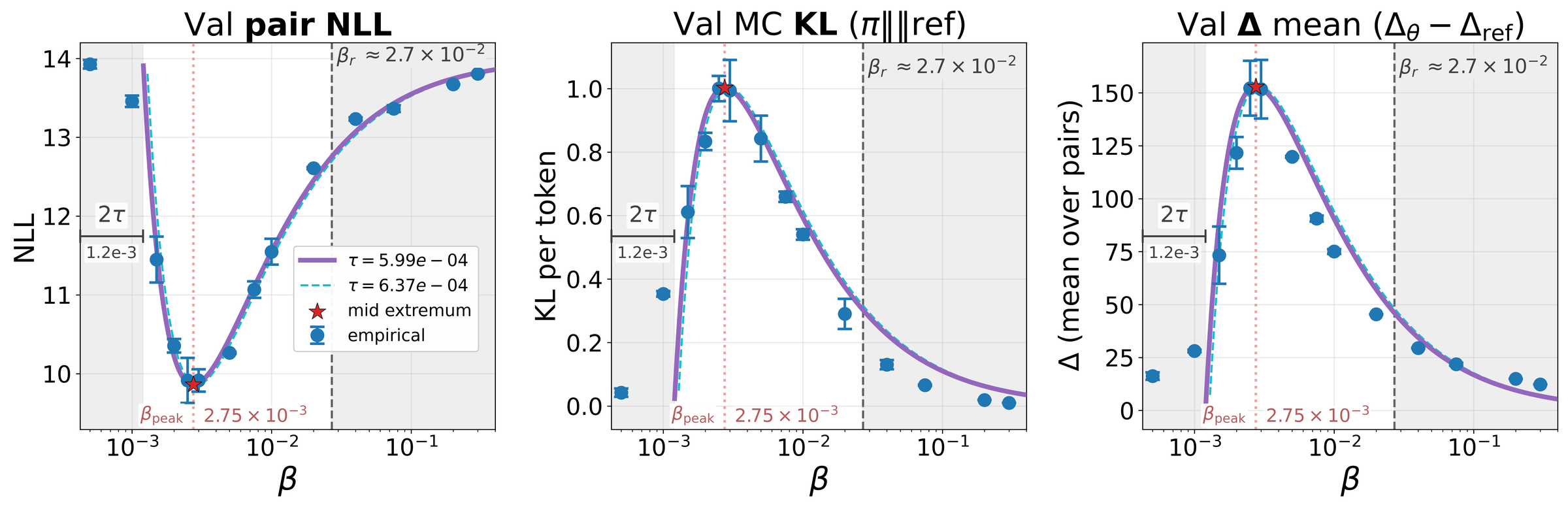}
  \caption{Standard DPO on HelpSteer3 with Qwen3-4B-Instruct-2507 (SGD, $\mathrm{lr}=2\times10^{-4}$, batch size~24): minimum validation pair NLL (with $\beta=1$; Appendix~\ref{app:tau-estimators}), and at that same epoch per-token Monte Carlo KL and mean validation margin $\overline{\Delta}=\Delta_{\theta}-\Delta_{\mathrm{ref}}$, versus $\beta$. Points are seed means $\pm$~std (Table~\ref{tab:app-nll-kl-delta}); far-right $\beta\in\{0.2,0.3\}$ are single-seed. Solid and dashed overlays follow the shape of Equation~\eqref{eq:sat-margin-dpo} with $\tau$ from $\beta_{\mathrm{peak}}$ ($\tau=5.99\times10^{-4}$) and from a shape+scale fit of $\overline{\Delta}(\beta)$ ($\tau=6.37\times10^{-4}$; Appendix~\ref{app:tau-estimators}). The left shaded band marks $\beta\le 2\tau$; the broad right-hand band is only an order-of-magnitude finite-horizon estimate because it freezes an initial batch-averaged Jacobian proxy (Appendix~\ref{app:right-dead-zone}).}
  \label{fig:beta-peak-nll-kl-delta}
\end{figure}

Figure~\ref{fig:beta-peak-nll-kl-delta} compares the observed dense $\beta$-sweep with Equation~\eqref{eq:sat-margin-dpo} under the same HelpSteer3 SGD protocol as Figure~\ref{fig:cause-drift}.
Validation pair NLL (with $\beta=1$; Appendix~\ref{app:tau-estimators}), per-token KL, and mean margin $\overline{\Delta}$ are qualitatively consistent with the predicted shape: a left dead zone, an intermediate peak, and a decaying right slope, with NLL an inverted copy of the saturation curve.
Appendix~\ref{app:tau-estimators} obtains two descriptive estimates of $\tau$: from Equation~\eqref{eq:sat-peak}, $\tau=\beta_{\mathrm{peak}}/4.6\approx 5.99\times10^{-4}$, and from a shape+scale fit of $\overline{\Delta}(\beta)$ to Equation~\eqref{eq:sat-margin-dpo} on $\beta>2\tau$, $\tau\approx 6.37\times10^{-4}$.
The two overlays in Figure~\ref{fig:beta-peak-nll-kl-delta} are almost indistinguishable.
The $\beta=0.002$ vs.\ $0.01$ ordering in Figure~\ref{fig:cause-drift} is consistent with the threshold model and need not indicate an instability.

\subsection{The Standard DPO Loss Is Not Comparable Across \texorpdfstring{$\beta$}{beta}}
\label{sec:loss-incomparability}

The same $\beta$ entanglement also undermines the standard DPO loss as a monitoring signal: the loss depends on $\beta$ and the preference margin only through their product $u=\beta\Delta$, so its value can remain nearly flat during training, or coincide across runs whose policies differ several-fold in KL.
One might expect a training loss to decrease more clearly when the model learns more aggressively and moves farther from the reference policy in KL.
The standard DPO loss need not satisfy this, as illustrated across Figure~\ref{fig:cause-drift}, Figure~\ref{fig:softplus-vs-centered-softplus}, and the HH-RLHF trajectories in Appendix~\ref{app:hh-rlhf-dynamics}: the training dynamics can be more aggressive (Appendix~\ref{app:hh-rlhf-dynamics}), the loss geometry is governed by the negative half-axis (Figure~\ref{fig:softplus-vs-centered-softplus}), and the standard DPO loss can still appear nearly flat during training (Figure~\ref{fig:cause-drift} and Appendix~\ref{app:hh-rlhf-dynamics}, Figure~\ref{fig:intro-overview}).
The reason is that on the negative half-axis, for $\beta_1 < \beta_2$ and $x<0$, we have
\begin{equation}
\label{eq:log-sigma-ordering}
\log \sigma(\beta_1 x) > \log \sigma(\beta_2 x),
\end{equation}
even though, under the KL-control interpretation, the smaller $\beta$ would be expected to permit more aggressive policy movement.
This ordering also breaks cross-$\beta$ comparability when $\beta$ is swept as a hyperparameter (see Section~\ref{sec:discussion}).
For small $\beta$, the standard DPO loss often changes little during training relative to its value at initialization and can appear nearly flat, which complicates monitoring.

\begin{figure}[t]
  \centering
  \includegraphics[width=0.713\linewidth]{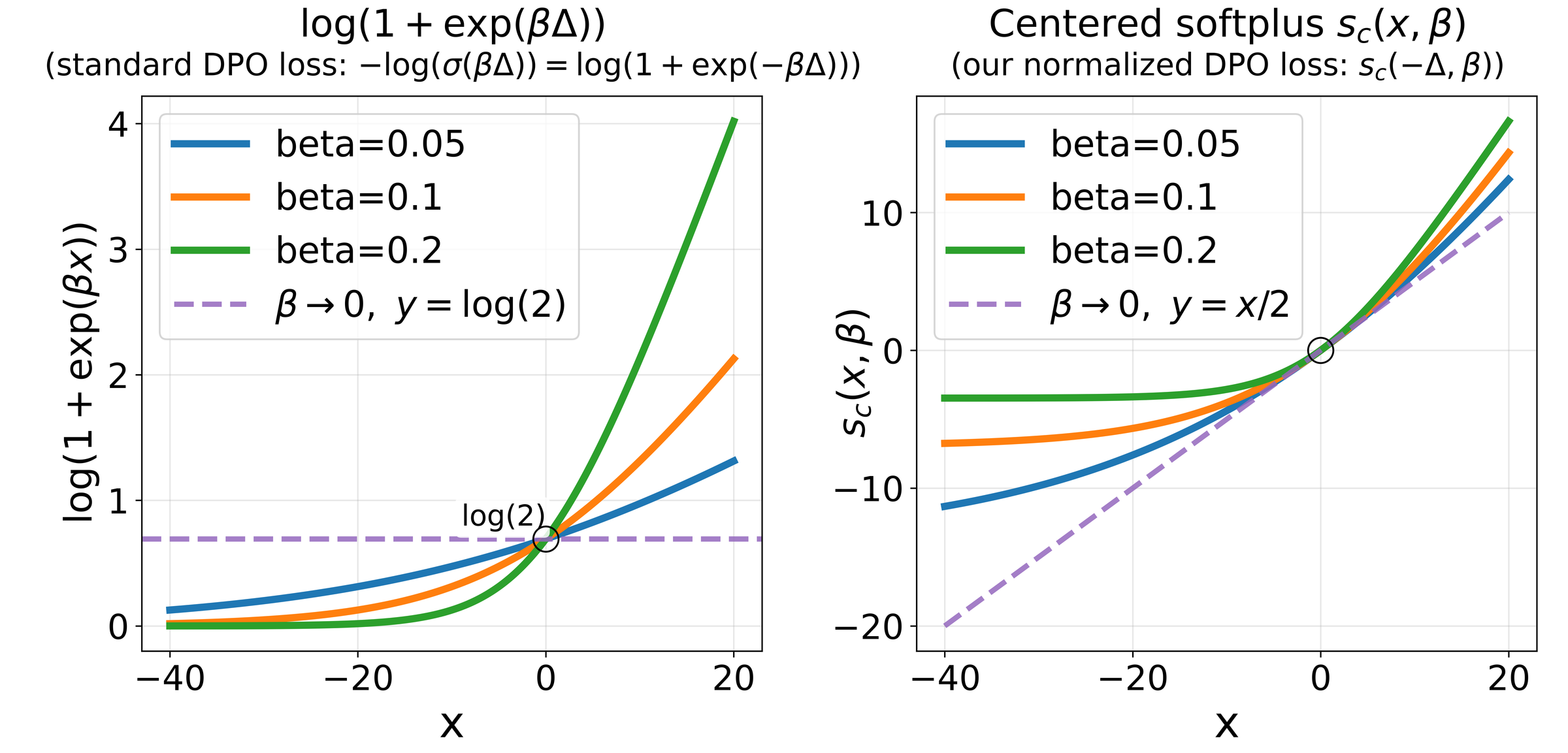}
  \caption{Standard DPO loss $\log\sigma(\beta\Delta)$ and centered softplus $s_c(\Delta;\beta)$. The negative half-axis is the region of primary interest because the loss is optimized mostly there. As $\beta\to 0$, the standard DPO loss degenerates into a constant.}
  \label{fig:softplus-vs-centered-softplus}
\end{figure}

\begin{figure}[t]
  \centering
  \includegraphics[width=0.92\linewidth]{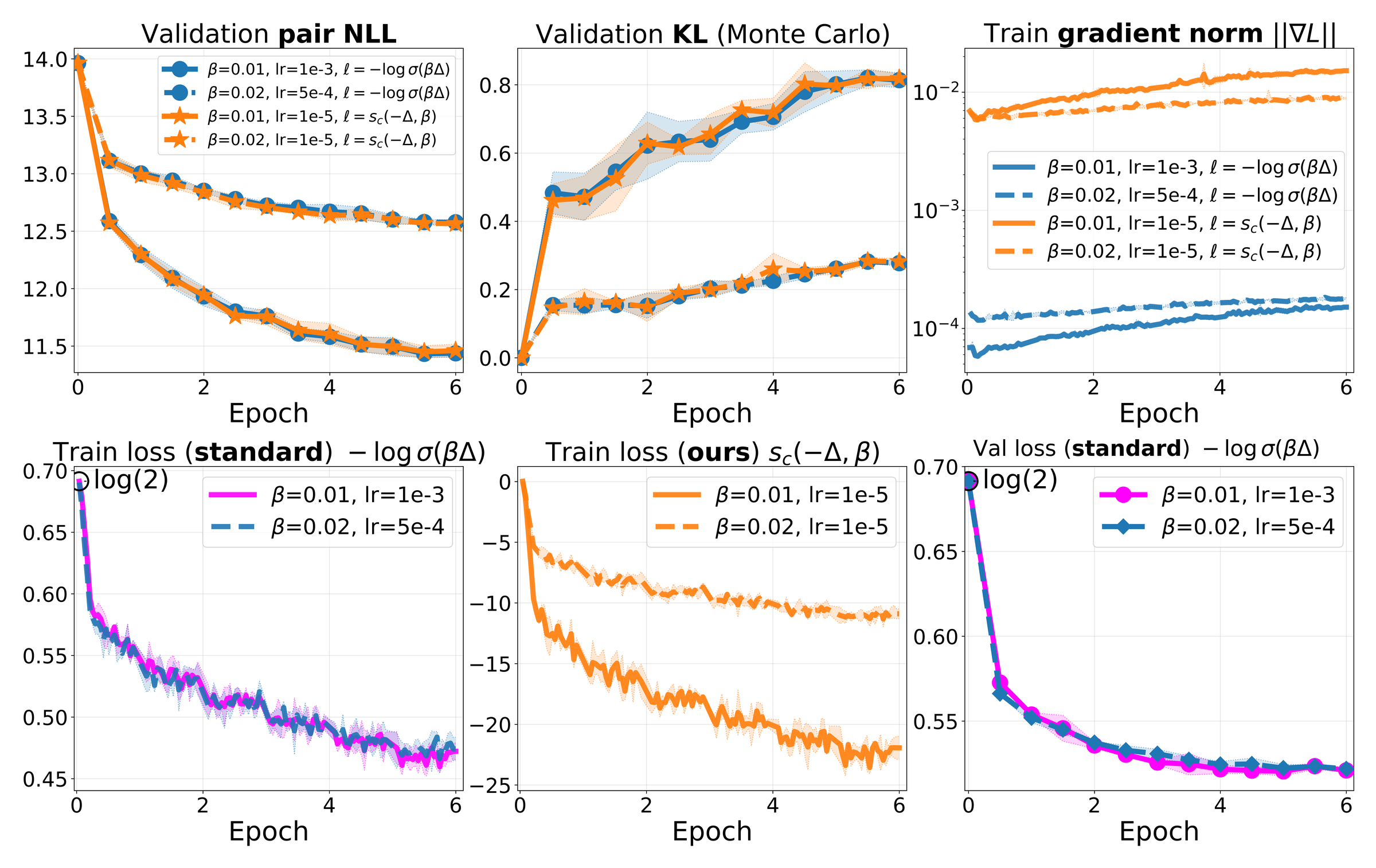}
  \caption{Comparison of training with the normalized centered-softplus DPO objective $s_c(-\Delta;\beta)$ and standard DPO with pairwise loss $-\log\sigma(\beta\Delta)$ on UltraFeedback with Qwen3-4B-Instruct-2507 (batch size 26). Learning rates satisfy $\mathrm{lr}_{\mathrm{norm}}=\beta\,\mathrm{lr}_{\mathrm{standard}}$. With SGD, validation metrics are nearly equivalent. Even when validation metrics differ substantially between $\beta\in\{0.01,0.02\}$, the standard training-loss curves are almost overlapping, whereas for $s_c(-\Delta;\beta)$ the training losses separate noticeably. Curves show the mean over four seeds; shaded bands denote one standard deviation. The comparison tests the predicted scale equivalence; it is not a claim that the normalized objective outperforms a retuned DPO baseline.}
  \label{fig:cause-failures}
\end{figure}

A sharper version of the problem appears when $\beta$ and the learning rate are traded off against each other.
The standard loss depends only on the product $u=\beta\Delta$, and under SGD the update of $\Delta$ carries the prefactor $\eta\beta$ (Equation~\eqref{eq:dpo-marginal-grad}).
Under the scalar frozen-Jacobian approximation, matching $\eta\beta$ predicts similar late-time trajectories in $u=\beta\Delta$.
This behavior is observed approximately in several of our sweeps but is not guaranteed for general minibatch training (Appendix~\ref{app:loss-invariance}).
In those cases, the runs produce nearly identical training and validation loss curves while their physical margins differ; their KL divergences from the reference can therefore differ substantially as well, although that ratio is not fixed by the same identity.
Figure~\ref{fig:cause-failures} shows this effect on UltraFeedback: the two standard-loss runs $(\beta,\mathrm{lr})=(0.01,\,10^{-3})$ and $(0.02,\,5\times10^{-4})$ have nearly identical validation DPO losses, yet their validation KL differs by a factor of about $3$ (Table~\ref{tab:loss-invariance-6epoch}), and validation NLL separates the runs as well.
The same coincidence of the standard DPO loss conceals a large downstream gap: AlpacaEval~2 length-controlled win rates for this pair differ by about $14$ percentage points (Table~\ref{tab:ultrafb-alpaca}, Appendix~\ref{app:normalized-trajectories}).
On HelpSteer3 the same construction yields a KL ratio of $\approx 10$; Appendix~\ref{app:loss-invariance} quantifies both cases, together with Ministral-3B and HH-RLHF comparisons.
Consequently, neither the training loss nor the validation DPO loss can be used to compare runs across $\beta$, to select $\beta$ by loss value, or to detect that one run has moved several times farther from the reference model.
The normalized loss of Section~\ref{sec:normalized} removes this artifact: it tracks the physical margin $\Delta$, so runs with different policy displacement produce visibly different loss curves.

\section{Normalized Objective Formulation}
\label{sec:normalized}

\subsection{Normalized DPO Loss}
\label{sec:normalized-dpo-loss}

A natural normalization is to divide Equation~\eqref{eq:dpo-loss} by $\beta > 0$.
Since multiplication by a positive constant does not change the $\arg\min$, this rescaling preserves the optimal solution.
For the standard softplus term, define
\begin{equation}
s(x;\beta)=\frac{1}{\beta}\log\!\left(1+e^{\beta x}\right).
\end{equation}

Using this notation, the normalized DPO objective can be written as
\begin{equation}
\mathcal{L}_{\mathrm{DPO}}^{\mathrm{norm}}(\Delta;\beta)=s(-\Delta;\beta).
\end{equation}
For $\beta>0$, its marginal gradient is
\begin{equation}
\frac{\partial \mathcal{L}_{\mathrm{DPO}}^{\mathrm{norm}}}{\partial \Delta}
= -\sigma(-\beta \Delta).
\end{equation}
Compared with Equation~\ref{eq:dpo-marginal-grad}, the explicit multiplicative $\beta$ prefactor has been removed.
Thus $\beta$ still controls the shape and saturation of the sigmoid response as an effective inverse preference-noise-scale coefficient, but no longer directly rescales the local update magnitude.
Quantitatively, $|\partial\mathcal{L}_{\mathrm{DPO}}^{\mathrm{norm}}/\partial\Delta|=\sigma(-\beta\Delta)\le 1$ while $\partial^2\mathcal{L}_{\mathrm{DPO}}^{\mathrm{norm}}/\partial\Delta^2=\beta\,\sigma'(\beta\Delta)\le\beta/4$: in the normalized parameterization, $\beta$ bounds the curvature of the pairwise loss without affecting its gradient scale, in line with the low-curvature motivation for centered softplus in \citet{Srinivas2022LCNN}.
For $\beta\Delta\approx 0$, $\log(1+e^{-\beta\Delta})/\beta\approx(\ln 2)/\beta-\Delta/2$.
The $\beta\to 0$ endpoint is handled rigorously in Appendix~\ref{app:prop1-proof}; in particular, the marginal gradient has the well-defined limit $-1/2$.
\subsection{Centered Softplus}
\label{sec:centered-softplus}

To eliminate the $\log(2)/\beta$ offset, we adopt the centered-softplus transformation introduced by \citet{Srinivas2022LCNN}; see Figure~\ref{fig:softplus-vs-centered-softplus} for an illustration of $\log(\sigma(\beta\Delta))$ versus centered softplus:
\begin{equation}
s_c(x;\beta)=s(x;\beta)-\frac{\ln 2}{\beta}
=\frac{1}{\beta}\log\!\left(\frac{1+e^{\beta x}}{2}\right).
\end{equation}
For $\beta>0$, the centered-softplus objective $s_c(-\Delta;\beta)$ is argmin-equivalent to the original DPO loss $-\log\sigma(\beta\Delta)$; the algebraic identity and the functional equivalence over $\theta$ are proved in Appendix~\ref{app:prop1-proof}.
In the limit $\beta\to 0$, the centered softplus converges to a linear map $s_c(x;0)=x/2$.
Accordingly, we define the updated normalized DPO objective as
\begin{equation}
\label{eq:dpo-norm-piecewise}
\mathcal{L}_{\mathrm{DPO}}^{\mathrm{norm}}(\Delta;\beta)=
\left\{
\begin{array}{ll}
s_c(-\Delta;\beta), & \beta>0,\\
-\frac{\Delta}{2}, & \beta=0.
\end{array}
\right.
\end{equation}
This parameterization removes the divergent offset while preserving the minimizers of the original objective for $\beta>0$.
In particular, at $\beta=0$, minimizing $\mathcal{L}_{\mathrm{DPO}}^{\mathrm{norm}}$ is equivalent (up to a constant factor and the fixed reference term) to maximizing the policy log-probability margin between winner and loser responses, i.e., $\log \pi_{\theta}(y_w\mid x)-\log \pi_{\theta}(y_l\mid x)$.
The $\beta=0$ branch should therefore be interpreted as a limiting case rather than as a universally preferable default.
Unlike the $\beta>0$ centered-softplus loss, the linear objective does not saturate for already well-separated preference pairs: its marginal gradient with respect to $\Delta$ remains constant, matching the limit $-\frac{1}{2}$ as $\beta\to 0$.
Consequently, in unconstrained parameterizations it does not by itself impose a finite preferred margin.
In practical language-model fine-tuning, the realized update is still controlled by the optimizer, learning-rate schedule, finite training horizon, gradient clipping, and any explicit or implicit regularization.
We therefore view $\beta=0$ primarily as an endpoint that cleanly separates preference-margin maximization from the $\beta$-controlled saturation behavior present at $\beta>0$.

The normalized DPO objective based on centered softplus addresses the Section~\ref{sec:beta-weakens-updates} issue that, for small $\beta$, the standard DPO loss can remain nearly flat even when optimization is relatively aggressive.
Unlike the standard DPO loss, its negative-half-axis geometry does not collapse toward a nearly constant curve as $\beta$ becomes small (Figure~\ref{fig:softplus-vs-centered-softplus}).
The same comparison highlights a further optimization advantage of the proposed parameterization.
The standard loss $-\log\sigma(\beta\Delta)$ is bounded below by $0$ for every $\beta$ and, as $\beta$ decreases, its entire profile flattens toward the constant $\log 2$, so gradients saturate quickly at small $\beta$.
The centered-softplus loss instead has a $\beta$-dependent lower plateau at $-\ln 2/\beta$: the smaller $\beta$, the deeper the plateau and the longer gradients keep flowing---the marginal gradient $-\sigma(-\beta\Delta)$ stays near $-1/2$ over a range of margins that grows as $1/\beta$---with no premature saturation.
From an optimization standpoint, avoiding vanishing gradients and premature saturation makes the normalized loss the better-behaved objective, while the minimizer is unchanged for $\beta>0$.
Over the same optimization trajectory it retains more variation and separates runs more clearly, making the loss easier to interpret as learning progress (Figure~\ref{fig:cause-failures}).
Together with the hard-margin endpoint in Equation~\ref{eq:softplus-hard-margin-limit} (Appendix~\ref{app:prop1-proof}), this shows that the normalized form has meaningful behavior at both extremes of $\beta$.

For the centered-softplus objective, $|\partial\mathcal{L}_{\mathrm{DPO}}^{\mathrm{norm}}/\partial\Delta|=\sigma(-\beta\Delta)$, so the stopping condition $\sigma(-\beta\Delta)=\tau$ gives the saturation margin
\begin{equation}
\label{eq:sat-margin-norm}
\Delta^{*}=\frac{1}{\beta}\log\frac{1-\tau}{\tau}.
\end{equation}
Unlike the classical threshold in Equation~\eqref{eq:sat-margin-dpo}, this margin scales as $1/\beta$ with no dead zone and no intermediate peak: under the normalized loss the achieved policy deviation is monotone in $1/\beta$ (Appendix~\ref{app:saturation}).

\subsection{Experiments}
\label{sec:experiments}

To verify that the normalized objective behaves as predicted by the gradient-scale analysis, we compare it against standard DPO under a learning-rate rescaling that equalizes the marginal gradient scale.
Because $\nabla_{\theta}\mathcal{L}_{\mathrm{DPO}}^{\mathrm{norm}}=\nabla_{\theta}\mathcal{L}_{\mathrm{DPO}}/\beta$ for $\beta>0$, equal SGD updates require
\[
\mathrm{lr}_{\mathrm{norm}}
=
\beta\,\mathrm{lr}_{\mathrm{standard}}.
\]
The experiment checks the gradient-scale analysis above; it is not meant to establish a performance advantage over a carefully retuned DPO baseline.
Thus SGD trajectories should match closely under this learning-rate relation, up to stochasticity and numerical effects.
Learning-rate retuning can compensate for much of the finite-time speed difference, but this compensation is a separate optimization choice rather than an intrinsic property of the original DPO parameterization.
Figure~\ref{fig:cause-failures} illustrates this prediction in a realistic LLM fine-tuning setup: each centered-softplus run uses $\mathrm{lr}_{\mathrm{norm}}=\beta\,\mathrm{lr}_{\mathrm{standard}}$, and the two objectives produce closely aligned validation trajectories, consistent with their objective-level affine relation.
The same figure shows that the normalized training loss remains distinguishable for $\beta\in\{0.01,0.02\}$, whereas the standard training-loss curves coincide (Section~\ref{sec:loss-incomparability}).
Table~\ref{tab:ultrafb-alpaca} (Appendix~\ref{app:normalized-trajectories}) reports AlpacaEval~2 and IFEval scores for these four curves: the overlapping standard-loss pair differs substantially downstream, while each standard-DPO configuration agrees with its learning-rate-rescaled centered-softplus counterpart.
Table~\ref{tab:hsteer-benchmarks} (Appendix~\ref{app:normalized-trajectories}) extends the comparison to post-training AlpacaEval and IFEval scores on HelpSteer3 over eight paired seeds. No statistically significant difference is detected at $n=8$ by the seed-wise Wilcoxon signed-rank tests ($p>0.05$ for all reported metrics); these tests do not establish practical equivalence or exclude small systematic effects.
Figure~\ref{fig:hsteer-sgd-centered-comparison} (Appendix~\ref{app:normalized-trajectories}) reproduces the same comparison on HelpSteer3 with curves averaged over eight seeds; shaded bands denote one standard deviation.
The remaining small gaps may reflect stochasticity and numerical nondeterminism in LLM fine-tuning, but the present sample size does not distinguish these sources from a small systematic effect.
The two \emph{standard-loss} UltraFeedback runs in Figure~\ref{fig:cause-failures} are exactly the loss-invariance pair analyzed in Section~\ref{sec:loss-incomparability} and Appendix~\ref{app:loss-invariance}.

We also trained with the normalized centered-softplus DPO objective under additional hyperparameter settings and on other preference datasets, with AdamW (Figure~\ref{fig:hsteer-adam}, Appendix~\ref{app:normalized-trajectories}), including $\beta=0$ with the linear branch in Equation~\ref{eq:dpo-norm-piecewise}; an additional graph for the normalized centered-softplus loss is shown in Figure~\ref{fig:hsteer-adam-single} (Appendix~\ref{app:normalized-trajectories}).
Across these runs, stronger policy evolution corresponds to a faster decrease in the normalized training loss.
The $\beta=0$ limiting case included in Figure~\ref{fig:hsteer-adam} is comparatively less stable than any $\beta>0$ run, since the linear branch in Equation~\ref{eq:dpo-norm-piecewise} is unbounded below; nevertheless, trajectories with stronger policy movement still show the fastest training-loss decrease.
Figure~\ref{fig:hsteer-adam} also plots the standard DPO pairwise loss on validation, although training optimizes the normalized centered-softplus objective.
The validation pair NLL echoes the training pattern: the $\beta=0$ and $\beta=0.002$ runs, which learn most aggressively and carry the highest gradient norms, attain the lowest pair NLL, whereas the comparatively weak $\beta=0.015$ run remains the highest.
The validation \emph{standard} loss, by contrast, shows the inverted ordering ($\beta=0.015$ is lowest, and $\beta=0$ sits identically at $\log 2$), reflecting the ordering artifact in Equation~\eqref{eq:log-sigma-ordering}: smaller $\beta$ pushes $-\log\sigma(\beta\Delta)$ toward the random-baseline value $\log 2$ regardless of the actual policy movement (see also Appendix~\ref{app:loss-invariance}).


\section{Discussion}
\label{sec:discussion}

The centered-softplus reformulation preserves the minimizer of the original DPO objective for $\beta>0$ while making the gradient-scale effect of $\beta$ explicit; it is not meant to define a different optimum or guarantee a universal gain over a carefully retuned DPO baseline. Its purpose is a cleaner parameterization in which the preference-noise scale and the learning rate control separate aspects of training, with $\beta=0$ providing a well-defined limiting objective, and in which the loss remains an informative training signal that is comparable across $\beta$ (Section~\ref{sec:loss-incomparability}, Appendix~\ref{app:loss-invariance}).

Conceptually, in KL-regularized RLHF, $\lambda$ is the coefficient of the explicit $\lambda\,\mathrm{KL}(\pi_{\theta}\,\|\,\pi_{\mathrm{ref}})$ term \citep{Xiong2024IterativeDPO}. By contrast, after substituting the reward--policy relation into a logistic preference model with annotator-noise scale $s_a$, the DPO objective depends on the effective coefficient $\beta=\lambda/s_a$ \citep[Eq.~6]{Mitchell2023DPO}. The scales $\lambda$ and $s_a$ are not separately identifiable from the DPO likelihood without fixing a reward normalization. Once the likelihood is used as the training loss, $\beta$ controls both its sharpness in policy-log-ratio coordinates and the scale of its gradients.
This reading is consistent with comparative and hybrid studies: \citet{Xu2024DPOvsPPO} contrast PPO's explicit, tunable KL penalty with DPO, which has no explicit KL term and controls deviation only through the preference likelihood and its coefficient $\beta$; hybrid methods such as DICE \citep{Chen2025DICE} and RTO \citep{Zhong2025RTO} likewise treat $\beta$ as a scale parameter of the preference loss rather than as a KL multiplier, and $\beta$-DPO \citep{Wu2024BetaDPO} adapts it as a loss-sharpness knob.

The centering constant $\ln 2$ in the normalized loss also has a statistical meaning.
In the Gumbel-noise view of Bradley--Terry \citep{Sun2025RethinkingBT}, $P(y_w\succ y_l)=\sigma((r_w-r_l)/s_a)$. Appendix~\ref{app:bt-temperature} shows that substituting $r_w-r_l=\lambda\Delta$ gives $P(y_w\succ y_l)=\sigma(\beta\Delta)$ with $\beta=\lambda/s_a$, and that $\mathcal{L}_{\mathrm{DPO}}-\log 2$ equals the negative log-likelihood ratio of the current preference model against the random-choice baseline $P=1/2$.
Dividing by $\beta$ removes the effective inverse preference-noise-scale prefactor from the marginal gradient, yielding the normalized centered-softplus loss; it should not be interpreted as measuring literal annotator noise unless the reward/KL normalization is fixed. The same construction carries over to the Gaussian (Thurstone) model (Appendix~\ref{app:bt-temperature}).
The practical consequence for cross-$\beta$ comparability of loss values was established in Section~\ref{sec:loss-incomparability}.

\section{Conclusion}
\label{sec:conclusion}

We recommend considering the normalized centered-softplus objective when $\beta$-dependent gradient scaling or loss comparability across $\beta$ is a practical concern.
This substitution:
\begin{itemize}
\item resolves gradient attenuation at small $\beta$, so that the inverse preference-noise scale no longer silently rescales the effective update magnitude;
\item makes changes in the preference margin $\Delta$ and validation pair NLL monotone in $\beta$ (equivalently, in $1/\beta$), without the dead-zone / intermediate-peak behavior of the classical threshold. In the reported sweeps, validation KL co-moves with that margin over a wide range of $\beta$, without a claim of a universal KL--margin law;
\item makes training and validation losses comparable across different $\beta$, and removes cases in which nearly identical training and validation loss curves hide substantially different policies.
\end{itemize}
Technically, the change is a local rewrite of the loss evaluation and is straightforward to drop into existing DPO codebases (e.g., TRL) without restructuring the trainer or the data pipeline.

\section*{Reproducibility, Data, and Compute}

Code for this work is publicly available at \url{https://github.com/ivankru/bayesian_dpo}.
We train Qwen3-4B-Instruct-2507, Qwen2.5-3B-Instruct, Ministral-3B-Instruct, and Mamba-2 2.7B (with LoRA) on three public pairwise preference datasets from Hugging Face: HelpSteer3-Preference \citep{HelpSteer3Dataset,HelpSteer3Preference2025}, UltraFeedback Binarized \citep{UltraFeedbackBinarizedDataset,UltraFeedback2023}, and PKU-processed HH-RLHF \citep{PKUProcessedHHRLHF,AnthropicHHRLHF,HHRLHF2022}.
For downstream instruction-following checks (Tables~\ref{tab:hsteer-benchmarks} and~\ref{tab:ultrafb-alpaca}), we additionally use Qwen2.5-14B-Instruct as an LLM judge for AlpacaEval~2 \citep{Dubois2024AlpacaEval2} and the official verifiable IFEval benchmark \citep{Zhou2023IFEval} without a judge model.
All experiments ran on NVIDIA A100-SXM4-80GB GPUs.
Dataset processing and splits, per-run hyperparameters, the Monte Carlo KL protocol, training times, and software versions are detailed in Appendix~\ref{app:training-details}; full run configurations are included in the released code.

\section*{Limitations and Broader Impacts}

Our empirical evidence covers three preference datasets and four backbones (Qwen3-4B-Instruct-2507, Qwen2.5-3B-Instruct, Ministral-3B-Instruct, and Mamba-2 2.7B), all fine-tuned with LoRA at moderate scale, so transfer to larger models, full fine-tuning, and other data regimes remains to be verified. Minimizer equivalence for $\beta>0$ is an objective-level statement: finite-time trajectories can still differ across optimizers and implementations. Our main sweeps use four to five seeds; downstream checks cover a HelpSteer3 comparison (AlpacaEval~2 and IFEval in Table~\ref{tab:hsteer-benchmarks}) and AlpacaEval~2 and IFEval on the UltraFeedback curves of Figure~\ref{fig:cause-failures} (Table~\ref{tab:ultrafb-alpaca}), and broader robustness evaluations (e.g., noisy or adversarial labels, out-of-domain prompts, long-horizon iterative preference updates) remain for future work.

This work reformulates an existing preference-optimization objective and does not introduce a new capability class; its risks are those of language-model alignment more broadly, including misuse of better-aligned systems and biases inherited from preference data and annotator judgments. We recommend standard deployment safeguards: safety evaluation, bias and robustness assessment, and human oversight in high-stakes settings.

\clearpage
{
\small
\bibliographystyle{plainnat}
\bibliography{custom}
}

\clearpage
\appendix
\setcounter{figure}{0}
\setcounter{table}{0}
\renewcommand{\thefigure}{\thesection\arabic{figure}}
\renewcommand{\thetable}{\thesection\arabic{table}}
\renewcommand{\theHfigure}{appendix.\thesection.\arabic{figure}}
\renewcommand{\theHtable}{appendix.\thesection.\arabic{table}}
\makeatletter
\@addtoreset{figure}{section}
\@addtoreset{table}{section}
\makeatother
\section{Marginal Gradients in Selected DPO-Family Variants}
\label{app:dpo-variant-gradients}

\setcounter{equation}{0}
\renewcommand{\theequation}{A.\arabic{equation}}
\renewcommand{\theHequation}{\theequation}

Table~\ref{tab:dpo-losses} summarizes pairwise objectives in the margin $\Delta$.
The unpaired KTO objective \citep{Ethayarajh2024KTO} instead uses binary labels $b\in\{0,1\}$ with $v=\log\pi_{\theta}(y\mid x)-\log\pi_{\mathrm{ref}}(y\mid x)$ and reference offset $z_0$, and its marginal gradients $\partial\mathcal{L}_{\mathrm{KTO}}/\partial v$ likewise carry an explicit $\beta$ prefactor alongside the class weights $\lambda_D$ and $\lambda_U$.
SimPO \citep{Meng2024SimPO} is reference-free: it replaces $\Delta$ with the length-normalized policy-only margin $m=\frac{1}{|y_w|}\log\pi_{\theta}(y_w\mid x)-\frac{1}{|y_l|}\log\pi_{\theta}(y_l\mid x)$ (the length-normalized counterpart of $\Delta_{\theta}$, to which it reduces when $|y_w|=|y_l|$) and subtracts a target margin $\gamma$, yet its marginal gradient $\partial\mathcal{L}_{\mathrm{SimPO}}/\partial m=-\beta\,\sigma(\gamma-\beta m)$ retains the same explicit $\beta$ prefactor.
\paragraph{Adaptive margin in $\alpha$-DPO.}
$\alpha$-DPO \citep[Eq.~13]{Wu2025AlphaDPO} uses the length-normalized policy-only margin
\[
m_{\theta}
:=
\frac{1}{|y_w|}\log\pi_{\theta}(y_w\mid x)
-
\frac{1}{|y_l|}\log\pi_{\theta}(y_l\mid x),
\qquad
u:=\beta m_{\theta},
\]
and the instance-adaptive offset
\[
c_{\alpha}
:=
\operatorname{sg}\!\left[\gamma+\alpha M^{*}(x,y_w,y_l)\right],
\qquad
M^{*}:=\frac{M-\mu_M}{\sigma_M},
\]
where
\[
M
:=
\beta\log
\frac{\pi_{\theta}(y_w\mid x)\pi_{\mathrm{ref}}(y_l\mid x)}
{\pi_{\mathrm{ref}}(y_w\mid x)\pi_{\theta}(y_l\mid x)}
\]
and $\operatorname{sg}$ denotes stop-gradient.
Its per-example objective and marginal gradient are therefore
\[
\mathcal{L}_{\alpha\text{-}\mathrm{DPO}}
=
-\log\sigma(u-c_{\alpha}),
\qquad
\frac{\partial\mathcal{L}_{\alpha\text{-}\mathrm{DPO}}}{\partial m_{\theta}}
=
-\beta\,\sigma(c_{\alpha}-u).
\]
Thus $\alpha$-DPO retains the explicit $\beta$ gradient prefactor, but its adaptive offset is not a fixed margin and the objective is not solely a function of the standard unnormalized DPO margin $\Delta$.
TDPO \citep{Zeng2024TDPO} moves to the token level and augments the margin with a sequential forward-KL correction, optimizing $-\log\sigma(\beta M)$ with $M=\Delta-\delta_{\mathrm{SeqKL}}$, where $\delta_{\mathrm{SeqKL}}=D_{\mathrm{SeqKL}}(x,y_l;\pi_{\mathrm{ref}}\,\|\,\pi_{\theta})-D_{\mathrm{SeqKL}}(x,y_w;\pi_{\mathrm{ref}}\,\|\,\pi_{\theta})$ (in TDPO$_2$ the correction is weighted by $\alpha$ under a stop-gradient); its marginal gradient $\partial\mathcal{L}_{\mathrm{TDPO}}/\partial M=-\beta\,\sigma(-\beta M)$ likewise carries the prefactor and equals $-\beta/2$ at initialization, exactly as in DPO.
The main exception is IPO, where $\beta$ has a different semantic role (as a target/margin scale in the objective) rather than acting as a direct multiplicative coefficient on the DPO-style gradient term; note, however, that the coupling is inverted rather than removed: at $\Delta=0$ the IPO marginal gradient equals $-1/\beta$, so decreasing $\beta$ inflates rather than suppresses the update magnitude.
For ROPO and AOT, the $\beta$ dependence is present but enters indirectly through the robust or distributional weighting terms $w_{\alpha}(\Delta)$ and $w_i^{\mathrm{dist}}$; the coupling may therefore be partially mediated by these weights in practice, and the effective suppression of update magnitude need not be strictly linear in $\beta$ for all operating points.

\section{Adam Compensation Breakdown: Closed-Form Threshold}
\label{app:adam-threshold}

\setcounter{equation}{0}
\renewcommand{\theequation}{B.\arabic{equation}}
\renewcommand{\theHequation}{\theequation}

We derive a closed-form condition under which Adam's adaptive scaling ceases to compensate for the linear $\beta$ factor in DPO gradients, and provide a threshold $\beta_{\mathrm{crit}}$ that characterizes the transition.

\paragraph{Setup.}
Write the per-parameter DPO gradient as $g_t = \beta h_t$, where $h_t$ is the full non-$\beta$ factor of the DPO gradient and $\beta>0$ is the effective DPO coefficient.
By the chain rule, the per-parameter gradient is
$\partial\mathcal{L}/\partial\theta_j = -\beta\,\sigma(-\beta\Delta_t)\cdot\partial\Delta_t/\partial\theta_j$,
so $h_{t,j} = -\sigma(-\beta\Delta_t)\cdot\partial\Delta_t/\partial\theta_j$ combines the sigmoid factor and the log-probability-gap Jacobian.
The quantity $\sqrt{\hat{v}_t^h}$ is therefore the root-mean-square of this product over the moment history, and it evolves during training as both $\Delta_t$ and the policy Jacobian change.
Under Adam with first and second moment decay rates $\rho_1,\rho_2\in(0,1)$, optimizer learning rate $\eta$, and numerical stabilizer $\epsilon>0$, the moment updates are:
\begin{align}
m_t &= \rho_1 m_{t-1} + (1-\rho_1)\,\beta h_t, \\
v_t &= \rho_2 v_{t-1} + (1-\rho_2)\,\beta^2 h_t^2.
\end{align}
Let $\hat{m}_t^h$ and $\hat{v}_t^h$ denote the bias-corrected moment estimates of $h_t$ and $h_t^2$, respectively, defined in the usual way ($\hat{m}_t^h = m_t^h/(1-\rho_1^t)$ and $\hat{v}_t^h = v_t^h/(1-\rho_2^t)$, where $m_t^h,v_t^h$ accumulate $h_t$ and $h_t^2$ without the $\beta$ prefactor). Then $\hat{m}_t = \beta\hat{m}_t^h$ and $\hat{v}_t = \beta^2\hat{v}_t^h$, so the Adam parameter update is:
\begin{equation}
\label{eq:adam-update-rewritten}
\theta_{t+1} - \theta_t
= -\eta\frac{\hat{m}_t}{\sqrt{\hat{v}_t}+\epsilon}
= -\eta\frac{\beta\hat{m}_t^h}{\sqrt{\beta^2\hat{v}_t^h}+\epsilon}
= -\eta\frac{\hat{m}_t^h}{\sqrt{\hat{v}_t^h}+\epsilon/\beta}.
\end{equation}

\paragraph{Proposition A1 (Two-regime characterization).}
Define the critical threshold
\begin{equation}
\label{eq:beta-crit}
\beta_{\mathrm{crit}}(\epsilon,\hat{v}^h) \;:=\; \frac{\epsilon}{\sqrt{\hat{v}_t^h}}.
\end{equation}
The Adam update in Eq.~\eqref{eq:adam-update-rewritten} satisfies:
\begin{itemize}
    \item \emph{Cancellation regime} ($\beta\gg\beta_{\mathrm{crit}}$): $\epsilon/\beta \ll \sqrt{\hat{v}_t^h}$, so the denominator $\approx\sqrt{\hat{v}_t^h}$ and
    \[
        \theta_{t+1}-\theta_t \;\approx\; -\eta\frac{\hat{m}_t^h}{\sqrt{\hat{v}_t^h}},
    \]
    which is independent of $\beta$. Adam fully compensates for the gradient-scale coupling.
    \item \emph{Non-cancellation regime} ($\beta\ll\beta_{\mathrm{crit}}$): $\epsilon/\beta \gg \sqrt{\hat{v}_t^h}$, so the denominator $\approx\epsilon/\beta$ and
    \[
        \theta_{t+1}-\theta_t \;\approx\; -\frac{\eta\beta}{\epsilon}\hat{m}_t^h,
    \]
    which is proportional to $\beta$, the same linear suppression as plain SGD.
\end{itemize}

\emph{Proof.} Equation~\eqref{eq:adam-update-rewritten} follows directly from the linearity of the moment recurrences in $g_t=\beta h_t$.
The two limiting cases follow from comparing $\epsilon/\beta$ to $\sqrt{\hat{v}_t^h}$. $\square$

\paragraph{Numerical estimates for DPO.}
The PyTorch AdamW default is $\epsilon=10^{-8}$.
At initialization, $\Delta\approx 0$ so $\sigma(-\beta\Delta)\approx 1/2$, giving
\[
    \sqrt{\hat{v}^h}\big|_{\mathrm{init}}
    \;\approx\;
    \tfrac{1}{2}\,\mathrm{RMS}_j\!\left(\frac{\partial\Delta}{\partial\theta_j}\right),
\]
where $\mathrm{RMS}_j(\cdot)$ is the per-element root-mean-square of the log-probability-gap Jacobian.
The sigmoid factor $1/2$ alone does \emph{not} determine $\sqrt{\hat{v}^h}$; the actual magnitude is set by the per-element Jacobian $|\partial\Delta/\partial\theta_j|$, which is architecture- and dataset-dependent.
Consequently, $\beta_{\mathrm{crit}}=\epsilon/\sqrt{\hat{v}^h}$ is best estimated from observed gradient statistics and can vary substantially across models and training stages.

\paragraph{Estimating $\beta_{\mathrm{crit}}$ from gradient logs.}
For the TRL runs used in this appendix, the logged gradient metric is the global pre-clipping norm
$G_t^{\mathrm{norm}}=\|g_t\|_2$.
With $g_t=\beta_{\mathrm{ref}}h_t$, this gives
\[
    G_t^{\mathrm{norm}}
    \;=\;
    \beta_{\mathrm{ref}}\|h_t\|_2.
\]
To map this global norm to Adam's per-coordinate denominator, we use
$\|h_t\|_2\approx \sqrt{d}\,\sqrt{\hat v_t^h}$, where $d$ is the number of
\emph{trainable} parameters that actually receive gradients and enter Adam's
per-parameter moment estimates. In our LoRA runs, $d$ is exactly the number of
LoRA adapter parameters (the frozen base-model weights contribute no gradient
and are excluded from both $G_t^{\mathrm{norm}}$ and the moment estimates), and we take $d$
directly from the \texttt{adapter\_config.json} / PEFT \texttt{print\_trainable\_parameters()}
log for each run rather than treating it as a free parameter. Then
\begin{equation}
\label{eq:bcrit-practical}
\beta_{\mathrm{crit}} \;\approx\; \frac{\epsilon\,\beta_{\mathrm{ref}}\sqrt{d}}{G_t^{\mathrm{norm}}}.
\end{equation}
Here $\beta_{\mathrm{crit}}$ is the 50\%-attenuation point (Adam denominator terms
$\sqrt{\hat v_t^h}$ and $\epsilon/\beta$ are equal).

For a practical ``noticeable'' threshold, define $\beta_{\mathrm{notice}}(\delta)$ by requiring
a relative attenuation $\delta$ of the compensated Adam step.
From Eq.~\eqref{eq:adam-update-rewritten}, the attenuation factor is
$A(\beta)=1/(1+\epsilon/(\beta\sqrt{\hat v_t^h}))$, hence
\begin{equation}
\beta_{\mathrm{notice}}(\delta)
\;=\;
\frac{1-\delta}{\delta}\,\beta_{\mathrm{crit}}.
\end{equation}
At $\delta=10\%$, this gives $\beta_{\mathrm{notice}}(10\%)=9\,\beta_{\mathrm{crit}}$.

\paragraph{Application to the reported experiments.}
Since $\sqrt{\hat{v}^h}$ is a property of the DPO loss landscape and the model's gradient structure,
independent of the optimizer, we estimate it from the directly logged TRL \texttt{grad\_norm}
values for the same runs, together with the exact LoRA trainable-parameter count $d$ for each run.
Both runs use LoRA with rank $r=16$ and $\alpha=32$, but with different target modules and therefore
different $d$: for Mamba-2 2.7B (target modules \texttt{in\_proj}, \texttt{x\_proj}, \texttt{dt\_proj}),
$d=13{,}451{,}264$; for Qwen3-4B (target modules \texttt{q\_proj}, \texttt{k\_proj}, \texttt{v\_proj},
\texttt{o\_proj}, \texttt{gate\_proj}, \texttt{up\_proj}, \texttt{down\_proj} across 36 layers),
$d=33{,}030{,}144$.
Using PyTorch default $\epsilon=10^{-8}$ and Eq.~\eqref{eq:bcrit-practical} with the corresponding $d$
for each run, we report both $\beta_{\mathrm{crit}}$ (50\% attenuation) and
$\beta_{\mathrm{notice}}(10\%)$ (10\% attenuation); no free or calibrated parameter enters this
computation.
Table~\ref{tab:bcrit-estimates} lists two rows per run (first and last epoch):
for Mamba-2 on HH-RLHF, step~100 and step~47{,}910;
for Qwen on UltraFeedback, step~10 and step~22{,}890.

\begin{table}[h]
\centering
\small
\setlength{\tabcolsep}{4pt}
\caption{Estimates of $\beta_{\mathrm{crit}}$ (50\% attenuation) and
$\beta_{\mathrm{notice}}(10\%)$ (10\% attenuation) from training logs.
Rows alternate first/last epoch within each run (see text above).
$G_t=\texttt{grad\_norm}$ from TRL \texttt{DPOTrainer}
(global gradient norm, pre-clipping, standard DPO loss).
$d$ is the exact number of trainable LoRA parameters for that run
(from \texttt{adapter\_config.json} / PEFT \texttt{print\_trainable\_parameters()}).
Numerics use $\epsilon=10^{-8}$.}
\label{tab:bcrit-estimates}
\begin{tabular}{lllccccc}
\toprule
Dataset & Model & Epoch & $\beta_{\mathrm{ref}}$ & $d$ & $G_t$ & $\beta_{\mathrm{crit}}\approx$ & $\beta_{\mathrm{notice}}$ \\
\midrule
HH-RLHF & Mamba-2 2.7B & first & $10^{-5}$ & $1.35\times10^{7}$ & $3.4\times10^{-5}$ & $1.1\times10^{-5}$ & $9.8\times10^{-5}$ \\
HH-RLHF & Mamba-2 2.7B & last & $10^{-5}$ & $1.35\times10^{7}$ & $2.0\times10^{-1}$ & $1.8\times10^{-9}$ & $1.7\times10^{-8}$ \\
UltraFB & Qwen3-4B & first & $10^{-6}$ & $3.30\times10^{7}$ & $8.9\times10^{-5}$ & $6.5\times10^{-7}$ & $5.8\times10^{-6}$ \\
UltraFB & Qwen3-4B & last & $10^{-6}$ & $3.30\times10^{7}$ & $1.3\times10^{-3}$ & $4.4\times10^{-8}$ & $4.0\times10^{-7}$ \\
\bottomrule
\end{tabular}
\end{table}

Three conclusions follow directly from the table.

\textbf{(i) At initialization, both runs sit close to their own $\beta_{\mathrm{crit}}$.}
For Mamba-2, $\beta_{\mathrm{ref}}=10^{-5}$ is within a factor of $1.1$ of
$\beta_{\mathrm{crit}}\approx1.1\times10^{-5}$ at initialization, i.e., the run starts
almost exactly at the 50\%-attenuation point, independently explaining why epoch-1
gradients are $\sim3\times10^{-5}$ (near-zero) and the DPO loss remains pinned at $\ln 2$
(Appendix~\ref{app:hh-rlhf-dynamics}). For UltraFeedback, $\beta_{\mathrm{ref}}=10^{-6}$
is within a factor of $1.5$ of $\beta_{\mathrm{crit}}\approx6.5\times10^{-7}$, and
$\beta_{\mathrm{notice}}(10\%)\approx5.8\times10^{-6}$ is of the same order of magnitude
as the empirically reported onset scale $\beta\sim10^{-5}$ for Adam-visible coupling, without
any calibrated or fitted parameter, since $d$ is fixed by the logged LoRA configuration.

\textbf{(ii) $\beta_{\mathrm{crit}}$ and $\beta_{\mathrm{notice}}$ are strongly stage-dependent.}
As training progresses and $G_t^{\mathrm{norm}}$ increases by three to four orders of magnitude,
both thresholds drop by a similar amount (e.g., $\beta_{\mathrm{crit}}$ falls from
$1.1\times10^{-5}$ to $1.8\times10^{-9}$ for Mamba-2). Hence Adam's $\epsilon$-dominated
regime is most consequential at initialization and early training, exactly when the
policy has not yet moved and $\Delta\approx0$.

\textbf{(iii) Mamba-2 and UltraFeedback differ in both $d$ and early gradient scale.}
Writing $\beta_{\mathrm{crit}}=\epsilon\sqrt{d}/\|h_t\|_2$ with $\|h_t\|_2=G_t^{\mathrm{norm}}/\beta_{\mathrm{ref}}$ separates the two effects.
Mamba-2's initial non-$\beta$ gradient scale is much smaller than Qwen's ($\|h\|_2\approx 3.4$ vs.\ $89$), which by itself raises Mamba-2's threshold by $\approx 26\times$; its $2.5\times$ fewer trainable LoRA parameters ($1.35\times10^7$ vs.\ $3.30\times10^7$, from the different target-module sets) partially offset this by lowering the threshold by $\sqrt{2.5}\approx1.6\times$.
The net effect, $26/1.6\approx 17\times$, matches the tabulated $\beta_{\mathrm{crit}}$ ratio at initialization, showing that both the gradient-norm scale and the trainable-parameter count materially shift the threshold, in opposite directions here.

The training dynamics of $\sqrt{\hat{v}^h}$ are $\beta$-asymmetric, as confirmed by the experimental figures.
For \emph{small} $\beta$, the sigmoid stays near $1/2$ and $\sqrt{\hat{v}^h}$ tracks the Jacobian: as the policy moves, the Jacobian can grow and gradients increase.
Across all reported runs, in Figure~\ref{fig:ultrafb-classic-trl} (AdamW, UltraFeedback, $\beta=10^{-6}$) gradient norms trend upward throughout training; in Figure~\ref{fig:hh-rlhf-drift} (HH-RLHF, SGD) small-$\beta$ trajectories show a delayed surge in gradient magnitude; and in Figure~\ref{fig:cause-drift} (HelpSteer3, SGD), larger $\beta$ generally produces larger gradient norms, while each curve rises over the run, with the $\beta=2\times10^{-3}$ curve showing the most pronounced increase.
Growing $\sqrt{\hat{v}^h}$ lowers $\beta_{\mathrm{crit}}$ over time, partially self-correcting the non-compensation condition.
However, this correction is delayed: learning is suppressed during the early phase before $\beta_{\mathrm{crit}}$ has fallen sufficiently.
For \emph{large} $\beta$, sigmoid saturation ($\sigma(-\beta\Delta)\to 0$ as $\Delta$ grows) suppresses $h_t$ even as the Jacobian increases, keeping $\sqrt{\hat{v}^h}$ bounded and $\beta_{\mathrm{crit}}$ relatively stable; in this regime $\beta\gg\beta_{\mathrm{crit}}$ from the outset.

For $\beta\in[0.01,0.2]$ (standard DPO range), $\beta$ exceeds $\beta_{\mathrm{crit}}$ by many orders of magnitude for any reasonable Jacobian scale, which is why the Adam compensation is effective and the coupling is routinely unnoticed.

\paragraph{Effect of gradient clipping and $\epsilon$ choices.}
Gradient clipping reduces the effective per-element gradient magnitude, which lowers $\sqrt{\hat{v}^h}$ and raises $\beta_{\mathrm{crit}}$.
Increasing $\epsilon$ (sometimes used for numerical stability in low-gradient regimes) directly raises $\beta_{\mathrm{crit}}$ and can push moderate $\beta$ values into the partial non-cancellation regime.
Both effects can exacerbate the gradient-scale coupling in practice, particularly in combination with very small $\beta$.

\newpage
\section{Additional Training Dynamics}
\label{app:hh-rlhf-dynamics}

\setcounter{equation}{0}
\renewcommand{\theequation}{C.\arabic{equation}}
\renewcommand{\theHequation}{\theequation}

\begin{figure}[H]
  \centering
  \includegraphics[width=0.98\linewidth]{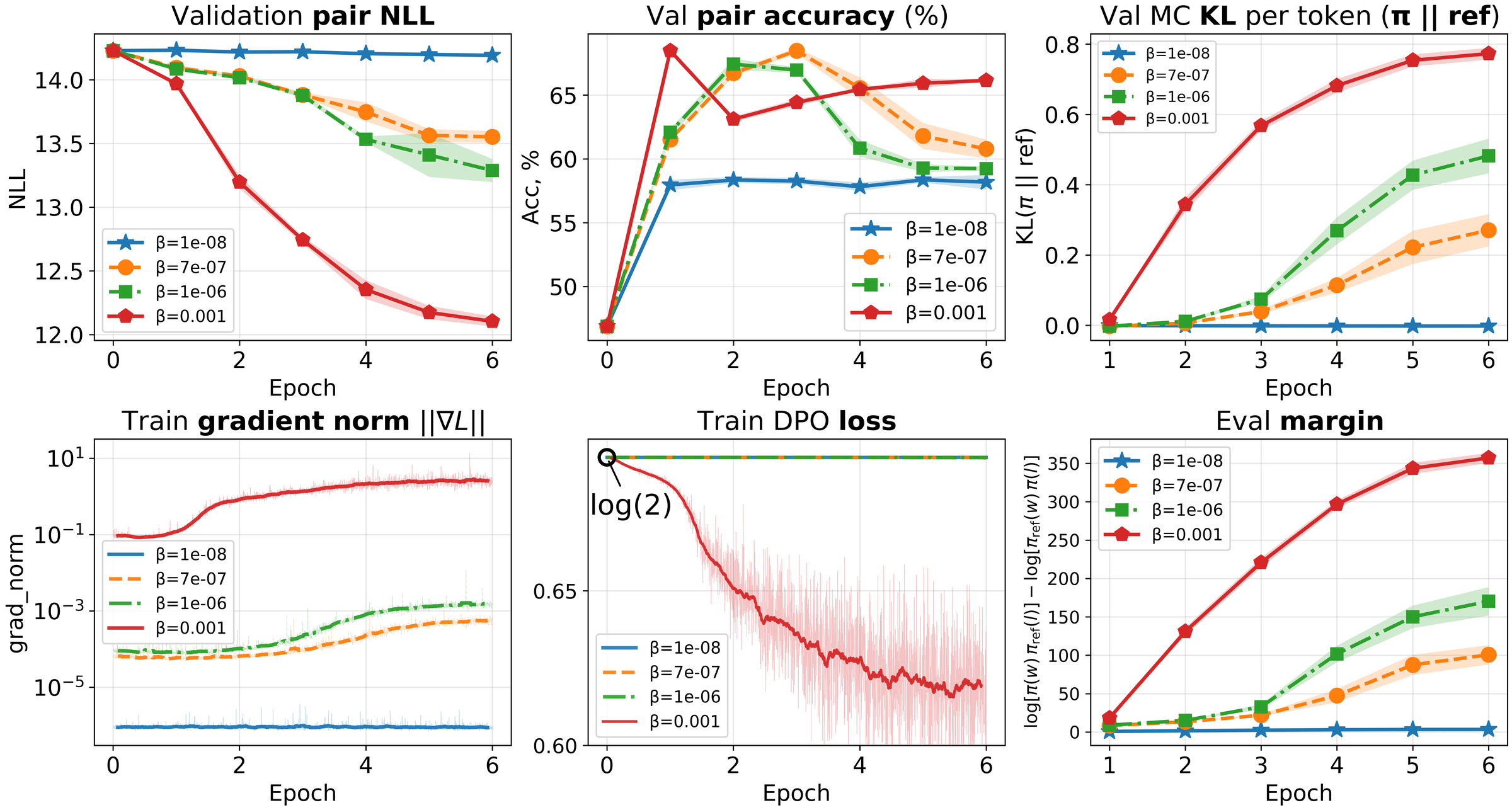}
  \caption{Training and validation dynamics of standard DPO on UltraFeedback with Qwen3-4B-Instruct-2507 using TRL \texttt{DPOTrainer} (AdamW, learning rate $5\times10^{-7}$, effective batch size 16), comparing $\beta\in\{10^{-8},7\times10^{-7},10^{-6},10^{-3}\}$. Curves show the mean over five seeds; shaded bands denote one standard deviation; see the discussion above.}
  \label{fig:intro-overview}
\end{figure}

\paragraph{UltraFeedback TRL baseline (Figure~\ref{fig:intro-overview}).}
To make the small-$\beta$ slowdown visible in a realistic setting, we train Qwen3-4B-Instruct-2507 on UltraFeedback for six epochs with the TRL \texttt{DPOTrainer}, comparing $\beta\in\{10^{-8},7\times10^{-7},10^{-6},10^{-3}\}$ (AdamW, learning rate $5\times10^{-7}$, effective batch size 16, five seeds per setting).
We report both training dynamics, including gradient norm, reward accuracy, and log-probability gap, and validation dynamics, including pair NLL, pair accuracy, and forward KL $\mathrm{KL}(\pi_{\theta}\,\|\,\pi_{\mathrm{ref}})$ per token.
The training-time metrics are taken from the TRL trainer logs.
Gradient norms rank-order with $\beta$: for $\beta\le 10^{-6}$ they remain low, the training loss stays pinned near $\log 2$, and validation KL stays close to zero, so these policies remain near the reference model, whereas the $\beta=10^{-3}$ run learns clearly.
Under the KL-control interpretation, the smaller-$\beta$ runs would instead be expected to deviate at least as much.
The pattern does not depend on the model or dataset: Figure~\ref{fig:hh-rlhf-mamba2-trl} below shows the same dynamics on PKU-processed HH-RLHF with Mamba-2.

\begin{figure}[H]
  \centering
  \includegraphics[width=0.98\linewidth]{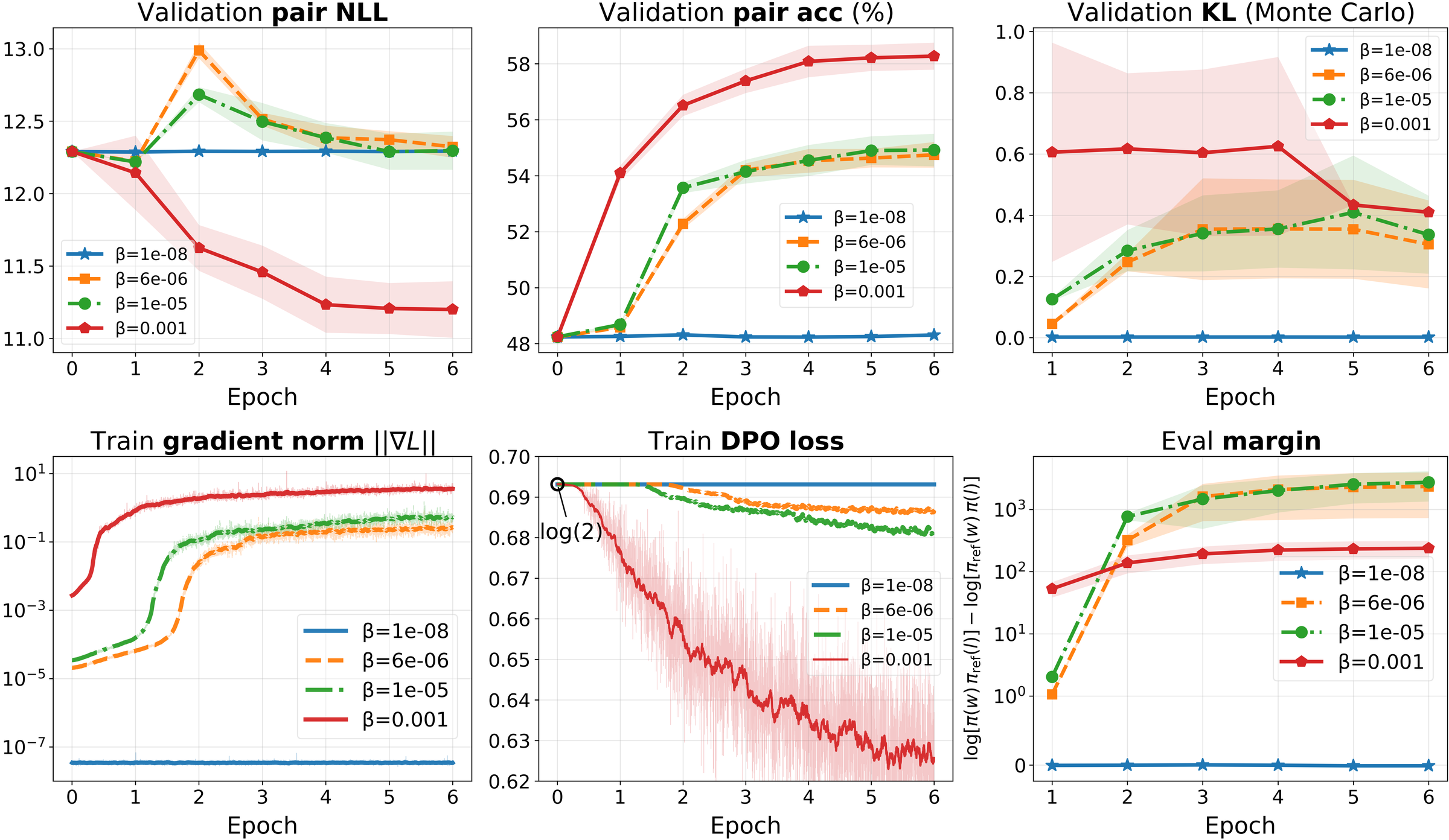}
  \caption{Training and validation dynamics of standard DPO on PKU-processed HH-RLHF with Mamba-2 using TRL \texttt{DPOTrainer} (AdamW, learning rate $5\times10^{-6}$, effective batch size 20). Curves show the mean over five seeds; shaded bands denote one standard deviation.}
  \label{fig:hh-rlhf-mamba2-trl}
\end{figure}

\begin{figure}[H]
  \centering
  \includegraphics[width=0.98\linewidth]{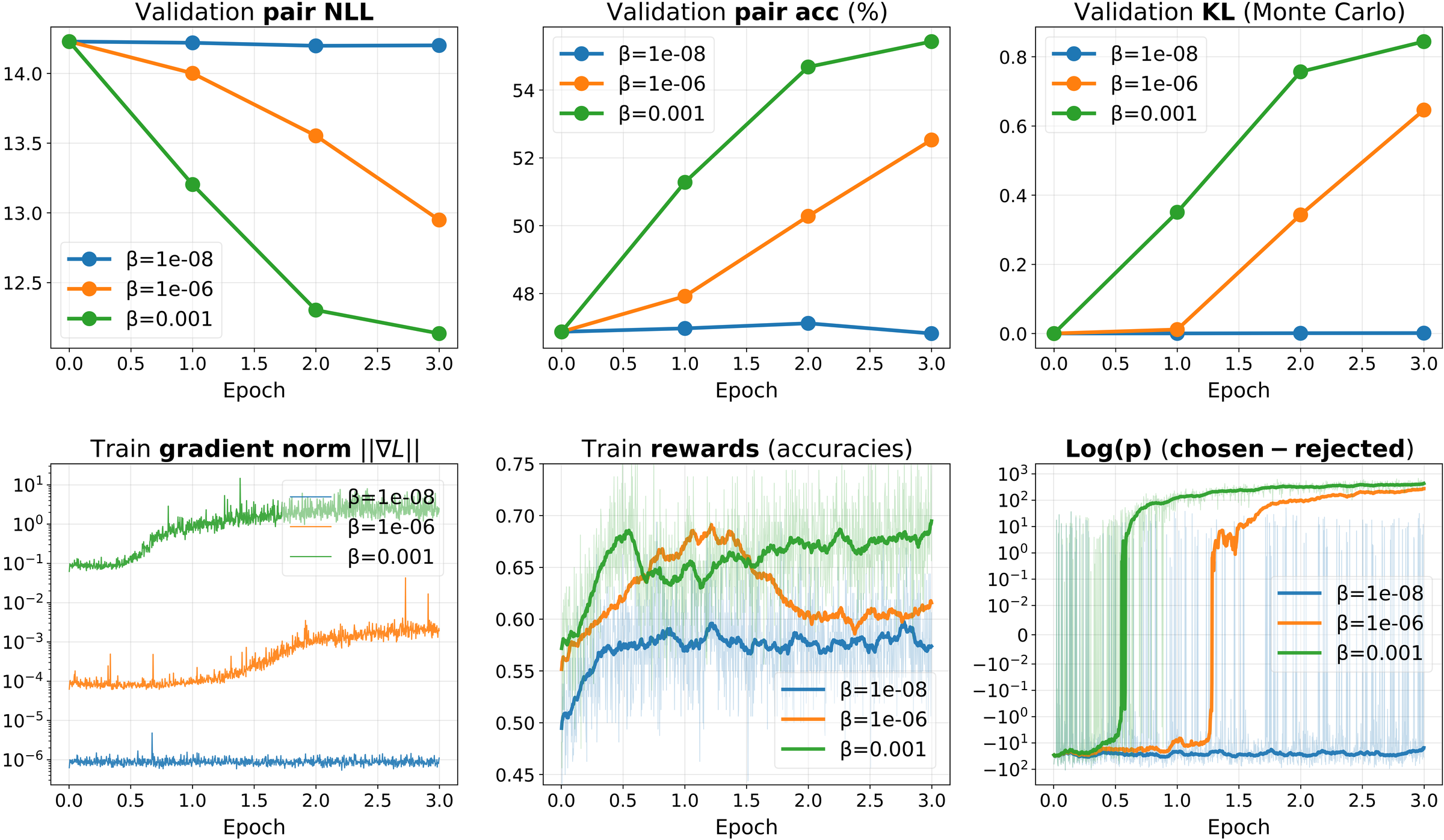}
  \caption{Training and validation dynamics of standard DPO on UltraFeedback with Qwen3-4B-Instruct-2507 using TRL \texttt{DPOTrainer} (AdamW, learning rate $1\times10^{-6}$, effective batch size 16), comparing $\beta\in\{10^{-8},10^{-6},10^{-3}\}$. For the two smaller $\beta$ values, gradient norms remain low and validation KL (per token) stays close to zero, so the policy remains near the reference model; only the $\beta=10^{-3}$ run moves appreciably.}
  \label{fig:ultrafb-classic-trl}
\end{figure}

\begin{figure}[H]
  \centering
  \includegraphics[width=0.92\linewidth]{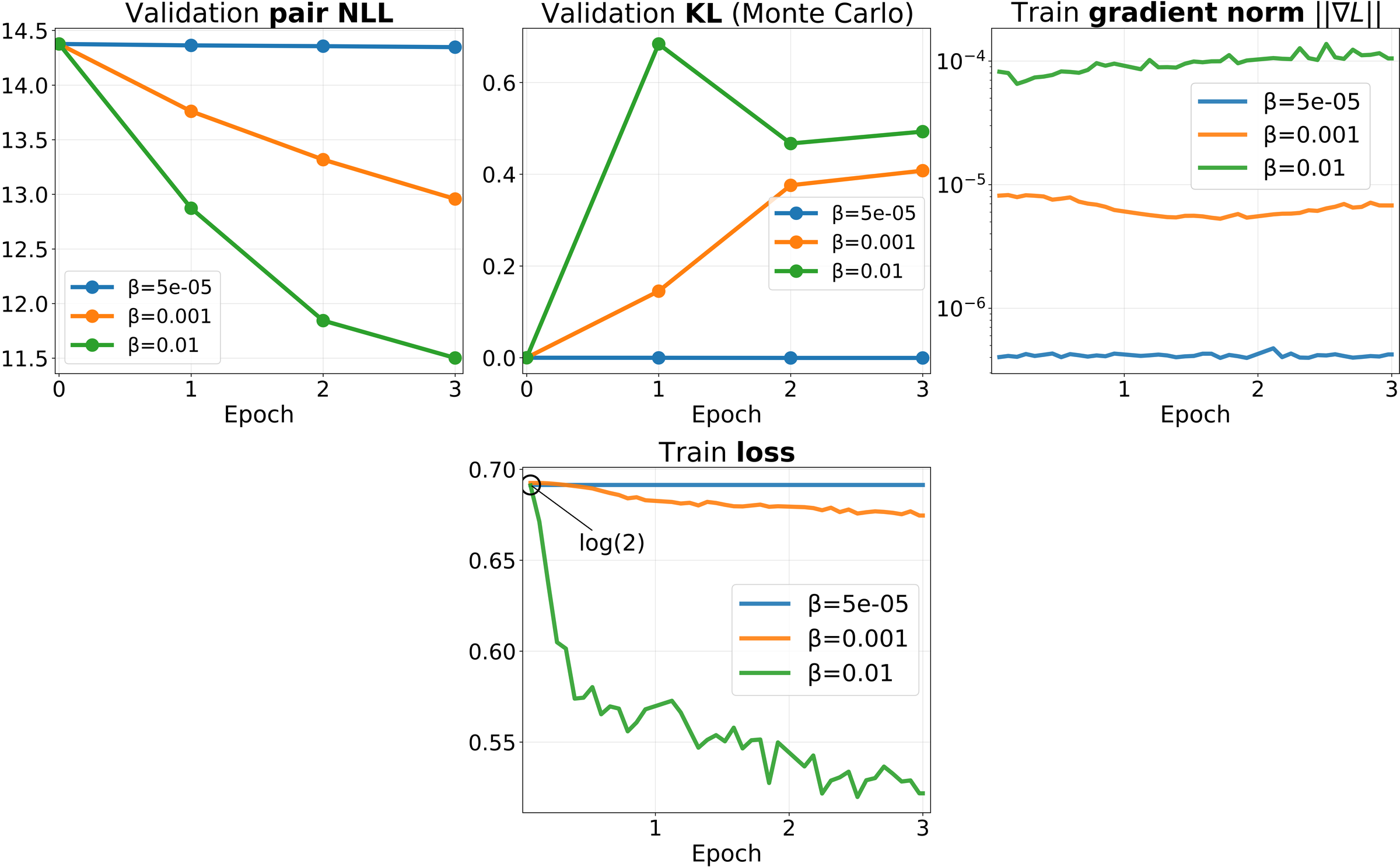}
  \caption{Training and validation dynamics of our DPO implementation on HelpSteer3 with Qwen3-4B-Instruct-2507 (SGD, $\mathrm{lr}=5\times10^{-4}$, batch size 24).}
  \label{fig:hsteer-sgd-lr5e4}
\end{figure}

Figure~\ref{fig:hsteer-sgd-lr5e4} shows an alternative HelpSteer3 SGD configuration at learning rate $5\times10^{-4}$, comparing $\beta\in\{5\times10^{-5}, 10^{-3}, 10^{-1}\}$.

\begin{figure}[H]
  \centering
  \includegraphics[width=0.92\linewidth]{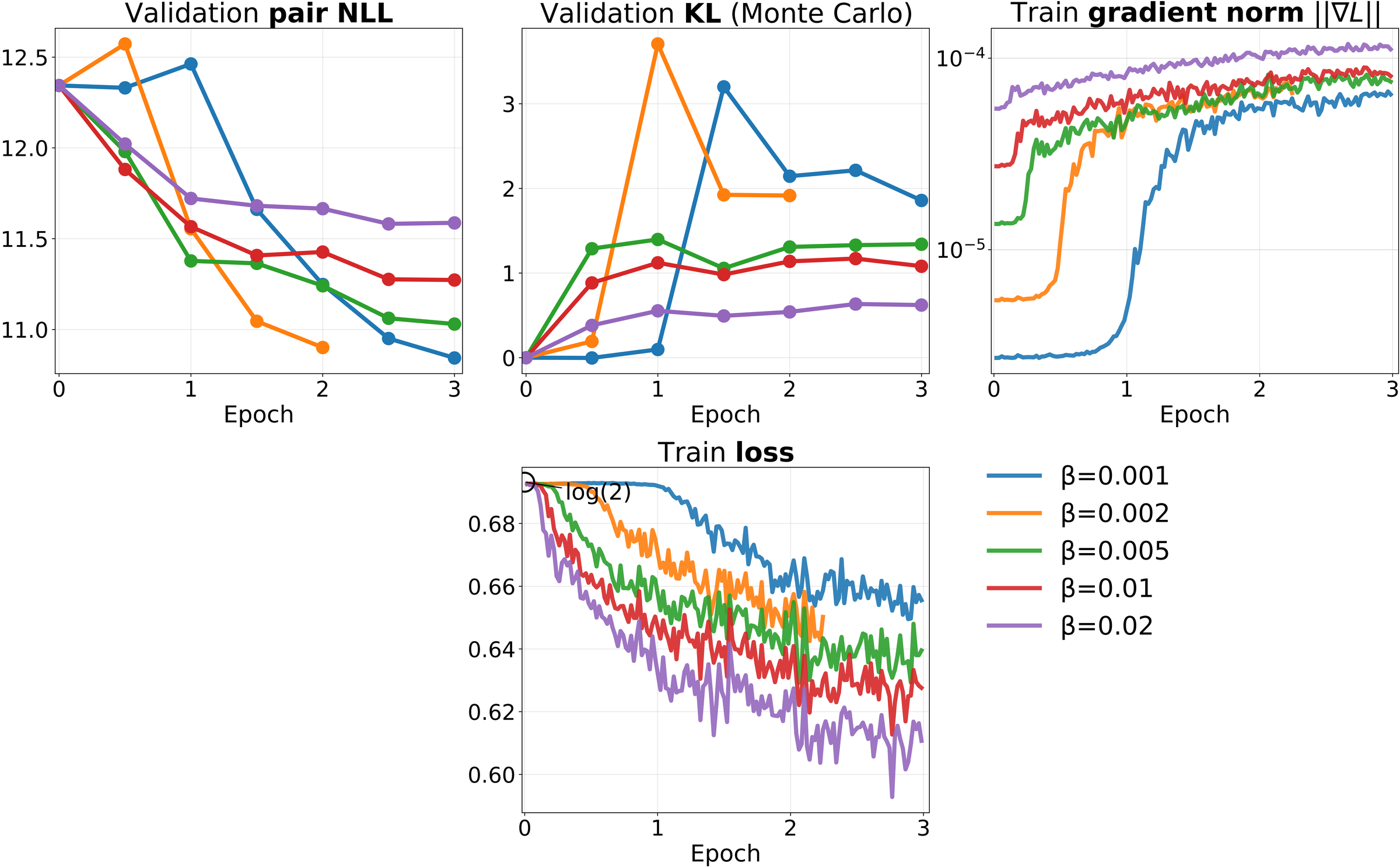}
  \caption{Training and validation dynamics of our DPO implementation on PKU-processed HH-RLHF with Qwen2.5-3B-Instruct (SGD, $\mathrm{lr}=10^{-3}$, batch size 32).}
  \label{fig:hh-rlhf-drift}
\end{figure}

Figure~\ref{fig:hh-rlhf-drift} shows a more intricate regime: across the HH-RLHF SGD trajectories ($\beta\in\{0.001,0.002,0.005,0.01,0.02\}$), a more aggressive policy trajectory can induce a delayed increase in gradient magnitude along the smaller-$\beta$ runs, partially compensating for the nominal linear $\beta$ prefactor in the DPO objective.

This delayed surge reflects the same stage-dependent growth of $G_t^{\mathrm{norm}}$ analyzed in Appendix~\ref{app:adam-threshold}: as the policy moves, $\sqrt{\hat{v}^h}$ increases and $\beta_{\mathrm{crit}}$ falls.
Under AdamW, that threshold drop would shift the optimizer from the $\epsilon$-dominated regime (in which updates are proportional to $\beta$ and learning is suppressed) toward the cancellation regime in which adaptive scaling no longer amplifies the gradient-scale coupling, so the effective $\beta$-induced slowdown is strongest at initialization and weakens once gradients have grown.
The HH-RLHF curves in Figure~\ref{fig:hh-rlhf-drift} use SGD rather than Adam, so there is no moment-based $\beta_{\mathrm{crit}}$ transition; nevertheless, the same rising Jacobian enlarges raw gradient norms and partially offsets the smaller $\beta$.
Under plain SGD with learning rate $\eta$, using the decomposition $g_t=\beta h_t$ from Appendix~\ref{app:adam-threshold},
\begin{equation}
\label{eq:sgd-dpo-update}
\theta_{t+1}-\theta_t = -\eta\,g_t,
\qquad
g_t = \beta\,h_t,
\qquad
\|g_t\|_2 = \beta\,\|h_t\|_2,
\end{equation}
where $h_t$ collects the sigmoid factor and the log-probability-gap Jacobian.
A smaller $\beta$ therefore directly shrinks $\|g_t\|_2$, but stage-dependent growth of $\|h_t\|_2$ as the policy moves can partially offset this reduction, as in the delayed surge visible in Figure~\ref{fig:hh-rlhf-drift}.

For sufficiently small $\beta$, however, this late-stage compensation remains weaker than the $\beta$-induced attenuation during the early phase, yielding dynamics that are noticeably non-monotone across practically used tuning ranges.
Such interaction effects between the inverse preference-noise scale and the optimization trajectory help explain why the phenomenon has often remained underappreciated in routine hyperparameter searches.
Consistent with this view, the $\beta=0.001$ trajectory attains substantially lower pair NLL and higher pair accuracy than the older $\beta=0.01$ run, while its temporal profile remains unusual and departs from the behavior one would expect from a simple $\beta$ rescaling.
The sharp late-stage surge in gradient norm and the qualitatively different curve shapes relative to Figures~\ref{fig:ultrafb-classic-trl} and~\ref{fig:cause-drift} are, among other factors, compounded by the comparatively large learning rate ($\mathrm{lr}=10^{-3}$, five times the HelpSteer3 SGD setting in Figure~\ref{fig:cause-drift}) acting on HH-RLHF preference pairs that are comparatively noisy and less separable.
Data-centric comparisons of public RLHF corpora quantify effective sample size, noise invariance, and information content and report lower informativeness for HH-RLHF relative to several alternatives \citep{Shen2024DataCentricRLHF}, while reward-modeling analyses of the same corpus emphasize annotator disagreement, ambiguous or incorrect preferences, and weak pairwise margins \citep{Wang2024SecretsRLHFPartII}.

\section{Thurstone Gradient and Comparison with DPO}
\label{app:thurstone-grad}

\setcounter{equation}{0}
\renewcommand{\theequation}{D.\arabic{equation}}
\renewcommand{\theHequation}{\theequation}

\paragraph{General link functions.}
Following the annotator-noise viewpoint of \citet{Sun2025RethinkingBT}, let $\varepsilon$ denote a pairwise comparison noise variable with CDF $F$ and PDF $f=F'$ on $\mathbb{R}$.
We assume the standard regularity conditions
\begin{equation}
F(-\infty)=0,\quad F(+\infty)=1,\quad f(x)\ge 0,\quad \int_{-\infty}^{\infty} f(x)\,dx=1,
\end{equation}
together with strict monotonicity ($f(x)>0$ for all $x$, hence $F(x)\in(0,1)$) and a finite first moment $\mathbb{E}[|\varepsilon|]<\infty$.
An annotator prefers $y_w$ over $y_l$ when the latent margin exceeds the noise, $\Delta>\varepsilon$, so
\begin{equation}
P(y_w \succ y_l \mid x)=P(\varepsilon<\Delta)=F(\Delta).
\end{equation}
Writing the effective inverse preference-noise scale in policy-margin coordinates as $\beta>0$ gives the scaled link $P(y_w \succ y_l \mid x)=F(\beta\Delta)$. As shown in Appendix~\ref{app:bt-temperature}, if the underlying reward model has pairwise noise scale $s_a$ and the RLHF objective has KL coefficient $\lambda$, then $\beta=\lambda/s_a$.
The corresponding negative log-likelihood is
\begin{equation}
\label{eq:general-link-loss}
\mathcal{L}_{F}(\Delta;\beta) = -\log F(\beta\Delta),
\end{equation}
and, by the chain rule,
\begin{equation}
\label{eq:general-link-grad}
\frac{\partial \mathcal{L}_{F}}{\partial \Delta}
= -\beta\,\frac{f(\beta\Delta)}{F(\beta\Delta)}.
\end{equation}
Thus the explicit multiplicative $\beta$ prefactor is not specific to the logistic or Gaussian branches: it follows for \emph{any} link CDF in this class.
Bradley--Terry/DPO and Thurstone correspond to $F=\sigma$ and $F=\Phi$; other examples on $\mathbb{R}$ with finite first moments include Laplace noise and symmetric Student-$t_\nu$ noise with $\nu>1$.
Distributions without finite first moments (e.g., Cauchy) are outside this interpretable class even though $P(\varepsilon<\Delta)$ is formally defined, because the latent-utility/noise-scale story no longer has finite expected bias.
Symmetry or membership in an exponential family is not required.
For plain SGD, decreasing $\beta$ therefore still linearly suppresses update magnitude across this class unless the learning rate is retuned.
Dividing by $\beta>0$ generalizes as well:
\begin{equation}
\frac{\partial}{\partial\Delta}\!\left[-\frac{1}{\beta}\log F(\beta\Delta)\right]
= -\frac{f(\beta\Delta)}{F(\beta\Delta)},
\end{equation}
with the same $\arg\min_{\Delta}\mathcal{L}_{F}$ because $1/\beta$ is a positive constant.

\paragraph{Thurstone case ($F=\Phi$).}
Beyond the logistic Bradley--Terry branch \citep{BradleyTerry1952}, the Gaussian annotator-noise model of \citet{Sun2025RethinkingBT} induces preferences $P(y_w\succ y_l)=\Phi(\beta\Delta)$, yielding the Thurstone pairwise loss
\begin{equation}
\label{eq:th-loss}
\mathcal{L}_{\mathrm{Th}}(\Delta;\beta)=-\log \Phi(\beta\Delta),
\end{equation}
the special case $F=\Phi$ of Equation~\eqref{eq:general-link-loss}, where $\Phi$ is the standard normal CDF and $\phi=\Phi'$ is the corresponding PDF.
Writing $u=\beta\Delta$, so that $\mathcal{L}_{\mathrm{Th}}(\Delta;\beta)=-\log\Phi(u)$ with $\partial u/\partial\Delta=\beta$, and using $\Phi'(u)=\phi(u)$, the chain rule gives
\begin{equation}
\label{eq:th-marginal-grad}
\frac{\partial \mathcal{L}_{\mathrm{Th}}}{\partial \Delta}
= -\frac{1}{\Phi(u)}\,\phi(u)\,\frac{\partial u}{\partial \Delta}
= -\beta\,\frac{\phi(\beta\Delta)}{\Phi(\beta\Delta)},
\end{equation}
which instantiates Equation~\eqref{eq:general-link-grad}.
As in the general case, dividing $\mathcal{L}_{\mathrm{Th}}$ by $\beta>0$ removes the front $\beta$ multiplier while preserving the argmin, yielding the normalized Thurstone marginal gradient $-\phi(\beta\Delta)/\Phi(\beta\Delta)$.

\paragraph{Comparison with the logistic branch.}
The Thurstone objective in Equations~\eqref{eq:th-loss}--\eqref{eq:th-marginal-grad} shares the same explicit $\beta$ prefactor as standard DPO (Eq.~\eqref{eq:dpo-marginal-grad}): only the shape factor changes from $\sigma(-\beta\Delta)$ to the Mills ratio $\phi(\beta\Delta)/\Phi(\beta\Delta)$.
At $\beta\Delta\approx 0$,
\begin{equation}
\sigma(0)=\tfrac12,
\qquad
\frac{\phi(0)}{\Phi(0)}=\sqrt{\frac{2}{\pi}},
\end{equation}
so the Gaussian branch is locally steeper by the constant factor
\begin{equation}
\frac{\phi(0)/\Phi(0)}{\sigma(0)}
=\frac{\sqrt{2/\pi}}{1/2}
=\sqrt{\frac{8}{\pi}}
\approx 1.60.
\end{equation}
For large negative margins, $\phi(x)/\Phi(x)\sim|x|$ grows rather than saturating, so the Thurstone loss increases quadratically, $-\log\Phi(\beta\Delta)\sim(\beta\Delta)^2/2$ as $\beta\Delta\to-\infty$, and penalizes strongly misranked pairs more aggressively than the logistic branch, whose loss grows only linearly, $-\log\sigma(\beta\Delta)\sim-\beta\Delta$, with marginal factor $\sigma(-\beta\Delta)$ bounded above by $1$.

\section{Formal Proof of Proposition~1 and Relation to Prior Work}
\label{app:prop1-proof}

\setcounter{equation}{0}
\renewcommand{\theequation}{E.\arabic{equation}}
\renewcommand{\theHequation}{\theequation}

\paragraph{Novelty statement.}
The centered-softplus function $s_c$ was introduced by \citet{Srinivas2022LCNN} as an
activation for lightweight convex neural networks.
We distinguish our contribution: we \emph{diagnose} the $\beta$-entanglement problem in DPO
(Section~\ref{sec:beta-weakens-updates}--\ref{sec:cause}), show that applying $s_c$ as a
\emph{loss normalization} resolves it while preserving the DPO minimizer
(Proposition~1 below),
extend the resulting objective to $\beta=0$ via a limiting argument (Proposition~2 below),
and provide a theoretical analysis of the Adam compensation regime
(Appendix~\ref{app:adam-threshold}).
These results are not reported in \citet{Srinivas2022LCNN} or in prior DPO work.

\paragraph{Proposition 1 (Argmin equivalence for $\beta>0$).}\label{prop:argmin}
For $\beta>0$,
\begin{equation}
\mathcal{L}_{\mathrm{DPO}}^{\mathrm{norm}}(\Delta;\beta)
=\frac{1}{\beta}\,\mathcal{L}_{\mathrm{DPO}}(\Delta;\beta)-\frac{\ln 2}{\beta},
\end{equation}
where $\mathcal{L}_{\mathrm{DPO}}(\Delta;\beta)=-\log \sigma(\beta \Delta)$ and $\mathcal{L}_{\mathrm{DPO}}^{\mathrm{norm}}(\Delta;\beta)=s_c(-\Delta;\beta)$.
Hence,
\begin{equation}
\arg\min_{\Delta}\mathcal{L}_{\mathrm{DPO}}^{\mathrm{norm}}(\Delta;\beta)
=\arg\min_{\Delta}\mathcal{L}_{\mathrm{DPO}}(\Delta;\beta).
\end{equation}
Equivalently, subtracting the additive constant $\ln 2/\beta$ from the normalized softplus loss does not change the minimizer:
\begin{equation}
\arg\min_{\Delta} s(-\Delta;\beta)=\arg\min_{\Delta}\bigl[s(-\Delta;\beta)-\ln 2/\beta\bigr].
\end{equation}

\paragraph{Proof of Proposition~1 (algebraic identity).}
We verify the identity
$\mathcal{L}_{\mathrm{DPO}}^{\mathrm{norm}}(\Delta;\beta)
= \tfrac{1}{\beta}\mathcal{L}_{\mathrm{DPO}}(\Delta;\beta) - \tfrac{\ln 2}{\beta}$
directly.
Recall $\mathcal{L}_{\mathrm{DPO}}(\Delta;\beta)=-\log\sigma(\beta\Delta)=\log(1+e^{-\beta\Delta})$
and $\mathcal{L}_{\mathrm{DPO}}^{\mathrm{norm}}(\Delta;\beta)=s_c(-\Delta;\beta)$.
By the definition of $s_c$,
\begin{align}
s_c(-\Delta;\beta)
&= \frac{1}{\beta}\log\frac{1+e^{-\beta\Delta}}{2}
\nonumber\\
&= \frac{1}{\beta}\mathcal{L}_{\mathrm{DPO}}(\Delta;\beta)-\frac{\ln 2}{\beta},
\label{eq:appd-algebraic-identity}
\end{align}
which is the required identity.\hfill$\square$

\paragraph{Proof of Proposition~1 (argmin equivalence over $\theta$).}
Let $\Theta$ be the parameter space and write
\begin{equation}
    F(\theta)=\mathbb{E}_{(x,y_w,y_l)\sim\mathcal{D}}\bigl[
        \mathcal{L}_{\mathrm{DPO}}(\Delta_\theta;\beta)\bigr],
    \quad
    G(\theta)=\mathbb{E}_{(x,y_w,y_l)\sim\mathcal{D}}\bigl[
        \mathcal{L}_{\mathrm{DPO}}^{\mathrm{norm}}(\Delta_\theta;\beta)\bigr],
\label{eq:appd-objectives}
\end{equation}
where $\Delta_\theta(x,y_w,y_l)$ is the log-ratio gap induced by $\theta$.
From the algebraic identity in Equation~\eqref{eq:appd-algebraic-identity},
\begin{equation}
    G(\theta)=\frac{1}{\beta}F(\theta)-\frac{\ln 2}{\beta}
    \qquad\forall\,\theta\in\Theta.
\label{eq:appd-affine-map}
\end{equation}
For any $\theta_1,\theta_2\in\Theta$,
\begin{equation}
    G(\theta_1)\le G(\theta_2)
    \;\iff\;
    \frac{1}{\beta}F(\theta_1)-\frac{\ln 2}{\beta}
    \le
    \frac{1}{\beta}F(\theta_2)-\frac{\ln 2}{\beta}
    \;\iff\;
    F(\theta_1)\le F(\theta_2),
\label{eq:appd-order-equivalence}
\end{equation}
where the last equivalence uses $1/\beta>0$.
Hence the two objectives induce the same total order on $\Theta$, and in particular
$\arg\min_\theta G(\theta)=\arg\min_\theta F(\theta)$.\hfill$\square$

\paragraph{Continuity at $\beta=0$ (Proposition~2).}
\label{prop:continuity}
For fixed $x\in\mathbb{R}$, define
\begin{equation}
    h_x(\beta) := \log\!\left(\frac{1+e^{\beta x}}{2}\right),\qquad \beta\ge 0.
\label{eq:appd-hx-definition}
\end{equation}
Then for $\beta>0$,
\begin{equation}
    s_c(x;\beta)=\frac{h_x(\beta)}{\beta}.
\label{eq:appd-sc-hx}
\end{equation}
Applying this identity at $x=-\Delta$ gives
\begin{equation}
    s_c(-\Delta;\beta)
    = \frac{h_{-\Delta}(\beta)}{\beta}.
\label{eq:appd-sc-minus-delta}
\end{equation}
Since $h_{-\Delta}(0)=0$, this is the right difference quotient of $h_{-\Delta}$ at $0$.
Because $h_{-\Delta}$ is differentiable at $\beta=0$,
\begin{equation}
    \lim_{\beta\to 0^+} s_c(-\Delta;\beta) = h_{-\Delta}'(0).
\label{eq:appd-right-derivative}
\end{equation}
Moreover,
\begin{equation}
    h_{-\Delta}'(\beta)
    = \frac{(-\Delta)e^{-\beta\Delta}}{1+e^{-\beta\Delta}}
    = -\frac{\Delta}{1+e^{\beta\Delta}},
\label{eq:appd-hx-derivative}
\end{equation}
so $h_{-\Delta}'(0)=-\Delta/2$. Therefore
\begin{equation}
    \lim_{\beta\to 0^+} s_c(-\Delta;\beta)=-\frac{\Delta}{2},
\label{eq:appd-sc-limit}
\end{equation}
which matches the piecewise definition at $\beta=0$ in
Equation~\eqref{eq:dpo-norm-piecewise}.\hfill$\square$

\paragraph{Gradient at $\beta=0$.}
For $\beta>0$,
$\partial s_c(-\Delta;\beta)/\partial\Delta=-\sigma(-\beta\Delta)$.
At $\beta=0$, from $s_c(-\Delta;0)=-\Delta/2$, we have
$\partial s_c(-\Delta;0)/\partial\Delta=-1/2$.
To verify continuity at the boundary, use that $\sigma$ is globally Lipschitz with
constant $1/4$:
\begin{equation}
    |\sigma(u)-\sigma(v)| \le \tfrac{1}{4}|u-v|.
\label{eq:appd-sigma-lipschitz}
\end{equation}
By direct substitution,
\begin{equation}
    \left|
    \frac{\partial s_c(-\Delta;\beta)}{\partial\Delta}
    -
    \frac{\partial s_c(-\Delta;0)}{\partial\Delta}
    \right|
    = \left|-\sigma(-\beta\Delta)+\tfrac{1}{2}\right|
    = |\sigma(-\beta\Delta)-\sigma(0)|.
\label{eq:appd-gradient-diff}
\end{equation}
Applying Equation~\eqref{eq:appd-sigma-lipschitz} to
Equation~\eqref{eq:appd-gradient-diff} with $u=-\beta\Delta$ and $v=0$ yields
\begin{equation}
    |\sigma(-\beta\Delta)-\sigma(0)|
    \le \tfrac{1}{4}|\beta\Delta|.
\label{eq:appd-gradient-bound}
\end{equation}
As $(\Delta,\beta)\to(\Delta_0,0^+)$, $\Delta$ remains bounded in a neighborhood of
$\Delta_0$, so the right-hand side tends to $0$.
Therefore $\partial s_c(-\Delta;\beta)/\partial\Delta$ is continuous at $\beta=0$
(from the admissible side $\beta\ge 0$), confirming that no special optimizer
treatment is required at the boundary.

\paragraph{Hard-margin limit of normalized softplus.}
At the opposite scale, the same $1/\beta$ normalization has the standard hard-margin limit of softplus, which matches the soft-margin-ranking interpretation of DPO \citep[Prop.~4.1]{Yang2026ConditionalDPO}:
\begin{equation}
\label{eq:softplus-hard-margin-limit}
\lim_{\beta\to\infty}\log\!\left(1+e^{-\beta\Delta}\right)/\beta
=\max\{-\Delta,0\}.
\end{equation}
Thus $\beta$ controls the sharpness of a soft hinge: finite $\beta$ gives a smooth transition, while $\beta\to\infty$ recovers the hard hinge boundary, consistent with the margin-ranking interpretation of DPO \citep{Yang2026ConditionalDPO}.
Centering subtracts the constant $\ln 2/\beta$ and therefore changes only the vertical offset: $s_c(0;\beta)=0$ exactly, so the decision boundary remains at $\Delta=0$.
As $\beta\to\infty$, $s_c(-\Delta;\beta)$ likewise converges to $\max\{-\Delta,0\}$.

\section{Bradley--Terry Scale and Log-Likelihood-Ratio View}
\label{app:bt-temperature}

\setcounter{equation}{0}
\renewcommand{\theequation}{F.\arabic{equation}}
\renewcommand{\theHequation}{\theequation}

\paragraph{Annotator-noise foundations.}
Following \citet{Sun2025RethinkingBT}, let $\Delta_r = r_{x,y_1}-r_{x,y_2}$ denote the latent utility gap and $\Delta_b = b(x,y_2,A)-b(x,y_1,A)$ the pairwise annotator-bias difference.
Then
\begin{equation}
P(y_1 \succ y_2 \mid x) = P(\Delta_r > \Delta_b).
\label{eq:bt-comparison}
\end{equation}

\emph{Logistic difference (Assumption~3).}
If the two annotator-bias terms are independently Gumbel-distributed with common scale $s_a$,
$b(x,y_i,A)\overset{\mathrm{i.i.d.}}{\sim}\mathrm{Gumbel}(0,s_a)$, then their difference is
$\Delta_b=b(x,y_2,A)-b(x,y_1,A)\sim\mathrm{Logistic}(0,s_a)$:
\begin{equation}
P(y_1 \succ y_2 \mid x)
= \sigma\!\left(\frac{\Delta_r}{s_a}\right).
\label{eq:bt-logistic}
\end{equation}

\emph{Gaussian difference (Assumption~4).}
If the two annotator-bias terms are independently Gaussian with common variance $s_a^2/2$,
$b(x,y_i,A)\overset{\mathrm{i.i.d.}}{\sim}\mathcal{N}(0,s_a^2/2)$, then their difference is
$\Delta_b=b(x,y_2,A)-b(x,y_1,A)\sim \mathcal{N}(0,s_a^2)$:
\begin{equation}
P(y_1 \succ y_2 \mid x)
= \Phi\!\left(\frac{\Delta_r}{s_a}\right),
\label{eq:bt-gaussian}
\end{equation}
where $\Phi$ is the standard normal CDF.
This parameterization absorbs the $\sqrt{2}$ factor from subtracting two independent Gaussian biases into the per-item variance, so $s_a$ denotes the standard deviation of the pairwise bias difference, matching the role of $s_a$ in the logistic branch.
This matches the Thurstone case in Remark~4 of \citet{Sun2025RethinkingBT} after this rescaling of the per-item variance.
At $s_a=1$, Eqs.~\eqref{eq:bt-logistic}--\eqref{eq:bt-gaussian} reproduce the unit-scale logistic and Gaussian curves compared in \citet{Sun2025RethinkingBT}; there $\Delta_b$ is modeled directly as $\mathrm{Logistic}(0,1)$ or $\mathcal{N}(0,1)$.

For an observed preference $y_w \succ y_l$ with $\Delta_r = r_w-r_l$, the corresponding pairwise reward-modeling losses are
\begin{equation}
\mathcal{L}_{\mathrm{BT}}
= -\log \sigma\!\left(\frac{\Delta_r}{s_a}\right),
\qquad
\mathcal{L}_{\mathrm{Th}}
= -\log \Phi\!\left(\frac{\Delta_r}{s_a}\right).
\label{eq:bt-rm-losses}
\end{equation}
The shared denominator in Eq.~\eqref{eq:bt-rm-losses} reflects the same pairwise-difference construction: $s_a$ is the scale of the induced pairwise bias difference in both the logistic and Gaussian cases. It is distinct from the KL coefficient $\lambda$ in Equation~\eqref{eq:rlhf}.

\paragraph{Connection to DPO.}
The optimal policy of the KL-regularized RLHF objective~\eqref{eq:rlhf} has the closed form
\begin{equation}
\label{eq:rlhf-optimal-policy}
\pi^{*}(y\mid x)=\frac{1}{Z(x)}\,\pi_{\mathrm{ref}}(y\mid x)\exp\!\left(\frac{r^{*}(x,y)}{\lambda}\right),
\end{equation}
which can be inverted to express the reward through the policy,
$r^*(x,y) = \lambda\log(\pi^*(y\mid x)/\pi_{\mathrm{ref}}(y\mid x)) + \lambda\log Z(x)$.
Substituting this implicit reward into the Bradley--Terry likelihood~\eqref{eq:bt-logistic} cancels the partition function $Z(x)$ and gives
\begin{equation}
\begin{aligned}
P(y_w \succ y_l \mid x)
&= \sigma\!\left(\frac{\lambda}{s_a}\Delta\right)
= \sigma(\beta\Delta),
\\
\Delta &= \log\frac{\pi(y_w\mid x)}{\pi_{\mathrm{ref}}(y_w\mid x)}-\log\frac{\pi(y_l\mid x)}{\pi_{\mathrm{ref}}(y_l\mid x)}.
\end{aligned}
\label{eq:dpo-bt}
\end{equation}
Here
\begin{equation}
\label{eq:dpo-effective-beta}
\beta := \frac{\lambda}{s_a}
\end{equation}
is the effective inverse preference-noise scale in policy-log-ratio coordinates. The DPO likelihood identifies this ratio, not $\lambda$ and $s_a$ separately; interpreting either factor physically requires a fixed reward normalization.
The DPO loss is the negative log-likelihood of this model over observed pairs:
$\mathcal{L}_{\mathrm{DPO}} = -\log\sigma(\beta\Delta)$.
At fixed $\lambda$, smaller $s_a$ (more decisive annotations) increases $\beta$ and sharpens the sigmoid, whereas larger $s_a$ decreases $\beta$ and yields softer preferences. At fixed $s_a$, changing $\lambda$ changes the same effective coefficient through the reward--policy conversion.

\paragraph{Connection to a log-likelihood-ratio baseline.}
Each training example presents an observation $O$: ``annotator chose $y_w$ over $y_l$''.
We compare two models of the world:
\begin{itemize}
  \item $H_0$ (null): both completions equally good, $r(y_w)=r(y_l)$,
        so $P(O\mid H_0)=\sigma(0)=\tfrac{1}{2}$.
  \item $H_1$ (current policy): the policy assigns log-ratio gap $\Delta$,
        so $P(O\mid H_1)=\sigma(\beta\Delta)$.
\end{itemize}
The per-example likelihood ratio against this null baseline is
\begin{equation}
\Lambda
= \frac{P(O\mid H_1)}{P(O\mid H_0)}
= \frac{\sigma(\beta\Delta)}{1/2}
= 2\,\sigma(\beta\Delta).
\label{eq:lr-stat}
\end{equation}
Taking logarithms and recalling $\mathcal{L}_{\mathrm{DPO}}=-\log\sigma(\beta\Delta)$,
\begin{equation}
\log\Lambda = \log 2 - \mathcal{L}_{\mathrm{DPO}}.
\label{eq:log-lr-dpo}
\end{equation}
Hence \textbf{minimizing the DPO loss is equivalent to maximizing $\log\Lambda$}:
the policy is trained to make each observed preference as much more likely
than the ``equal-quality'' baseline as possible.
Subtracting $\log 2$ removes the random-choice baseline, and
$\mathcal{L}_{\mathrm{DPO}}-\log 2=-\log\Lambda$ is the negative log-likelihood ratio relative to chance.
Dividing by the effective coefficient gives
$(\mathcal{L}_{\mathrm{DPO}}-\log 2)/\beta=(s_a/\lambda)(\mathcal{L}_{\mathrm{DPO}}-\log 2)$.
This normalization removes the direct $\beta$ prefactor from the marginal gradient. It equals a conversion to annotator-noise units only under an explicitly fixed reward/KL normalization.

Note also that $\Delta$ itself is a log-likelihood ratio comparing the current policy to the reference:
\begin{equation}
\Delta
= \underbrace{\log\frac{\pi(y_w\mid x)}{\pi_{\mathrm{ref}}(y_w\mid x)}}_{\text{log-LR for }y_w}
- \underbrace{\log\frac{\pi(y_l\mid x)}{\pi_{\mathrm{ref}}(y_l\mid x)}}_{\text{log-LR for }y_l}.
\label{eq:delta-log-lr}
\end{equation}
Thus $\beta\Delta=(\lambda/s_a)\Delta$ is a scale-controlled evidence score: how strongly the current policy
departs from the reference in the direction of the observed preference.
In ML terms, $\sigma(\beta\Delta)$ is a soft preference score. At fixed $\lambda$, varying the annotator-noise scale gives
\begin{equation}
\begin{aligned}
s_a\to 0\;(\beta\to\infty):\quad &\sigma(\beta\Delta)\to\mathbf{1}[\Delta>0]
\quad\text{(hard preference boundary, near-deterministic labels)},\\
s_a\to\infty\;(\beta\to0):\quad &\sigma(\beta\Delta)\to\tfrac{1}{2}
\quad\text{(maximally noisy labels, no discrimination)}.
\end{aligned}
\label{eq:bt-limits}
\end{equation}
Because only $\beta=\lambda/s_a$ appears in the DPO likelihood, a sweep over $\beta$ varies an effective inverse scale; it cannot by itself distinguish a change in annotator noise from a change in reward/KL normalization.

\paragraph{Implication for the disentanglement.}
The two roles of the effective coefficient $\beta=\lambda/s_a$ identified in the main paper are:
\begin{itemize}
  \item \emph{Statistical inverse scale $\beta=\lambda/s_a$}:
        the ratio of the KL/reward scale to the annotator-noise scale sets the width of the soft decision boundary in policy-log-ratio coordinates.
  \item \emph{Optimization scale $\beta$} (gradient multiplier):
        an unintended coupling to effective step size that arises from the standard DPO parameterization.
\end{itemize}
The centered-softplus reformulation removes the second role while preserving the first.
It lets the practitioner tune the effective statistical coefficient $\beta$ and the optimizer independently. Separating $\lambda$ from $s_a$ additionally clarifies the modeling interpretation, but the two quantities require an external normalization or additional information to be identified individually.

\section{Saturation Thresholds: Where Training Stops as a Function of \texorpdfstring{$\beta$}{beta}}
\label{app:saturation}

\setcounter{equation}{0}
\renewcommand{\theequation}{G.\arabic{equation}}
\renewcommand{\theHequation}{\theequation}

\paragraph{Setup.}
We model effective convergence with a stopping tolerance: updates on a preference pair become negligible once the magnitude of the marginal gradient drops below a threshold $\tau>0$, which lumps together the SGD noise floor, learning-rate decay, and the finite training horizon.
The invariant quantity behind this tolerance is the per-step drift of the margin.
A parameter step $\theta\leftarrow\theta-\eta\nabla_{\theta}\mathcal{L}$ with $\nabla_{\theta}\mathcal{L}=\ell'(\Delta)\,\nabla_{\theta}\Delta$ changes the scalar margin by $\Delta\leftarrow\Delta+\eta\,g^{2}\,|\ell'(\Delta)|$ to first order, where $g=\|\nabla_{\theta}\Delta\|_{2}$: the factor $g^{2}$ is the squared Jacobian of the margin, not a square of the marginal loss derivative (Appendix~\ref{app:right-dead-zone}, Equation~\eqref{eq:rdz-disc}).
A run stalls once this drift falls below a noise-floor scale $\varepsilon>0$ of the stochastic margin increments (Appendix~\ref{app:margin-diffusion}),
\[
\eta\,g^{2}\,\beta\,\sigma(-\beta\Delta)\;\lesssim\;\varepsilon .
\]
At a fixed learning rate---the setting of this appendix---this criterion is equivalent to the marginal-gradient threshold above with $\tau=\varepsilon/(\eta g^{2})$; under the learning-rate rescaling $\eta\propto 1/\beta$ of Appendix~\ref{app:loss-invariance}, the same criterion instead reduces to a condition on $\sigma(-\beta\Delta)$ alone (used there).
This is a coarse description of where training stops, not an exact optimization statement; it turns the claim that $\beta$ controls saturation into scaling predictions that can be checked against the reported runs.

\paragraph{Normalized objective: a clean $1/\beta$ law.}
For the centered-softplus objective, $|\partial\mathcal{L}_{\mathrm{DPO}}^{\mathrm{norm}}/\partial\Delta|=\sigma(-\beta\Delta)$, so the stopping condition $\sigma(-\beta\Delta)=\tau$ gives the saturation margin in Equation~\eqref{eq:sat-margin-norm},
\begin{equation}
\Delta^{*}=\frac{1}{\beta}\log\frac{1-\tau}{\tau}.
\end{equation}
The tolerance enters only through the numerator, so for any fixed $\tau<1/2$ the product $\beta\,\Delta^{*}$ is a constant independent of $\beta$.
Unlike Equation~\eqref{eq:sat-margin-dpo}, there is no dead zone and no intermediate peak: $\Delta^{*}$ is strictly monotone in $1/\beta$.

\paragraph{Standard DPO: dead zone, drifting product, and a peak.}
Section~\ref{sec:loss-geometry} states the main result: for the classical loss, $|\partial\mathcal{L}_{\mathrm{DPO}}/\partial\Delta|=\beta\,\sigma(-\beta\Delta)$, and the stopping condition $\beta\,\sigma(-\beta\Delta)=\tau$ yields Equation~\eqref{eq:sat-margin-dpo}.
We reproduce the derivation here.
Setting $\beta\,\sigma(-\beta\Delta)=\tau$ and solving for $\Delta$ gives
\begin{equation}
\Delta^{*}_{\mathrm{DPO}}=\frac{1}{\beta}\log\!\left(\frac{\beta}{\tau}-1\right),
\qquad \beta>2\tau.
\end{equation}
This scalar is an idealized one-pair saturation ceiling under the stopping model---not a train-set mean margin, and not a quantity recovered by re-evaluating frozen weights on the training pairs.
Empirical $\Delta$ statistics remain below the ceiling in general and relate to $\Delta^{*}_{\mathrm{DPO}}(\beta)$ only through their $\beta$-shape (Section~\ref{app:tau-estimators}).
Three regimes follow (as summarized in Section~\ref{sec:loss-geometry}).
\begin{itemize}
\item \emph{Dead zone} ($\beta\le 2\tau$): the marginal gradient is largest at initialization, where $\Delta=0$ and $|\partial\mathcal{L}_{\mathrm{DPO}}/\partial\Delta|=\beta/2$. If $\beta/2\le\tau$, the gradient is below tolerance from the start and training never escapes initialization: the run is suppressed outright rather than merely slowed. The normalized objective has no dead zone: its initial marginal gradient is $1/2$ for every $\beta$.
\item \emph{No clean scaling} ($\beta>2\tau$): the product $\beta\,\Delta^{*}_{\mathrm{DPO}}=\log(\beta/\tau-1)$ grows logarithmically in $\beta$, in contrast to the constant product in Equation~\eqref{eq:sat-margin-norm}.
\item \emph{Non-monotone peak and right-hand saturation}: maximizing Equation~\eqref{eq:sat-margin-dpo} over $\beta$ gives Equation~\eqref{eq:sat-peak}; the detailed derivation is given below.
The threshold model therefore predicts this non-monotone shape for the saturation margin. In the HelpSteer3 SGD sweep of Figure~\ref{fig:beta-peak-nll-kl-delta}, per-token KL co-moves with the mean margin over a wide $\beta$ range (a left dead zone, an intermediate peak, and a decaying right slope), which we treat as an empirical observation rather than as a general mapping from $\Delta$ to KL.
\end{itemize}

\paragraph{Derivation of the peak location.}
Differentiating Equation~\eqref{eq:sat-margin-dpo} with respect to $\beta$,
\begin{equation}
\frac{\mathrm{d}\Delta^{*}_{\mathrm{DPO}}}{\mathrm{d}\beta}
=\frac{1}{\beta^{2}}\left[\frac{\beta/\tau}{\beta/\tau-1}-\log\!\left(\frac{\beta}{\tau}-1\right)\right],
\end{equation}
so the stationarity condition is
\begin{equation}
\frac{\beta/\tau}{\beta/\tau-1}=\log\!\left(\frac{\beta}{\tau}-1\right).
\end{equation}
Substituting $v=\beta/\tau-1$ turns the left-hand side into $(v+1)/v=1+1/v$, giving the transcendental equation
\begin{equation}
\label{eq:sat-peak-transcendental}
1+\frac{1}{v}=\log v .
\end{equation}
The left-hand side decreases from $+\infty$ to $1$ and the right-hand side increases from $-\infty$, so Equation~\eqref{eq:sat-peak-transcendental} has a unique root; numerically $v^{*}\approx 3.591$ (at $v=3.591$: $\log v\approx 1.2784$ and $1+1/v\approx 1.2785$).
Hence
\begin{equation}
\label{eq:app-g6-beta-peak}
\beta_{\mathrm{peak}}=\tau\,(1+v^{*})\approx 4.59\,\tau\approx 4.6\,\tau,
\end{equation}
which is Equation~\eqref{eq:sat-peak}.
The second derivative is negative at $v^{*}$ (the bracket above changes sign from positive to negative), so this stationary point is the unique maximum.
The peak value itself has a compact form: using $\log v^{*}=1+1/v^{*}$,
\begin{equation}
\label{eq:app-g7-delta-peak}
\Delta^{*}_{\mathrm{DPO}}(\beta_{\mathrm{peak}})
=\frac{\log v^{*}}{\tau(1+v^{*})}
=\frac{1+1/v^{*}}{\tau(1+v^{*})}
=\frac{1}{\tau\,v^{*}}
\approx\frac{0.278}{\tau},
\end{equation}
so the maximal reachable margin under standard DPO at fixed learning rate is set entirely by the tolerance $\tau$.
For $\beta\gg\beta_{\mathrm{peak}}$, $\Delta^{*}_{\mathrm{DPO}}\approx\log(\beta/\tau)/\beta$ decays toward zero.
These two identities are linked by $\beta_{\mathrm{peak}}\,\Delta^{*}_{\mathrm{DPO}}(\beta_{\mathrm{peak}})=\log v^{*}\approx 1.278$: they are two writings of the same one-dimensional threshold model, not independent estimators.

\paragraph{Empirical illustration.}
Figure~\ref{fig:beta-peak-nll-kl-delta} overlays Equation~\eqref{eq:sat-margin-dpo} on a dense HelpSteer3 SGD sweep of the same protocol as Figure~\ref{fig:cause-drift}.
The four-point subset in Section~\ref{sec:beta-weakens-updates} is already non-monotone: seed-averaged final per-token KL $\{0.04,\,0.35,\,0.83,\,0.54\}$ peaks at the nearest sampled cell $\beta=2\times10^{-3}$, and $\beta\cdot\mathrm{KL}$ grows by more than two orders of magnitude.
On the dense grid of Figure~\ref{fig:beta-peak-nll-kl-delta} and Table~\ref{tab:app-nll-kl-delta}, min NLL / max KL / max mean $\overline{\Delta}$ sit between $\beta=2.5\times10^{-3}$ and $3\times10^{-3}$.
The ordering anomaly noted in Section~\ref{sec:beta-weakens-updates} (the $\beta=0.002$ trajectory overtaking $\beta=0.01$) is thus a predictable consequence of the threshold model rather than an instability.

\subsection{Estimating \texorpdfstring{$\tau$ from $\beta_{\mathrm{peak}}$ and from $\Delta(\beta)$}{tau from beta-peak and from Delta(beta)}}
\label{app:tau-estimators}

\paragraph{Two readings of $\tau$.}
Equations~\eqref{eq:app-g6-beta-peak}--\eqref{eq:app-g7-delta-peak} suggest two practical ways to recover $\tau$ from a $\beta$-sweep under standard DPO at fixed learning rate.
Figure~\ref{fig:beta-peak-nll-kl-delta} overlays both resulting curves:
\begin{enumerate}
  \item \emph{From the peak location} (Eq.~\eqref{eq:app-g6-beta-peak}):
  $\tau=\beta_{\mathrm{peak}}/(1+v^{*})$.
  This is our primary estimate: $\beta_{\mathrm{peak}}$ is read jointly from validation NLL, KL, and preference-margin statistics.
  \item \emph{From the $\Delta(\beta)$ curve}: either plug a height into Eq.~\eqref{eq:app-g7-delta-peak}, or fit the shape of the equilibrium margin
  $\Delta^{*}_{\mathrm{cl}}(\beta;\tau)=\frac1\beta\log(\beta/\tau-1)$ ($\beta>2\tau$) to an empirical margin statistic across $\beta$.
\end{enumerate}
The second route is a consistency check, not a replacement for the first: the one-dimensional saturating scalar $\Delta^{*}_{\mathrm{cl}}$ is not directly observed.

\paragraph{Why $\Delta^{*}_{\mathrm{cl}}$ is not observed.}
The threshold derivation treats a single preference pair with scalar margin $\Delta$.
On validation we instead see a heavy-tailed distribution of margins over $\sim 1920$ pairs.
Empirical summaries---mean $\overline{\Delta}$, median, and upper quantile $\Delta_{p95}$---are therefore not interchangeable with $\Delta^{*}_{\mathrm{cl}}$.
In particular, $\overline{\Delta}$ is typically several times smaller than the one-pair peak height implied by Eq.~\eqref{eq:app-g7-delta-peak}, so a naive G.7-style inversion of the mean is not a valid $\tau$ reading.
The upper quantile $\Delta_{p95}$ is closer in magnitude and can be used cautiously for height-based checks; the preferred $\Delta$-based estimator below uses the \emph{shape} of $\overline{\Delta}(\beta)$ with a free vertical scale.

\paragraph{Uncertainty in locating $\beta_{\mathrm{peak}}$.}
On a discrete $\beta$ grid the exact peak is often ambiguous: neighbouring cells can agree within seed noise for mean $\Delta$ and min NLL, while the median may peak one cell away.
It is therefore useful to keep several $\beta$ points near the apparent maximum and to report a mid-cell anchor with a grid half-width.
On HelpSteer3 / Qwen3-4B (SGD, $\mathrm{lr}=2\cdot10^{-4}$, batch~24, 6~epochs; Figure~\ref{fig:beta-peak-nll-kl-delta}), mean $\Delta$ and min NLL at $\beta=0.0025$ and $\beta=0.003$ are tied within seed error, while the median is slightly higher at $0.003$.
We take
\[
\beta_{\mathrm{peak}}^{\mathrm{mid}}=0.00275,
\qquad
\tau_{\mathrm{G.6}}^{\mathrm{mid}}
=
\frac{0.00275}{1+v^{*}}
\approx 5.99\cdot10^{-4},
\]
with grid uncertainty given by the neighbouring cells $0.0025$ and $0.003$:
\[
\tau_{\mathrm{G.6}}=(5.99\pm0.54)\cdot10^{-4}
\quad(\pm\sim9\%\text{ relative to mid}),
\]
since $\delta\tau/\tau=\delta\beta_{\mathrm{peak}}/\beta_{\mathrm{peak}}$.
With this $\tau$, the left shaded region of Figure~\ref{fig:beta-peak-nll-kl-delta} is the dead zone $\beta\le 2\tau\approx 1.2\times10^{-3}$: it contains $\beta=5\times10^{-4}$ (near-zero KL) and the edge point $\beta=10^{-3}$.
On the right, the rough finite-horizon scale $\beta_{\mathrm{r}}\sim 10^{-2}$ (the scalar proxy gives $2.7\times10^{-2}$; Appendix~\ref{app:right-dead-zone}) marks the broad region where the $\log(\beta/\tau)/\beta$ ceiling is already decreasing; by $\beta=0.2$--$0.3$ the measured KL has collapsed.

\paragraph{Why validation NLL uses $\beta=1$.}
The training DPO loss is $-\log\sigma(\beta\Delta)$ with the run's own $\beta$.
That quantity is not comparable across a $\beta$-sweep: it depends on the product $\beta\Delta$, so a smaller training $\beta$ pushes the loss toward $\log 2$ even when the policy ranks chosen vs.\ rejected more strongly (Section~\ref{sec:loss-incomparability}).
The left panel of Figure~\ref{fig:beta-peak-nll-kl-delta} therefore reports a different diagnostic, \emph{validation pair NLL with $\beta=1$}:
\[
\mathbb{E}\bigl[-\log\sigma(s_c-s_r)\bigr]
=
\mathbb{E}\bigl[-\log\sigma(\Delta_{\theta})\bigr],
\]
where $s_c-s_r=\Delta_{\theta}=\log\pi_{\theta}(y_w\mid x)-\log\pi_{\theta}(y_l\mid x)$ on held-out pairs, and the training $\beta$ of that run is not inserted into the sigmoid.
This is a monotone decreasing function of the policy preference gap $\Delta_{\theta}=\Delta+\Delta_{\mathrm{ref}}$ ($\Delta_{\mathrm{ref}}$ is fixed by the reference model and data).
Consequently NLL falls when the learned margin $\Delta$ rises: a $\beta$-sweep that peaks in $\Delta$ exhibits a trough in min NLL at the same location, as in Figure~\ref{fig:beta-peak-nll-kl-delta}.
This motivates reading the NLL panel as an inverted copy of the saturation curve,
\begin{equation}
\label{eq:app-nll-delta}
\mathrm{NLL}(\beta)
\approx
a-b\,\Delta^{*}_{\mathrm{cl}}(\beta;\tau),
\qquad b>0,
\end{equation}
where the intercept $a$ and slope $b$ are obtained by least squares on the active region $\beta>2\tau$ (or with an empirical margin statistic in place of $\Delta^{*}_{\mathrm{cl}}$).
These two constants absorb the unknown map from the one-pair ceiling to the pair-averaged NLL; they are not estimators of $\tau$.
The tolerance continues to be read from $\beta_{\mathrm{peak}}$ or from the shape of $\overline{\Delta}(\beta)$ as below.

\paragraph{Validation NLL / KL and margin statistics.}
Table~\ref{tab:app-nll-kl-delta} reports the numbers plotted in Figure~\ref{fig:beta-peak-nll-kl-delta}: for each run we take the epoch of minimum validation pair NLL (with $\beta=1$), and at that same epoch report per-token MC KL, $\beta\cdot\mathrm{KL}$, and validation margin quantiles $\Delta=\Delta_{\theta}-\Delta_{\mathrm{ref}}$ (full val set).
Uncertainties are mean$\pm$std over seeds when $n>1$, matching the error bars in Figure~\ref{fig:beta-peak-nll-kl-delta}; far-right points $\beta\in\{0.2,0.3\}$ are single-seed (seed~43).
Bold entries mark column maxima: min NLL / max KL / max mean and $p95$ $\Delta$ at $\beta=0.0025$, max median $\Delta$ at $\beta=0.003$, and max $\beta\cdot\mathrm{KL}$ at $\beta=0.02$.

\begin{table}[t]
\centering
\scriptsize
\setlength{\tabcolsep}{3.5pt}
\caption{HelpSteer3 / Qwen3-4B standard DPO (SGD, $\mathrm{lr}=2\cdot10^{-4}$): min validation pair NLL (with $\beta=1$); per-token MC KL, $\beta\cdot\mathrm{KL}$, and validation margin quantiles at the same epoch (rounded). Same sweep as Figure~\ref{fig:beta-peak-nll-kl-delta}; uncertainties are mean$\pm$std over seeds.}
\label{tab:app-nll-kl-delta}
\begin{tabular}{rcccccccc}
\toprule
$\beta$ & $n$ & min NLL & KL & $\beta\cdot\mathrm{KL}$ & mean $\Delta$ & med.\ $\Delta$ & $p95$ $\Delta$ \\
\midrule
0.0005 & 4 & $13.93\pm 0.03$ & $0.042\pm 0.007$ & $2.1\cdot10^{-5}$ & $16.2\pm 0.9$ & $10.2\pm 0.5$ & $112\pm 7$ \\
0.001  & 4 & $13.46\pm 0.04$ & $0.353\pm 0.004$ & $3.5\cdot10^{-4}$ & $28.1\pm 0.3$ & $18.0\pm 0.0$ & $177\pm 1$ \\
0.0015 & 3 & $11.45\pm 0.15$ & $0.611\pm 0.041$ & $9.2\cdot10^{-4}$ & $73.3\pm 6.8$ & $40.2\pm 2.0$ & $431\pm 44$ \\
0.002  & 4 & $10.35\pm 0.04$ & $0.833\pm 0.014$ & $1.7\cdot10^{-3}$ & $122\pm 4$ & $69.5\pm 1.9$ & $607\pm 13$ \\
\textbf{0.0025} & \textbf{5} & $\mathbf{9.92\pm 0.14}$ & $\mathbf{1.00\pm 0.02}$ & $2.5\cdot10^{-3}$ & $\mathbf{152\pm 6}$ & $87.1\pm 4.4$ & $\mathbf{708\pm 30}$ \\
0.003  & 4 & $9.91\pm 0.07$ & $0.994\pm 0.048$ & $3.0\cdot10^{-3}$ & $152\pm 7$ & $\mathbf{88.4\pm 2.1}$ & $707\pm 24$ \\
0.005  & 2 & $10.26\pm 0.01$ & $0.842\pm 0.036$ & $4.2\cdot10^{-3}$ & $120\pm 0$ & $72.1\pm 0.2$ & $536\pm 0$ \\
0.0075 & 2 & $11.07\pm 0.05$ & $0.659\pm 0.008$ & $4.9\cdot10^{-3}$ & $90.6\pm 0.7$ & $58.0\pm 0.0$ & $399\pm 4$ \\
0.01   & 4 & $11.55\pm 0.08$ & $0.540\pm 0.008$ & $5.4\cdot10^{-3}$ & $75.1\pm 0.6$ & $49.2\pm 0.9$ & $321\pm 6$ \\
0.02   & 2 & $12.61\pm 0.01$ & $0.290\pm 0.024$ & $\mathbf{5.8\cdot10^{-3}}$ & $45.4\pm 0.0$ & $33.0\pm 1.4$ & $177\pm 1$ \\
0.04   & 2 & $13.23\pm 0.01$ & $0.130\pm 0.007$ & $5.2\cdot10^{-3}$ & $29.4\pm 0.0$ & $24.0\pm 0.0$ & $106\pm 0$ \\
0.075  & 2 & $13.36\pm 0.02$ & $0.066\pm 0.001$ & $4.9\cdot10^{-3}$ & $21.9\pm 0.3$ & $19.0\pm 1.4$ & $68\pm 0$ \\
0.2    & 1 & $13.67$ & $0.019$ & $3.8\cdot10^{-3}$ & $14.9$ & $14.0$ & $40$ \\
0.3    & 1 & $13.80$ & $0.010$ & $2.9\cdot10^{-3}$ & $12.3$ & $12.0$ & $32$ \\
\bottomrule
\end{tabular}
\end{table}

\begin{figure}[t]
\centering
\includegraphics[width=0.589\linewidth]{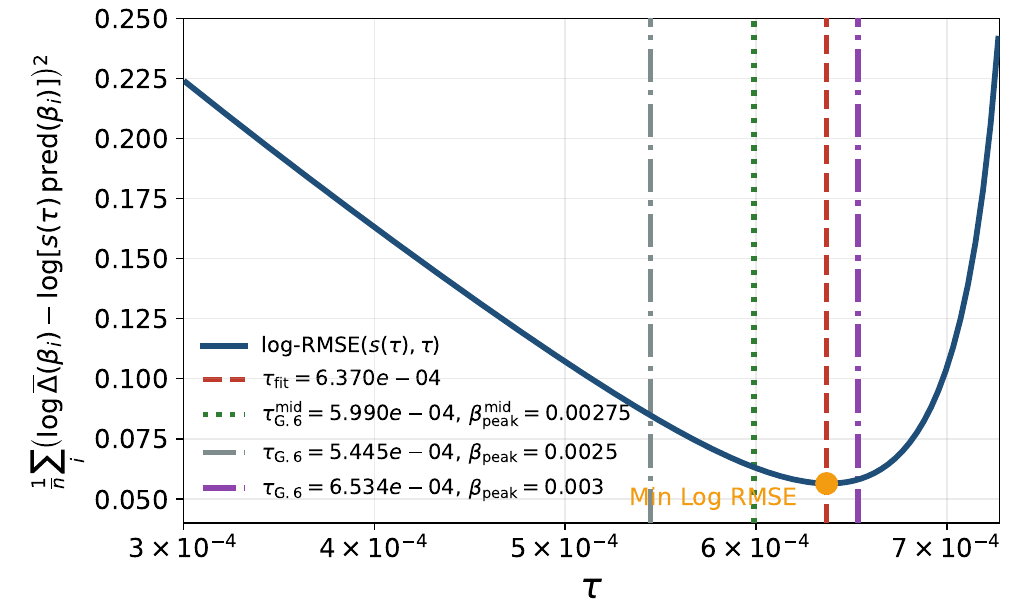}
\caption{Shape+scale log-RMSE versus $\tau$ for mean validation $\Delta$ on $\beta>2\tau_{\mathrm{rough}}$ (HelpSteer3 / Qwen3-4B, including $\beta=0.2,0.3$).
Vertical lines: $\tau_{\mathrm{fit}}$ and G.6 anchors at $\beta_{\mathrm{peak}}\in\{0.0025,0.00275,0.003\}$.}
\label{fig:app-tau-shape-scale-loss}
\end{figure}

\paragraph{Shape+scale fit of $\overline{\Delta}(\beta)$.}
On the active region $\beta>2\tau$ we relate the observed mean to the one-pair formula by
\begin{equation}
\label{eq:app-shape-scale}
\overline{\Delta}(\beta)
\approx
s\cdot\Delta^{*}_{\mathrm{cl}}(\beta;\tau).
\end{equation}
In practice we form a rough peak estimate $\beta_{\mathrm{peak}}^{\mathrm{mid}}=0.00275$, set $\tau_{\mathrm{rough}}=\beta_{\mathrm{peak}}^{\mathrm{mid}}/(1+v^{*})$, and fit on all sweep points with $\beta>2\tau_{\mathrm{rough}}$ (here $\beta\in\{0.0015,\ldots,0.075,0.2,0.3\}$; $n=12$).
Parameters are obtained by log-RMSE in two steps.
First, for each candidate $\tau$ on a grid, set $\mathrm{pred}(\beta)=\Delta^{*}_{\mathrm{cl}}(\beta;\tau)$.
Second, the optimal scale at that $\tau$ is the closed-form log-space least-squares value
\begin{equation}
\log s(\tau)
=
\frac1n\sum_{i}\Bigl(\log\overline{\Delta}(\beta_i)-\log\mathrm{pred}(\beta_i)\Bigr),
\qquad
s(\tau)=e^{\log s(\tau)}.
\end{equation}
We then choose $\tau$ to minimize
\begin{equation}
\frac1n\sum_{i}
\Bigl(\log\overline{\Delta}(\beta_i)-\log\bigl(s(\tau)\,\mathrm{pred}(\beta_i)\bigr)\Bigr)^{2}.
\end{equation}
Thus $\tau$ is read from the \emph{shape} of the curve and $s(\tau)$ from its height.
Once $\tau$ is obtained, the implied peak $\beta_{\mathrm{peak}}^{\mathrm{pred}}=\tau(1+v^{*})$ follows from Eq.~\eqref{eq:app-g6-beta-peak} without reusing $\Delta$.
Figure~\ref{fig:app-tau-shape-scale-loss} shows the objective versus $\tau$; the minimum is at $\tau_{\mathrm{fit}}\approx 6.37\cdot10^{-4}$ with $s\approx 0.35$.
The two overlays in Figure~\ref{fig:beta-peak-nll-kl-delta} are this shape+scale curve and the G.6 curve from $\beta_{\mathrm{peak}}^{\mathrm{mid}}$.

\paragraph{Comparison of $\tau$ estimates.}
Table~\ref{tab:app-tau-compare} compares G.6 anchors on the peak grid cell, the shape+scale fit, and a height-based check that inverts Eq.~\eqref{eq:app-g7-delta-peak} using $\Delta_{p95}$ at $\beta=0.0025$ (not the mean).
Upright entries are observed or fitted inputs; italic entries are quantities derived by the indicated map ($\beta\to\tau$ or $\tau\to\beta_{\mathrm{peak}}^{\mathrm{pred}}$).

\begin{table}[t]
\centering
\small
\caption{Consistency of $\tau$ estimates on HelpSteer3 / Qwen3-4B, corresponding to the two overlays in Figure~\ref{fig:beta-peak-nll-kl-delta}.
Reference: $\tau_{\mathrm{G.6}}^{\mathrm{mid}}$ at $\beta_{\mathrm{peak}}^{\mathrm{mid}}=0.00275$.
Shape+scale uses $\beta>2\tau_{\mathrm{rough}}$ with $\tau_{\mathrm{rough}}=\tau_{\mathrm{G.6}}^{\mathrm{mid}}$.}
\label{tab:app-tau-compare}
\begin{tabular}{llccc}
\toprule
method & map & $\tau$ & $\beta_{\mathrm{peak}}$ & $\tau/\tau_{\mathrm{G.6}}^{\mathrm{mid}}$ \\
\midrule
G.6, mid $[0.0025,0.003]$ &
  $\beta\to\tau$ &
  $\mathit{5.99\cdot10^{-4}}$ & $0.00275$ & $1.00$ \\
G.6, lower grid cell &
  $\beta\to\tau$ &
  $\mathit{5.45\cdot10^{-4}}$ & $0.0025$ & $0.91$ \\
G.6, upper grid cell &
  $\beta\to\tau$ &
  $\mathit{6.53\cdot10^{-4}}$ & $0.003$ & $1.09$ \\
shape+scale, mean, $\beta>2\tau_{\mathrm{rough}}$ &
  $\tau\to\beta$ &
  $6.37\cdot10^{-4}$ & $\mathit{0.00292}$ & $1.06$ \\
Eq.~\eqref{eq:app-g7-delta-peak} on $\Delta_{p95}$ at $\beta=0.0025$ &
  $\Delta_{p95}\to\tau\to\beta$ &
  $\mathit{3.94\cdot10^{-4}}$ & $\mathit{0.00181}$ & $0.66$ \\
\bottomrule
\end{tabular}
\end{table}

\paragraph{Takeaway.}
The primary estimate remains Eq.~\eqref{eq:app-g6-beta-peak} from $\beta_{\mathrm{peak}}$, with an explicit grid uncertainty when the peak cell is not unique.
Fitting $\overline{\Delta}\approx s\,\Delta^{*}_{\mathrm{cl}}$ on $\beta>2\tau_{\mathrm{rough}}$ recovers a compatible $\tau$ (here $+6\%$ relative to mid and $-2.5\%$ relative to the $\beta=0.003$ G.6 anchor), providing an internal consistency check that the shape of the mean margin tracks the threshold model once a height scale is freed; the two curves in Figure~\ref{fig:beta-peak-nll-kl-delta} nearly coincide.
Because both estimates use the same $\beta$-sweep, their agreement is an internal consistency check rather than an out-of-sample validation.
A height-only inversion of Eq.~\eqref{eq:app-g7-delta-peak} should use an upper quantile such as $\Delta_{p95}$, not the raw mean.

\paragraph{Scope and caveats.}
First, the threshold model predicts the $\beta$-shape of the saturation margin, and pair NLL is monotone in that gap via Equation~\eqref{eq:app-nll-delta}. We do not treat per-token KL as a general monotone function of the pair margin. In the HelpSteer3 SGD sweep the two quantities co-move over a wide range, which is why Figure~\ref{fig:beta-peak-nll-kl-delta} reports both, but we do not fit a global $\mathrm{KL}$--$\Delta$ law or proportionality constants.
Second, the threshold model applies only when training reaches saturation before the epoch budget ends.
For the very small $\beta$ values in Figure~\ref{fig:intro-overview}, the implied $\Delta^{*}_{\mathrm{DPO}}$ is too large to reach in practice; those runs stop early because updates are weak and Adam partially compensates (Appendix~\ref{app:adam-threshold}), not because the loss has saturated.
Third, under a decaying learning-rate schedule, $\tau$ is only a coarse effective tolerance, so we interpret its magnitude and the functional form, not a precise fit.
Finally, the extra $\beta$ prefactor in standard DPO (Equation~\eqref{eq:sat-margin-dpo}) is a fixed-learning-rate effect: rescaling the learning rate by $1/\beta$ (Section~\ref{sec:normalized}) or operating Adam in its cancellation regime (Appendix~\ref{app:adam-threshold}) largely removes it and restores the cleaner $1/\beta$ scaling of the normalized objective (Equation~\eqref{eq:sat-margin-norm}).
Appendix~\ref{app:right-dead-zone} complements the left dead zone $\beta\le 2\tau$ with an order-of-magnitude right-hand horizon scale obtained from a scalar margin closure. On the HelpSteer3 SGD sweep, the initial batch-Jacobian proxy gives $\beta_{\mathrm{r}}\approx0.027$, but time variation and cross-pair Gram terms limit this estimate to a broad \(\sim10^{-2}\) region rather than a calibrated cutoff.

\section{Loss-Level Scale Invariance: Identical Loss Curves, Different Policies}
\label{app:loss-invariance}

\setcounter{equation}{0}
\renewcommand{\theequation}{H.\arabic{equation}}
\renewcommand{\theHequation}{\theequation}

\begin{figure}[H]
  \centering
  \includegraphics[width=0.78\linewidth]{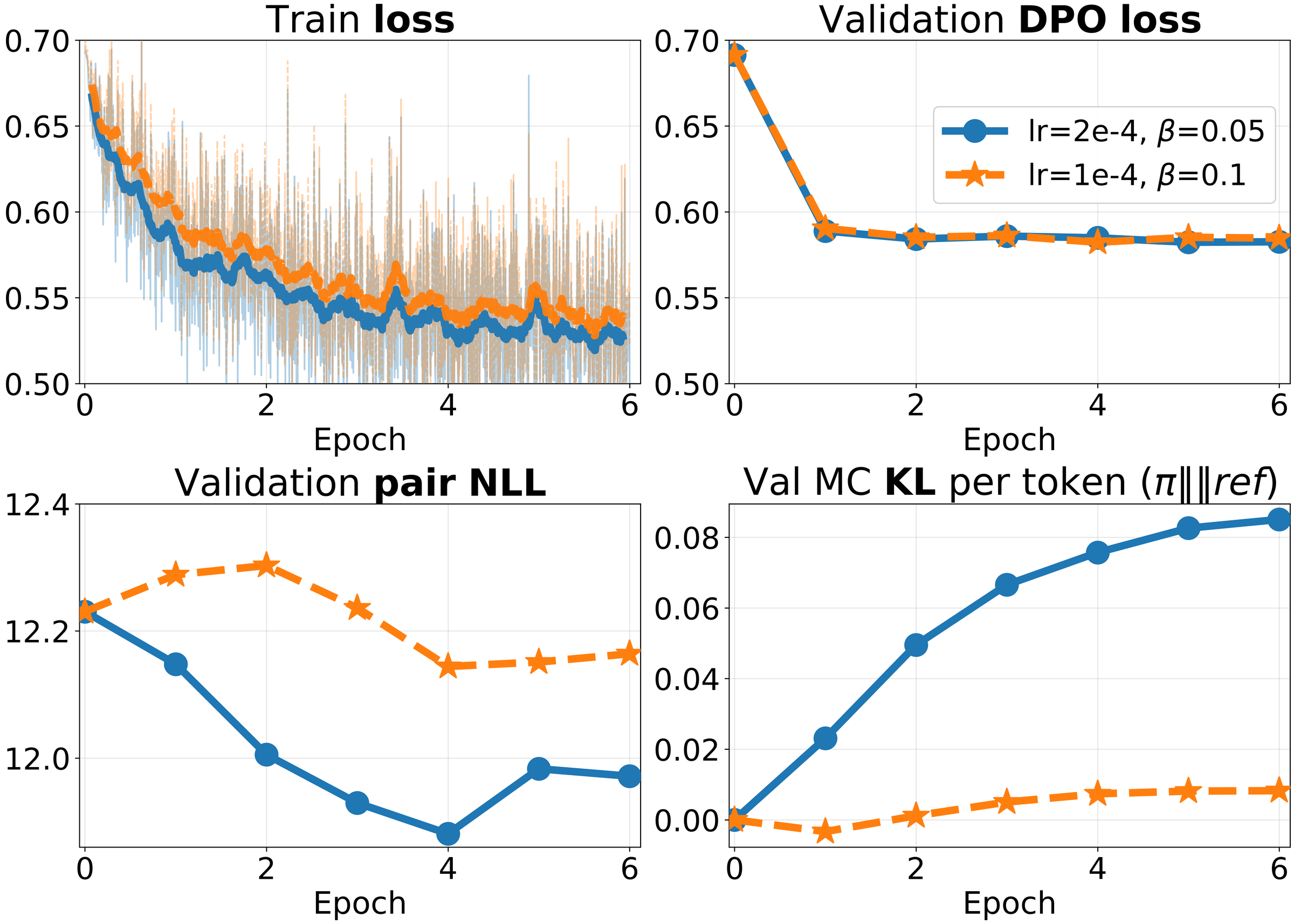}
  \caption{HelpSteer3 standard-DPO comparison with scale-equivalent learning rates using TRL \texttt{DPOTrainer} (Qwen3-4B-Instruct-2507, SGD, batch size~4, gradient accumulation~4, seed~51): $(\beta,\mathrm{lr})=(0.05,\,2\times10^{-4})$ and $(0.1,\,1\times10^{-4})$, so $\eta\beta=10^{-5}$ in both runs. Table~\ref{tab:loss-invariance-hsteer} lists the same checkpoints.}
  \label{fig:hsteer-beta-lr-const}
\end{figure}

\begin{figure}[H]
  \centering
  \includegraphics[width=0.78\linewidth]{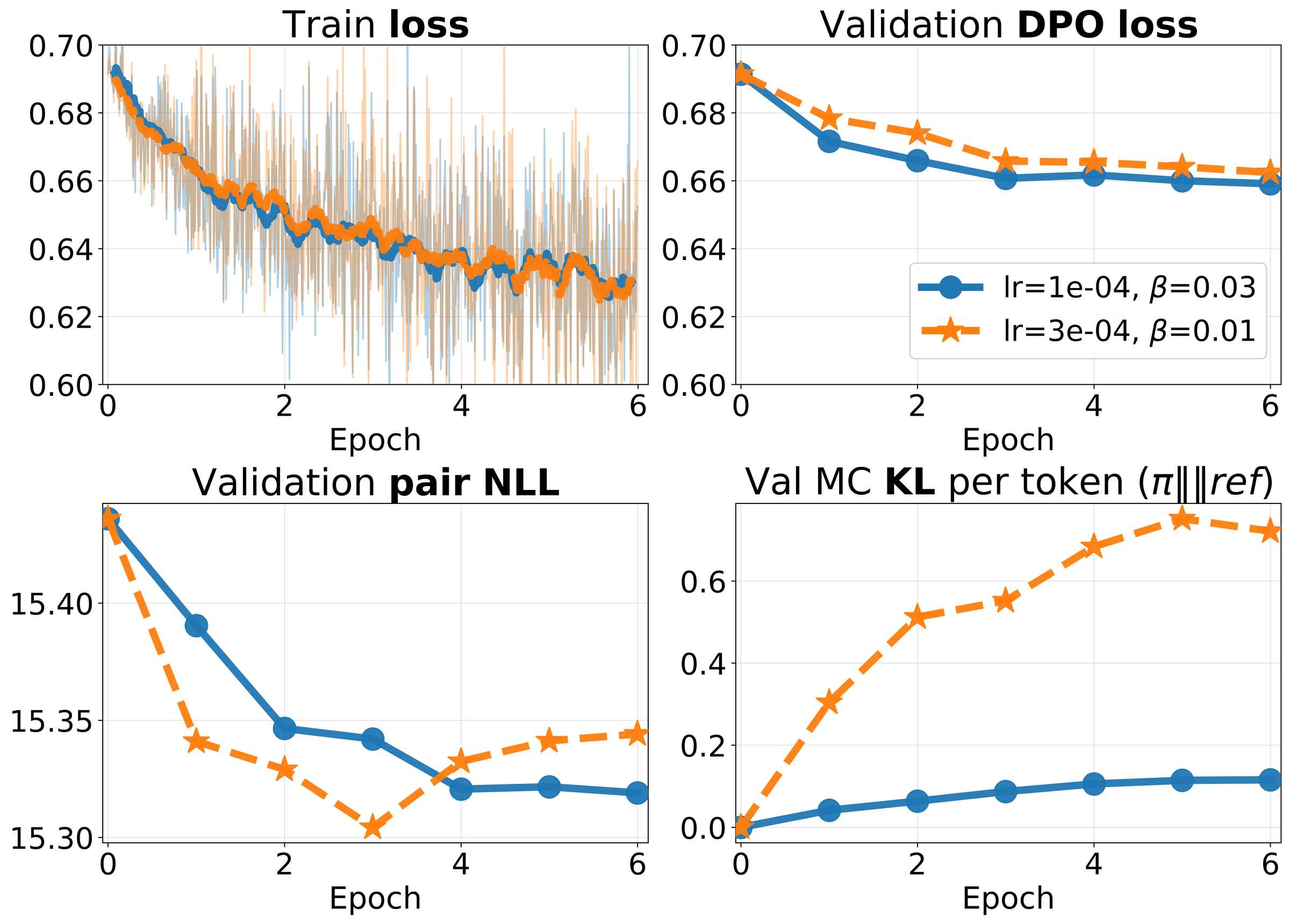}
  \caption{UltraFeedback Binarized standard-DPO comparison with scale-equivalent learning rates using TRL \texttt{DPOTrainer} (Ministral-3B-Instruct, SGD, batch size~4, gradient accumulation~4, seed~52): $(\beta,\mathrm{lr})=(0.01,\,3\times10^{-4})$ and $(0.03,\,1\times10^{-4})$, so $\eta\beta=3\times10^{-6}$ in both runs. Table~\ref{tab:loss-invariance-mistral3b} lists the same checkpoints.}
  \label{fig:mistral3b-beta-lr-const}
\end{figure}

\begin{figure}[H]
  \centering
  \includegraphics[width=0.78\linewidth]{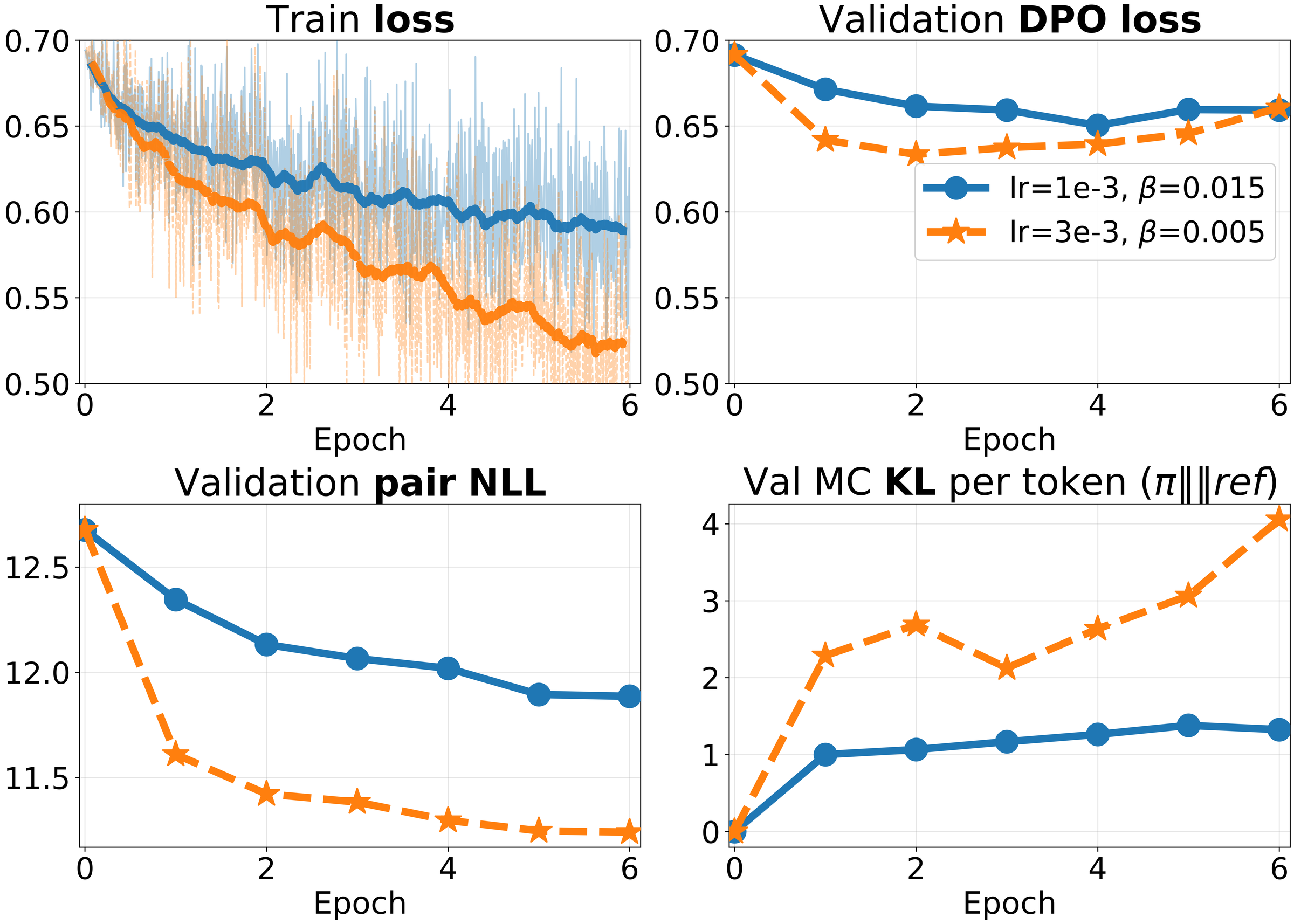}
  \caption{PKU-processed HH-RLHF standard-DPO comparison with scale-equivalent learning rates using TRL \texttt{DPOTrainer} (Qwen2.5-3B-Instruct, SGD, batch size~4, gradient accumulation~4, seed~52): $(\beta,\mathrm{lr})=(0.015,\,10^{-3})$ and $(0.005,\,3\times10^{-3})$, so $\eta\beta=1.5\times10^{-5}$ in both runs. Table~\ref{tab:loss-invariance-hhrlhf} lists the same checkpoints. Raw training losses (top-left) do not coincide, unlike the smoothed HelpSteer3 and Ministral-3B curves above.}
  \label{fig:hhrlhf-beta-lr-const}
\end{figure}

\paragraph{Observation.}
Throughout this appendix, the \emph{KL ratio} at a given checkpoint is the ratio of Monte-Carlo forward per-token validation KL values,
$\mathrm{KL\ ratio}=\mathrm{KL}(\pi_{\theta}\,\|\,\pi_{\mathrm{ref}})_{\beta_1}/\mathrm{KL}(\pi_{\theta}\,\|\,\pi_{\mathrm{ref}})_{\beta_2}$,
with the smaller-$\beta$ run in the numerator and the larger-$\beta$ run in the denominator (equivalently, the ratio of the physical policy displacement readouts when the two runs share the same dimensionless margin $u=\beta\Delta$).
In the UltraFeedback comparison of Figure~\ref{fig:cause-failures}, the two standard-DPO runs ($(\beta,\mathrm{lr})=(0.01,\,10^{-3})$ and $(0.02,\,5\times10^{-4})$) produce training- and validation-loss curves that are nearly indistinguishable, while their validation NLL and KL differ substantially.
Table~\ref{tab:loss-invariance-6epoch} quantifies this from the six-epoch logs (mean over four seeds; validation every half epoch): from epoch~1 onward the validation DPO losses differ by at most $0.005$~nats, while the KL ratio stays near $2.9$--$4.1$.
Table~\ref{tab:loss-invariance} reports the same metrics for the three-epoch single-run comparison of Figure~\ref{fig:ultrafb-comparison-3epoch} (seed~42): from epoch~1 onward the validation DPO losses differ by at most $0.0017$~nats, while the KL ratio is stable at $3.3$--$3.9$.
The same pattern holds on HelpSteer3 (Figure~\ref{fig:hsteer-beta-lr-const}, Table~\ref{tab:loss-invariance-hsteer}): with $(\beta,\mathrm{lr})=(0.05,\,2\times10^{-4})$ and $(0.1,\,1\times10^{-4})$ (seed~51, effective batch size~16), validation DPO losses again agree to the third decimal, but the KL ratio reaches $\approx 10$ by epoch~6 because the larger-$\beta$ run remains close to the reference model.
A third TRL \texttt{DPOTrainer} comparison on UltraFeedback Binarized with Ministral-3B-Instruct (Figure~\ref{fig:mistral3b-beta-lr-const}, Table~\ref{tab:loss-invariance-mistral3b}): with $(\beta,\mathrm{lr})=(0.01,\,3\times10^{-4})$ and $(0.03,\,1\times10^{-4})$ (seed~52, effective batch size~16, $\eta\beta=3\times10^{-6}$), validation DPO losses stay within $0.009$~nats while the KL ratio stabilizes near $6.5$.
By contrast, on PKU-processed HH-RLHF with Qwen2.5-3B-Instruct (Figure~\ref{fig:hhrlhf-beta-lr-const}, Table~\ref{tab:loss-invariance-hhrlhf}), the same scale-equivalent setting $(\beta,\mathrm{lr})=(0.015,\,10^{-3})$ and $(0.005,\,3\times10^{-3})$ (seed~52, $\eta\beta=1.5\times10^{-5}$) yields separated raw training losses and validation DPO losses that differ by up to $0.03$~nats early on, even though the KL ratio remains stable at $\approx 1.8$--$3.1$.

\begin{table}[h]
\centering
\small
\caption{Validation DPO loss and per-token Monte-Carlo KL for the two standard-loss runs of Figure~\ref{fig:ultrafb-comparison-3epoch} (UltraFeedback, Qwen3-4B-Instruct-2507, SGD, learning rate scaled as $1/\beta$, seed~42). KL ratio $=\mathrm{KL}(\pi_{\theta}\,\|\,\pi_{\mathrm{ref}})_{\beta=0.01}/\mathrm{KL}(\pi_{\theta}\,\|\,\pi_{\mathrm{ref}})_{\beta=0.02}$ at each checkpoint. The loss curves coincide to the third decimal from epoch~1 onward, while the KL ratio stays near $3.3$--$3.9$.}
\label{tab:loss-invariance}
\begin{tabular}{lccccc}
\toprule
Epoch & \multicolumn{2}{c}{Val.\ DPO loss} & \multicolumn{2}{c}{Val.\ KL per token} & KL ratio \\
 & $\beta=0.01$ & $\beta=0.02$ & $\beta=0.01$ & $\beta=0.02$ & \\
\midrule
0.5 & 0.5728 & 0.5620 & 0.436 & 0.128 & 3.4 \\
1.0 & 0.5552 & 0.5557 & 0.455 & 0.140 & 3.3 \\
1.5 & 0.5463 & 0.5462 & 0.579 & 0.168 & 3.5 \\
2.0 & 0.5387 & 0.5385 & 0.616 & 0.158 & 3.9 \\
2.5 & 0.5361 & 0.5368 & 0.622 & 0.189 & 3.3 \\
3.0 & 0.5346 & 0.5363 & 0.630 & 0.194 & 3.3 \\
\bottomrule
\end{tabular}
\end{table}

\begin{table}[h]
\centering
\small
\caption{Validation DPO loss and per-token Monte-Carlo KL for the two standard-loss runs of Figure~\ref{fig:cause-failures} (UltraFeedback, Qwen3-4B-Instruct-2507, SGD, learning rate scaled as $1/\beta$; mean over four seeds). KL ratio $=\mathrm{KL}(\pi_{\theta}\,\|\,\pi_{\mathrm{ref}})_{\beta=0.01}/\mathrm{KL}(\pi_{\theta}\,\|\,\pi_{\mathrm{ref}})_{\beta=0.02}$ at each checkpoint. From epoch~1 onward the validation DPO losses differ by at most $0.005$~nats, while the KL ratio stays near $2.9$--$4.1$.}
\label{tab:loss-invariance-6epoch}
\begin{tabular}{lccccc}
\toprule
Epoch & \multicolumn{2}{c}{Val.\ DPO loss} & \multicolumn{2}{c}{Val.\ KL per token} & KL ratio \\
 & $\beta=0.01$ & $\beta=0.02$ & $\beta=0.01$ & $\beta=0.02$ & \\
\midrule
0.5 & 0.5727 & 0.5662 & 0.466 & 0.154 & 3.0 \\
1.0 & 0.5539 & 0.5520 & 0.484 & 0.154 & 3.1 \\
1.5 & 0.5458 & 0.5450 & 0.545 & 0.155 & 3.5 \\
2.0 & 0.5357 & 0.5373 & 0.622 & 0.152 & 4.1 \\
2.5 & 0.5303 & 0.5328 & 0.633 & 0.180 & 3.5 \\
3.0 & 0.5257 & 0.5307 & 0.639 & 0.203 & 3.2 \\
3.5 & 0.5246 & 0.5272 & 0.692 & 0.212 & 3.3 \\
4.0 & 0.5216 & 0.5245 & 0.706 & 0.226 & 3.1 \\
4.5 & 0.5209 & 0.5247 & 0.780 & 0.245 & 3.2 \\
5.0 & 0.5204 & 0.5223 & 0.802 & 0.261 & 3.1 \\
5.5 & 0.5235 & 0.5233 & 0.820 & 0.282 & 2.9 \\
6.0 & 0.5210 & 0.5219 & 0.814 & 0.278 & 2.9 \\
\bottomrule
\end{tabular}
\end{table}

\begin{table}[h]
\centering
\small
\caption{Validation DPO loss and per-token Monte-Carlo KL for the HelpSteer3 TRL \texttt{DPOTrainer} runs of Figure~\ref{fig:hsteer-beta-lr-const} (SGD, $\eta\beta=10^{-5}$, seed~51). KL ratio $=\mathrm{KL}(\pi_{\theta}\,\|\,\pi_{\mathrm{ref}})_{\beta=0.05}/\mathrm{KL}(\pi_{\theta}\,\|\,\pi_{\mathrm{ref}})_{\beta=0.1}$; entries marked ``---'' are omitted when the denominator KL is non-positive or too small for a stable ratio (Monte-Carlo noise near the reference policy).}
\label{tab:loss-invariance-hsteer}
\begin{tabular}{lccccc}
\toprule
Epoch & \multicolumn{2}{c}{Val.\ DPO loss} & \multicolumn{2}{c}{Val.\ KL per token} & KL ratio \\
 & $\beta=0.05$ & $\beta=0.1$ & $\beta=0.05$ & $\beta=0.1$ & \\
\midrule
1 & 0.5889 & 0.5904 & 0.0231 & $-0.0033$ & --- \\
2 & 0.5842 & 0.5852 & 0.0495 & 0.0012 & --- \\
3 & 0.5859 & 0.5862 & 0.0666 & 0.0050 & 13.2 \\
4 & 0.5849 & 0.5823 & 0.0757 & 0.0074 & 10.3 \\
5 & 0.5823 & 0.5852 & 0.0826 & 0.0081 & 10.1 \\
6 & 0.5826 & 0.5847 & 0.0851 & 0.0083 & 10.3 \\
\bottomrule
\end{tabular}
\end{table}

\begin{table}[h]
\centering
\small
\caption{Validation DPO loss and per-token Monte-Carlo KL for the UltraFeedback Binarized TRL \texttt{DPOTrainer} runs of Figure~\ref{fig:mistral3b-beta-lr-const} (Ministral-3B-Instruct, SGD, $\eta\beta=3\times10^{-6}$, seed~52). KL ratio $=\mathrm{KL}(\pi_{\theta}\,\|\,\pi_{\mathrm{ref}})_{\beta=0.01}/\mathrm{KL}(\pi_{\theta}\,\|\,\pi_{\mathrm{ref}})_{\beta=0.03}$ at each checkpoint.}
\label{tab:loss-invariance-mistral3b}
\begin{tabular}{lccccc}
\toprule
Epoch & \multicolumn{2}{c}{Val.\ DPO loss} & \multicolumn{2}{c}{Val.\ KL per token} & KL ratio \\
 & $\beta=0.01$ & $\beta=0.03$ & $\beta=0.01$ & $\beta=0.03$ & \\
\midrule
1 & 0.6784 & 0.6716 & 0.304 & 0.041 & 7.3 \\
2 & 0.6740 & 0.6659 & 0.512 & 0.064 & 8.0 \\
3 & 0.6658 & 0.6607 & 0.552 & 0.087 & 6.3 \\
4 & 0.6654 & 0.6617 & 0.683 & 0.106 & 6.5 \\
5 & 0.6641 & 0.6600 & 0.752 & 0.114 & 6.6 \\
6 & 0.6624 & 0.6591 & 0.721 & 0.116 & 6.2 \\
\bottomrule
\end{tabular}
\end{table}

\begin{table}[h]
\centering
\small
\caption{Validation DPO loss and per-token Monte-Carlo KL for the PKU-processed HH-RLHF TRL \texttt{DPOTrainer} runs of Figure~\ref{fig:hhrlhf-beta-lr-const} (Qwen2.5-3B-Instruct, SGD, $\eta\beta=1.5\times10^{-5}$, seed~52). KL ratio $=\mathrm{KL}(\pi_{\theta}\,\|\,\pi_{\mathrm{ref}})_{\beta=0.005}/\mathrm{KL}(\pi_{\theta}\,\|\,\pi_{\mathrm{ref}})_{\beta=0.015}$ at each checkpoint.}
\label{tab:loss-invariance-hhrlhf}
\begin{tabular}{lccccc}
\toprule
Epoch & \multicolumn{2}{c}{Val.\ DPO loss} & \multicolumn{2}{c}{Val.\ KL per token} & KL ratio \\
 & $\beta=0.015$ & $\beta=0.005$ & $\beta=0.015$ & $\beta=0.005$ & \\
\midrule
1 & 0.6714 & 0.6419 & 1.001 & 2.287 & 2.3 \\
2 & 0.6616 & 0.6335 & 1.067 & 2.688 & 2.5 \\
3 & 0.6593 & 0.6375 & 1.168 & 2.125 & 1.8 \\
4 & 0.6504 & 0.6395 & 1.263 & 2.634 & 2.1 \\
5 & 0.6596 & 0.6457 & 1.380 & 3.064 & 2.2 \\
6 & 0.6593 & 0.6607 & 1.325 & 4.056 & 3.1 \\
\bottomrule
\end{tabular}
\end{table}

\paragraph{When do the losses coincide?}
Matching $\eta\beta$ is necessary but not sufficient for overlapping loss curves.
The standard loss depends only on $u=\beta\Delta$ (Eq.~\eqref{eq:loss-invariance}), so training and validation losses coincide only if the two runs follow a common $u$-trajectory.
With $\mathrm{lr}\propto 1/\beta$, the physical margin grows at comparable rates ($\mathrm{d}\Delta/\mathrm{d}t\propto\sigma(-u)$), but $u$ advances faster for larger $\beta$, so that run initially leads.
Saturation, however, is a condition on $u$ alone: the per-step drift criterion $\eta g^{2}\beta\,\sigma(-\beta\Delta)\lesssim\varepsilon$ (Appendix~\ref{app:saturation}, Setup) depends on $(\eta,\beta)$ only through the matched product $\eta\beta$, so both runs share one threshold $\sigma(-u)\lesssim\varepsilon/(\eta\beta g^{2})$: the larger-$\beta$ run reaches it first and slows down, while the smaller-$\beta$ run---still at smaller $u$ for the same $\Delta$---keeps catching up until both settle near a common $u^{*}$.
Their losses then coincide even though $\Delta=u/\beta$ (and KL) remain different.
This catch-up succeeds on UltraFeedback, HelpSteer3, and Ministral-3B (Tables~\ref{tab:loss-invariance}--\ref{tab:loss-invariance-mistral3b}): validation DPO losses agree to $10^{-3}$--$10^{-2}$~nats and smoothed training losses are nearly overlapping.
It does not fully occur on HH-RLHF (Figure~\ref{fig:hhrlhf-beta-lr-const}, Table~\ref{tab:loss-invariance-hhrlhf}): both runs move far from the reference (KL $\gtrsim 1$~nat/token), the corpus is comparatively noisy, the $\beta$ ratio is $3$, and the early transient does not fully close within six epochs---validation DPO losses differ by up to $0.03$~nats and raw training losses remain separated (final values $0.62$ vs.\ $0.53$).
In all cases, however, the KL ratio is a more stable readout of physical displacement: it stays near $2.9$--$4.1$ (UltraFeedback, six epochs) and $3.3$--$3.9$ (three epochs), $\approx 10$ (HelpSteer3), $6.2$--$8.0$ (Ministral-3B), and $\approx 1.8$--$3.1$ (HH-RLHF) once the early transient passes, even when the loss curves do not fully coincide.
Table~\ref{tab:loss-invariance-summary} collects these scale-equivalent comparisons side by side.

\begin{table}[h]
\centering
\scriptsize
\setlength{\tabcolsep}{2.5pt}
\renewcommand{\arraystretch}{1.05}
\caption{Summary of scale-equivalent standard-DPO pairs ($\mathrm{lr}\propto 1/\beta$, matched $\eta\beta$). $|\Delta\,\mathrm{val}|$: absolute difference between the two validation DPO losses (nats); max / final. Train-loss rows: \emph{all} = overall verdict; \emph{sm.} = moving-average curve ($2.5\%$ window); \emph{raw} = final per-step values logged at the last checkpoint (HelpSteer3 raw values differ by $\approx 0.025$ despite overlapping smoothed curves). KL ratio $=\mathrm{KL}_{\beta_{\mathrm{small}}}/\mathrm{KL}_{\beta_{\mathrm{large}}}$; HelpSteer3 from epoch~3 onward.}
\label{tab:loss-invariance-summary}
\resizebox{\linewidth}{!}{%
\begin{tabular}{@{}llcc c p{0.62in} c@{}}
\toprule
Data & Model & $\eta\beta$ & $\beta_2/\beta_1$ &
\shortstack{$|\Delta\,\mathrm{val}|$\\max / final} &
\shortstack{Train loss\\ \scriptsize (all / sm.\ / raw)} &
\shortstack{KL\\ratio} \\
\midrule
UF 3\,ep. & Q3-4B & $10^{-5}$ & 2 & $\le$0.0017 / $\le$0.0017 &
\shortstack{overlap\\ overlap\\ overlap} & 3.3--3.9 \\
\cmidrule(lr){1-7}
UF 6\,ep. & Q3-4B & $10^{-5}$ & 2 & 0.0051 / 0.0009 &
\shortstack{overlap\\ overlap\\ overlap} & 2.9--4.1 \\
\cmidrule(lr){1-7}
HSteer3 & Q3-4B & $10^{-5}$ & 2 & 0.0029 / 0.0021 &
\shortstack{overlap\\ overlap\\ diverge} & $\approx 10$ \\
\cmidrule(lr){1-7}
UF bin. & M3-3B & $3{\times}10^{-6}$ & 3 & 0.0081 / 0.0033 &
\shortstack{overlap\\ overlap\\ overlap} & 6.2--8.0 \\
\cmidrule(lr){1-7}
HH-RLHF & Q2.5-3B & $1.5{\times}10^{-5}$ & 3 & 0.0295 / 0.0014 &
\shortstack{diverge\\ diverge\\ diverge} & 1.8--3.1 \\
\bottomrule
\end{tabular}%
}
\end{table}

\paragraph{Invariance of the loss.}
The coincidence is structural, not accidental.
The standard pairwise loss depends on $(\beta,\Delta)$ only through the dimensionless product $u=\beta\Delta$:
\begin{equation}
\label{eq:loss-invariance}
\mathcal{L}_{\mathrm{DPO}}(\Delta;\beta)=\log\!\left(1+e^{-\beta\Delta}\right)
=\mathrm{softplus}(-u),
\qquad
\mathcal{L}_{\mathrm{DPO}}\!\left(\Delta/c;\,c\beta\right)=\mathcal{L}_{\mathrm{DPO}}(\Delta;\beta)
\quad\forall\,c>0.
\end{equation}
Its level sets in the $(\beta,\Delta)$ plane are hyperbolas $\beta\Delta=\mathrm{const}$: the loss value identifies the dimensionless margin $u$ and is blind to the physical margin $\Delta$, and can therefore hide large differences in policy displacement, including KL from the reference.
For example, $(\beta,\Delta)=(0.01,100)$ and $(0.02,50)$ yield the identical loss $\log(1+e^{-1})\approx 0.313$ although the first policy has moved twice as far from the reference.
The same invariance holds for every objective of the form $L(\beta\Delta)$ in Table~\ref{tab:dpo-losses}.

\paragraph{Why the two runs can approach similar $u$-trajectories.}
Under the scalar frozen-Jacobian approximation of Appendix~\ref{app:saturation}, $\mathrm{lr}\propto 1/\beta$ gives $\mathrm{d}\Delta/\mathrm{d}t\propto\sigma(-u)$ and the same drift-based saturation threshold $\sigma(-u)\lesssim\varepsilon/(\eta\beta g^{2})$ for both runs.
This approximation predicts late-time catch-up in $u$, consistent with several reported sweeps, but it is not a general trajectory-equivalence result for minibatch training because the Jacobians, cross-pair Gram terms, and stochastic gradients evolve during optimization.
The KL ratio is not fixed by this identity alone: Table~\ref{tab:loss-invariance-6epoch} gives $\approx 2.9$--$4.1$ on UltraFeedback (six epochs) and Table~\ref{tab:loss-invariance} gives $\approx 3.3$--$3.9$ (three epochs), Table~\ref{tab:loss-invariance-mistral3b} gives $\approx 6.2$--$8.0$ on UltraFeedback Binarized with Ministral-3B-Instruct, Table~\ref{tab:loss-invariance-hsteer} gives $\approx 10$ once the larger-$\beta$ run has moved only slightly from the reference model, and Table~\ref{tab:loss-invariance-hhrlhf} gives $\approx 1.8$--$3.1$ on HH-RLHF.

\paragraph{Early transient.}
The only phase in which the loss curves should differ is the start of training.
While $u\ll 1$, both runs grow the physical margin at nearly the same rate because $\sigma(-u)\approx 1/2$, so over the same time interval $u=\beta\Delta$ is initially larger for the larger-$\beta$ run and its loss drops faster.
Table~\ref{tab:loss-invariance} makes this visible at epoch~$0.5$: the $\beta=0.02$ validation loss is $0.5620$ versus $0.5728$ for $\beta=0.01$, even though the smaller-$\beta$ run already has higher KL ($0.436$ vs.\ $0.128$).
Once saturation sets in, the larger-$\beta$ run slows first and the smaller-$\beta$ run catches up in $u$; by epoch~$1$ the validation losses differ by only $5\times10^{-4}$~nats and remain within $10^{-3}$ thereafter.

\paragraph{The normalized loss reads the physical scale.}
The centered-softplus objective is \emph{not} a function of $u$ alone:
$s_c(-\Delta;\beta)=\bigl(\mathrm{softplus}(-u)-\ln 2\bigr)/\beta$, so at equal $u$ its value scales as $1/\beta$ and the two runs are predicted to separate by a factor of $2$.
The final logged training losses confirm this: $-17.5$ for $\beta=0.01$ versus $-9.0$ for $\beta=0.02$, a ratio of $1.95$ (bottom-right panel of Figure~\ref{fig:ultrafb-comparison-3epoch}); the four-seed six-epoch comparison in Figure~\ref{fig:cause-failures} exhibits the same factor-of-two separation (bottom-middle panel).
The overlap of the standard losses and the $2\times$ separation of the normalized losses are thus two sides of the same identity~\eqref{eq:loss-invariance}.

\section{Per-run Trajectories under the Normalized Objective}
\label{app:normalized-trajectories}

\begin{figure}[H]
  \centering
  \includegraphics[width=0.92\linewidth]{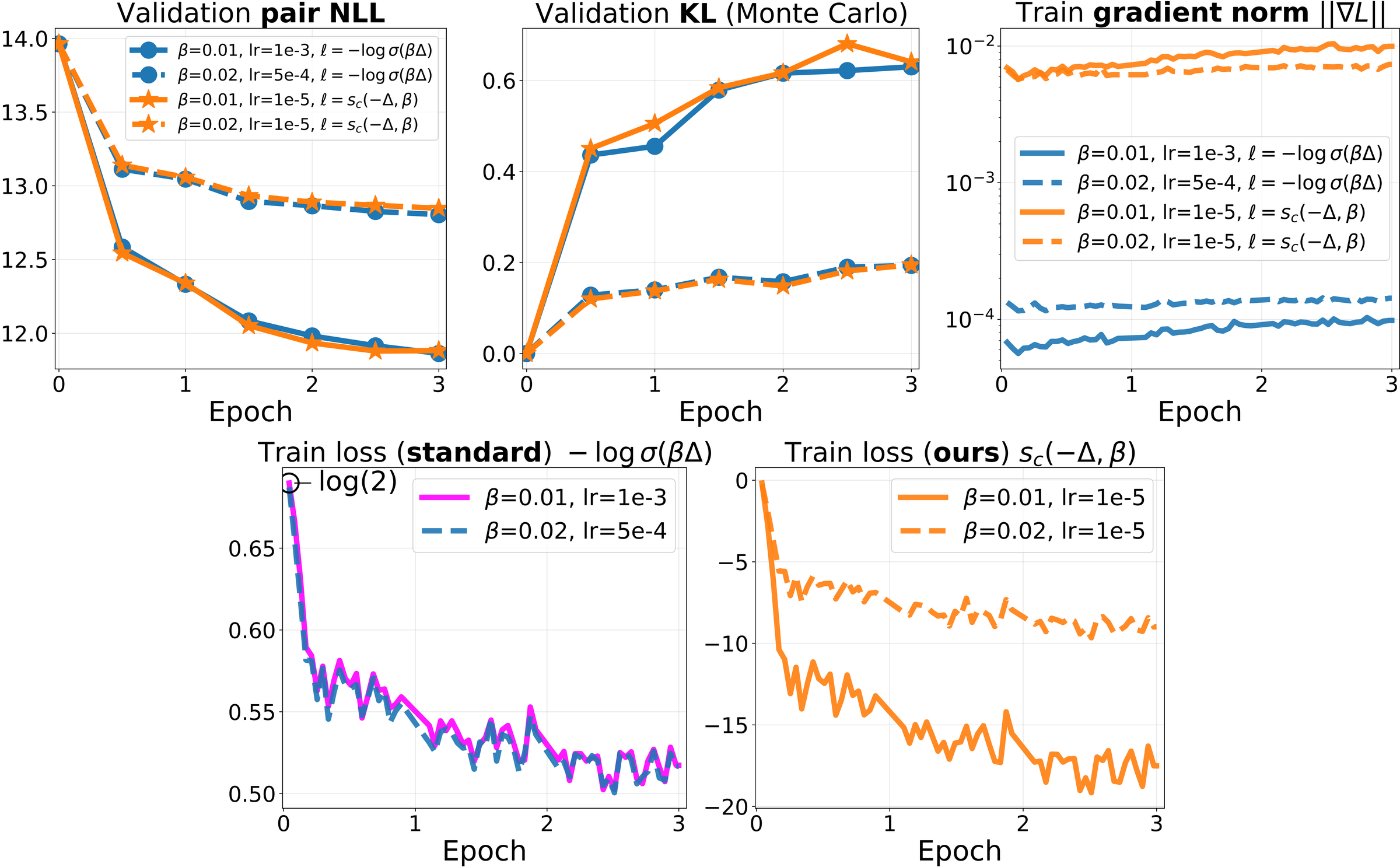}
  \caption{Comparison of training with the normalized centered-softplus DPO objective $s_c(-\Delta;\beta)$ and standard DPO with pairwise loss $-\log\sigma(\beta\Delta)$ on UltraFeedback with Qwen3-4B-Instruct-2507 (batch size 26). Learning rates satisfy $\mathrm{lr}_{\mathrm{norm}}=\beta\,\mathrm{lr}_{\mathrm{standard}}$. With SGD, validation metrics are nearly equivalent. Even when validation metrics differ substantially between $\beta\in\{0.01,0.02\}$, the standard training-loss curves are almost overlapping, whereas for $s_c(-\Delta;\beta)$ the training losses separate noticeably (Appendix~\ref{app:loss-invariance}). Single-run, three epochs. The comparison tests the predicted scale equivalence; it is not a claim that the normalized objective outperforms a retuned DPO baseline.}
  \label{fig:ultrafb-comparison-3epoch}
\end{figure}

Figure~\ref{fig:cause-failures} in the main text reports the six-epoch UltraFeedback comparison of standard DPO and the centered-softplus objective (four-seed mean $\pm$ one standard deviation).
Figure~\ref{fig:ultrafb-comparison-3epoch} shows the corresponding three-epoch single-run trajectories.

\paragraph{Downstream evaluation on UltraFeedback.}
Table~\ref{tab:ultrafb-alpaca} reports AlpacaEval~2 and IFEval scores for the four curves of Figure~\ref{fig:cause-failures} (Qwen3-4B-Instruct-2507, SGD, batch size 26, four paired seeds).
We use the same protocol as Table~\ref{tab:hsteer-benchmarks}: official AlpacaEval~2 instructions and GPT-4-turbo reference outputs, scored by a fixed \texttt{Qwen/Qwen2.5-14B-Instruct} judge, and the official verifiable IFEval benchmark without an LLM judge.
Entries are mean $\pm$ std.
The two standard-DPO configurations have nearly identical training and validation DPO losses (Figure~\ref{fig:cause-failures}), yet their length-controlled win rates differ by about $14$ percentage points and their IFEval accuracies by about $2$ points in the same direction; the centered-softplus pair shows the same downstream split, while each standard-DPO setting agrees with its learning-rate-rescaled centered-softplus counterpart.

\begin{table}[h]
\centering
\small
\setlength{\tabcolsep}{3pt}
\renewcommand{\arraystretch}{1.15}
\caption{AlpacaEval~2 and IFEval scores for the four UltraFeedback curves of Figure~\ref{fig:cause-failures} (Qwen3-4B-Instruct-2507, SGD, batch size 26). Centered softplus uses a shared $\mathrm{lr}_{\mathrm{norm}}=10^{-5}$, and the paired standard-DPO rates satisfy $\mathrm{lr}_{\mathrm{norm}}=\beta\,\mathrm{lr}_{\mathrm{standard}}$. AlpacaEval~2 scores use \texttt{Qwen/Qwen2.5-14B-Instruct} as the pairwise judge; IFEval uses the official verifiable benchmark. Entries are mean$_{\pm\mathrm{std}}$ over 4 seeds (\%).}
\label{tab:ultrafb-alpaca}
\begin{tabular}{@{}lcc cc cccc@{}}
\toprule
 & & & \multicolumn{2}{c}{AlpacaEval~2} & \multicolumn{4}{c}{IFEval} \\
\cmidrule(lr){4-5}\cmidrule(lr){6-9}
Objective & $\beta$ & $\mathrm{lr}$ & WR & LC WR &
\shortstack{Strict\\prompt} & \shortstack{Strict\\instr} &
\shortstack{Loose\\prompt} & \shortstack{Loose\\instr} \\
\midrule
\shortstack[l]{Standard\\DPO} & $0.01$ & $10^{-3}$ & $41.46_{\pm 2.98}$ & $48.61_{\pm 2.11}$ & $78.19_{\pm 0.40}$ & $84.98_{\pm 0.38}$ & $81.42_{\pm 0.44}$ & $87.35_{\pm 0.32}$ \\
\shortstack[l]{Standard\\DPO} & $0.02$ & $5\times10^{-4}$ & $61.43_{\pm 1.88}$ & $62.37_{\pm 1.31}$ & $79.81_{\pm 0.49}$ & $86.06_{\pm 0.44}$ & $83.50_{\pm 0.63}$ & $88.70_{\pm 0.54}$ \\
\shortstack[l]{Centered\\softplus} & $0.01$ & $10^{-5}$ & $41.74_{\pm 1.15}$ & $49.37_{\pm 2.04}$ & $77.77_{\pm 0.80}$ & $84.68_{\pm 0.45}$ & $81.05_{\pm 1.09}$ & $87.17_{\pm 0.75}$ \\
\shortstack[l]{Centered\\softplus} & $0.02$ & $10^{-5}$ & $62.48_{\pm 1.91}$ & $63.41_{\pm 1.77}$ & $79.71_{\pm 0.35}$ & $86.09_{\pm 0.17}$ & $83.46_{\pm 0.57}$ & $88.67_{\pm 0.52}$ \\
\bottomrule
\end{tabular}
\end{table}

\begin{figure}[H]
  \centering
  \includegraphics[width=0.92\linewidth]{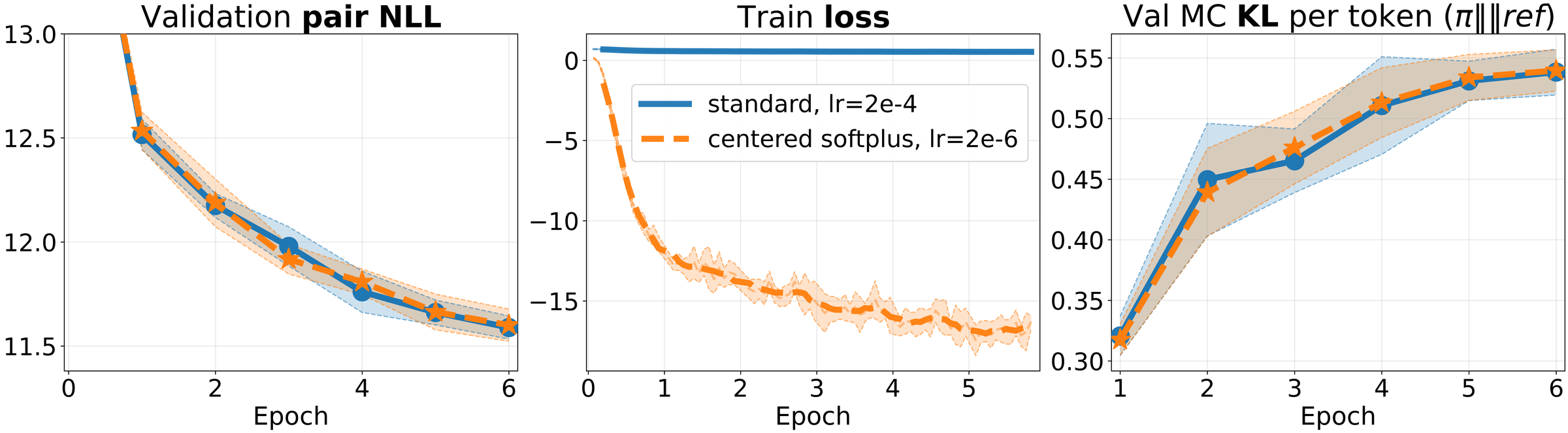}
  \caption{Comparison of training with the normalized centered-softplus DPO objective $s_c(-\Delta;\beta)$ and standard DPO with pairwise loss $-\log\sigma(\beta\Delta)$ on HelpSteer3 with Qwen3-4B-Instruct-2507 (batch size 24). SGD with $\beta=0.01$. The standard run uses $\mathrm{lr}_{\mathrm{standard}}=2\times10^{-4}$ and the centered-softplus run uses $\mathrm{lr}_{\mathrm{norm}}=\beta\,\mathrm{lr}_{\mathrm{standard}}=2\times10^{-6}$. Curves show the mean over eight experiments; shaded bands denote one standard deviation.}
  \label{fig:hsteer-sgd-centered-comparison}
\end{figure}

\paragraph{Downstream evaluation.}
Table~\ref{tab:hsteer-benchmarks} summarizes post-training instruction-following checks for the HelpSteer3 SGD comparison of Figure~\ref{fig:hsteer-sgd-centered-comparison}: standard DPO ($\beta=0.01$, $\mathrm{lr}_{\mathrm{standard}}=2\times10^{-4}$) versus the centered-softplus objective with $\mathrm{lr}_{\mathrm{norm}}=\beta\,\mathrm{lr}_{\mathrm{standard}}=2\times10^{-6}$, across eight paired seeds (43--50).
For \emph{AlpacaEval~2} \citep{Dubois2024AlpacaEval2}, we use the official instruction set and GPT-4-turbo reference outputs, generate candidate completions from each fine-tuned checkpoint, and score pairwise preferences with a fixed \texttt{Qwen/Qwen2.5-14B-Instruct} judge; we report raw win rate, length-controlled (LC) win rate, and length-filtered win rate.
For \emph{IFEval} \citep{Zhou2023IFEval}, we use the official verifiable instruction-following benchmark (strict and loose prompt- and instruction-level accuracy) without an LLM judge.
Entries report mean $\pm$ std; the $p$ column gives two-sided Wilcoxon signed-rank test $p$-values for seed-matched pairs.
No statistically significant difference is detected at $n=8$ for any reported metric. This absence of significance does not establish practical equivalence or exclude small systematic effects.

\begin{table}[h]
\centering
\small
\caption{Post-training benchmarks for the HelpSteer3 SGD comparison of Figure~\ref{fig:hsteer-sgd-centered-comparison} (Qwen3-4B-Instruct-2507, batch size 24, $\beta=0.01$, eight paired seeds). Standard DPO uses $\mathrm{lr}_{\mathrm{standard}}=2\times10^{-4}$; centered softplus uses $\mathrm{lr}_{\mathrm{norm}}=\beta\,\mathrm{lr}_{\mathrm{standard}}=2\times10^{-6}$. AlpacaEval~2 scores use \texttt{Qwen/Qwen2.5-14B-Instruct} as the pairwise judge; IFEval uses the official verifiable benchmark. Entries are mean $\pm$ std (\%); $p$ is the two-sided Wilcoxon signed-rank test $p$-value for seed-matched pairs. No statistically significant difference is detected at $n=8$; this test does not establish practical equivalence.}
\label{tab:hsteer-benchmarks}
\begin{tabular}{lccc}
\toprule
Metric & Standard DPO & Centered softplus & $p$ \\
\midrule
Alpaca win & $54.89 \pm 2.82$ & $55.47 \pm 1.83$ & 0.94 \\
Alpaca LC win & $54.72 \pm 2.37$ & $55.50 \pm 1.44$ & 0.63 \\
Alpaca len-filtered win & $43.83 \pm 2.67$ & $45.09 \pm 1.91$ & 0.63 \\
Tie & $0.02 \pm 0.04$ & $0.02 \pm 0.04$ & 1.00 \\
Loss & $45.09 \pm 2.82$ & $44.52 \pm 1.85$ & 0.94 \\
IFEval strict prompt & $80.11 \pm 0.98$ & $81.22 \pm 0.94$ & 0.11 \\
IFEval strict instr & $86.18 \pm 0.72$ & $87.02 \pm 0.65$ & 0.08 \\
IFEval loose prompt & $83.76 \pm 0.74$ & $84.61 \pm 1.27$ & 0.23 \\
IFEval loose instr & $88.74 \pm 0.48$ & $89.44 \pm 0.81$ & 0.16 \\
\bottomrule
\end{tabular}
\end{table}

\begin{figure}[H]
  \centering
  \includegraphics[width=0.92\linewidth]{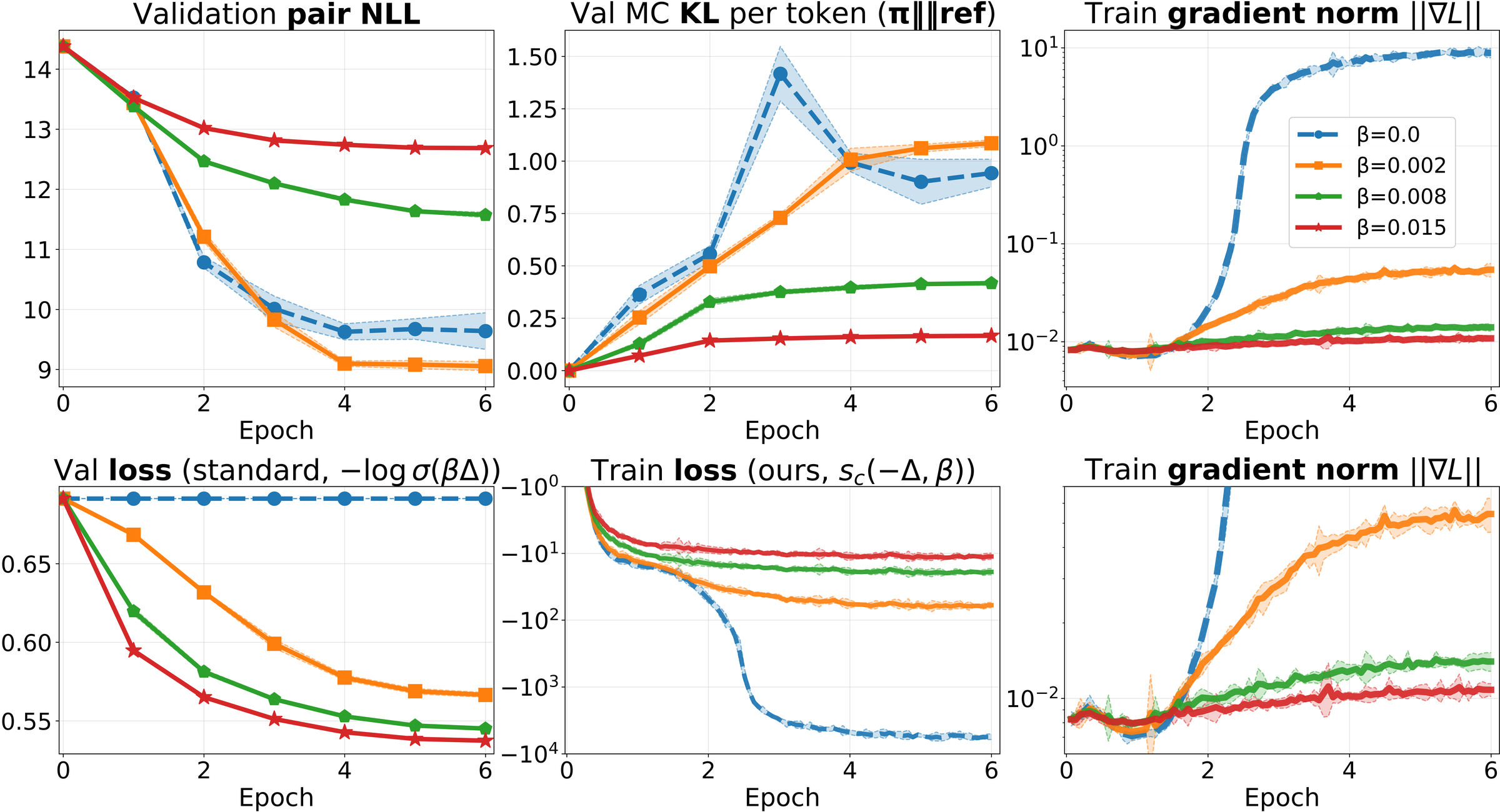}
  \caption{Training and validation dynamics on HelpSteer3 with Qwen3-4B-Instruct-2507 using the centered-softplus DPO objective $s_c(-\Delta,\beta)$ (batch size 24, $\mathrm{lr}=1\times10^{-6}$, Adam). Curves show the mean over four experiments; shaded bands denote one standard deviation.}
  \label{fig:hsteer-adam}
\end{figure}

\begin{figure}[H]
  \centering
  \includegraphics[width=0.92\linewidth]{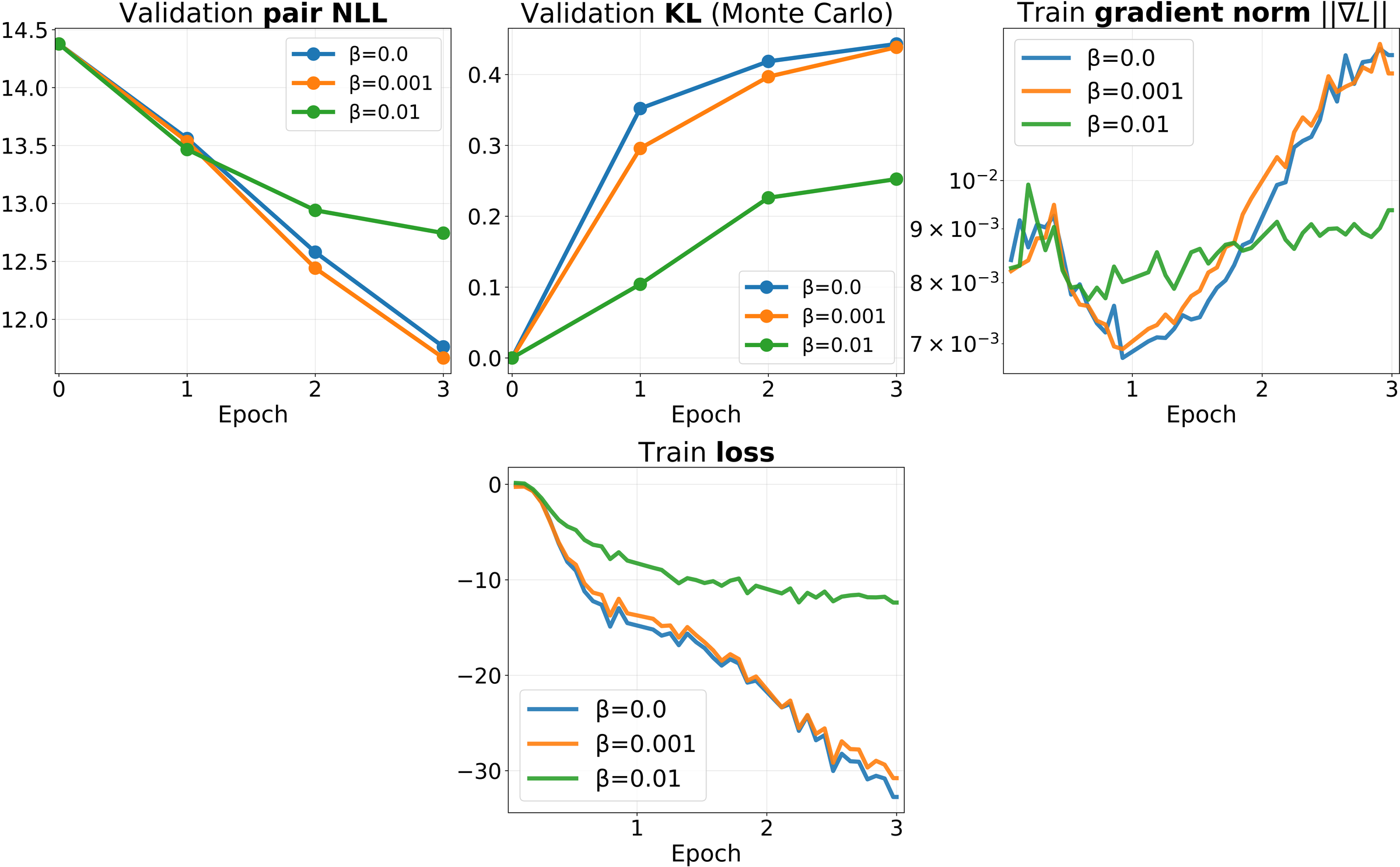}
  \caption{Centered Softplus loss $s_c(-\Delta,\beta)$ on HelpSteer3 with Qwen3-4B-Instruct-2507 (batch size 24, $\mathrm{lr}=1\times10^{-6}$), trained with Adam.}
  \label{fig:hsteer-adam-single}
\end{figure}

Figure~\ref{fig:hsteer-adam} shows the seed-averaged HelpSteer3 AdamW trajectories for the centered-softplus objective analyzed in Section~\ref{sec:normalized}; Figure~\ref{fig:hsteer-adam-single} shows the corresponding single-run trajectories.
Consistent with the $1/\beta$ readout in Appendix~\ref{app:loss-invariance}, the normalized training losses separate across $\beta$, tracking the physical aggressiveness of each run rather than collapsing onto a common curve.

\section{Logged Gradients, Mean-Abs Magnitudes, and the Jacobian Factor \texorpdfstring{$g$}{g}}
\label{app:grad-decomposition}

\setcounter{equation}{0}
\renewcommand{\theequation}{J.\arabic{equation}}
\renewcommand{\theHequation}{\theequation}

This appendix collects the gradient questions that arise in the main text: what the training logs record, how that scalar relates to $\nabla_{\theta}\mathcal{L}$, and how both relate to the Jacobian scale used in the saturation and horizon formulas.
The exact identities below first apply to a single preference pair. For a minibatch, the logged norm identifies a coherence-weighted norm of the batch-averaged margin Jacobian rather than a typical per-pair Jacobian; we make this distinction explicit before using the logged value as an order-of-magnitude proxy in Appendix~\ref{app:right-dead-zone}.

\paragraph{Chain rule.}
For a single preference pair the standard DPO loss depends on $\theta$ only through the scalar margin $\Delta(\theta)$:
\begin{equation}
\label{eq:app-I-loss}
\mathcal{L}(\theta)=\ell\bigl(\Delta(\theta)\bigr),
\qquad
\ell(u)=-\log\sigma(\beta u),
\qquad
\ell'(u)=-\beta\,\sigma(-\beta u).
\end{equation}
The parameter gradient is therefore exactly
\begin{equation}
\label{eq:app-I-chain}
\nabla_{\theta}\mathcal{L}
=
\ell'(\Delta)\,
\nabla_{\theta}\Delta
=
-\beta\,\sigma(-\beta\Delta)\,
\nabla_{\theta}\Delta.
\end{equation}
Taking Euclidean norms,
\begin{equation}
\label{eq:app-I-decomp}
\|\nabla_{\theta}\mathcal{L}\|_{2}
=
\beta\,\sigma(-\beta\Delta)\,
g,
\qquad
g
:=
\|\nabla_{\theta}\Delta\|_{2}.
\end{equation}
Thus every loss-gradient magnitude factors as
\[
\underbrace{\beta}_{\text{explicit scale}}
\times
\underbrace{\sigma(-\beta\Delta)}_{\text{sigmoid / saturation}}
\times
\underbrace{g}_{\text{Jacobian of the margin}}.
\]
At initialization $\Delta=0$ and $\sigma(0)=1/2$, so
\begin{equation}
\label{eq:app-I-init}
\|\nabla_{\theta}\mathcal{L}\|_{2}
=
\frac{\beta}{2}\,g
\qquad\Rightarrow\qquad
g
=
\frac{2}{\beta}\,\|\nabla_{\theta}\mathcal{L}\|_{2}.
\end{equation}
Equation~\eqref{eq:app-I-decomp} is an identity for one pair. Treating $g$ as constant along a trajectory and identifying it from a minibatch-gradient log are separate approximations used only for the rough horizon estimate.

\paragraph{Minibatch correction.}
For a minibatch of $B$ pairs, write $j_i=\nabla_\theta\Delta_i$ and $a_i=\sigma(-\beta\Delta_i)$. Then
\begin{equation}
\label{eq:app-I-batch-gradient}
\nabla_\theta\mathcal{L}_{B}
=
-\frac{\beta}{B}\sum_{i=1}^{B}a_i j_i.
\end{equation}
Consequently, a global gradient norm does not recover a typical
$g_i=\|j_i\|_2$. At initialization, where $a_i=1/2$,
\begin{equation}
\label{eq:app-I-batch-proxy}
\frac{2}{\beta}\|\nabla_\theta\mathcal{L}_{B}\|_2
=
\left\|\frac{1}{B}\sum_{i=1}^{B}j_i\right\|_2
=:g_B.
\end{equation}
The quantity $g_B$ includes cross-pair alignment and cancellation. It may therefore differ substantially from both
$B^{-1}\sum_i\|j_i\|_2$ and a typical per-pair norm. The Gram-matrix form of the resulting margin drift is given in Appendix~\ref{app:right-dead-zone}; throughout the numerical horizon estimate we use the observed $g_B$ only as an effective scalar proxy.

\paragraph{Logged scalars: mean-abs versus Euclidean.}
Our DPO implementation logs two scalars every $100$ steps over trainable (LoRA) coordinates $a=\nabla_{\theta}\mathcal{L}$:
\begin{align}
G
&:=
\mathrm{mean}_{j}|a_{j}|
=
\texttt{grad\_abs\_mean},
\label{eq:app-I-G}
\\
\|\nabla_{\theta}\mathcal{L}\|_{2}
&:=
\sqrt{\sum_{j}a_{j}^{2}}
=
\texttt{grad\_norm}.
\label{eq:app-I-L2}
\end{align}
TRL \texttt{DPOTrainer} TensorBoard curves report the Euclidean quantity as \texttt{train/grad\_norm}.
The two are related by a shape factor
\begin{equation}
\label{eq:app-I-r}
\|\nabla_{\theta}\mathcal{L}\|_{2}
=
C\,G,
\qquad
C=\sqrt{d}\,r,
\qquad
r
=
\frac{\mathrm{RMS}_{j}|a_{j}|}{\mathrm{mean}_{j}|a_{j}|}\ge 1,
\end{equation}
with $d$ the number of trainable parameters.
Equality $r=1$ holds only if all $|a_{j}|$ are equal; empirically, for Qwen3-4B, $r\approx 2.8$--$3.1$ on HelpSteer3 and $r\approx 3.7$--$4.1$ on UltraFeedback (below).
Table~\ref{tab:init-grad-linearity} in the main text uses $G_0$ in the sense of~\eqref{eq:app-I-G}.

The dynamics panels in the main text and appendices inherit a mixed convention from the original experimental logs: the SGD runs of our implementation recorded $G$, and those curves were kept so that the published figures match the logs, even where an axis is labelled $\|\nabla L\|$.
TRL panels report Euclidean $\|\nabla_{\theta}\mathcal{L}\|_{2}$.
This is a residual inconsistency of the logging pipeline, not a claim that $G$ equals the Euclidean norm.
By~\eqref{eq:app-I-r} the two differ by the nearly constant factor $C=\sqrt{d}\,r$ ($r$ varies only weakly along a run), so cross-$\beta$ orderings and time trends are the same for either scalar.
Horizon formulas use the Euclidean reading, not $G$. For actual minibatch logs, Equation~\eqref{eq:app-I-batch-proxy}, rather than the one-pair inversion in~\eqref{eq:app-I-init}, is the appropriate interpretation.

\begin{figure}[t]
\centering
\includegraphics[width=\textwidth]{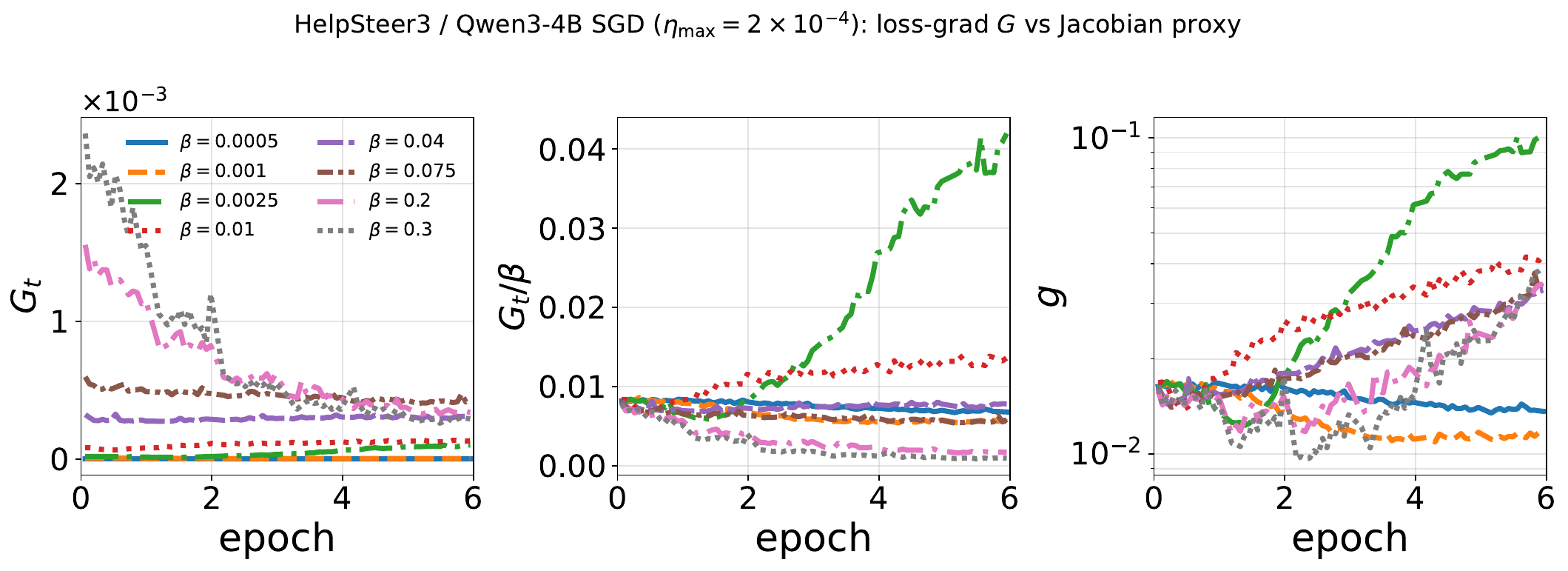}
\caption{HelpSteer3 / Qwen3-4B, our DPO implementation, SGD, $\eta_{\max}=2\times10^{-4}$.
The three panels isolate the factors in~\eqref{eq:app-I-decomp}: logged $G=\texttt{grad\_abs\_mean}$ (left), $G/\beta$ (center), and a scalar batch-Jacobian proxy $g_B(t)$ after dividing out $\beta\sigma(-\beta\Delta)$ (right). The last quantity is not a typical per-pair Jacobian norm.}
\label{fig:app-I-hsteer}
\end{figure}

\begin{figure}[t]
\centering
\includegraphics[width=\textwidth]{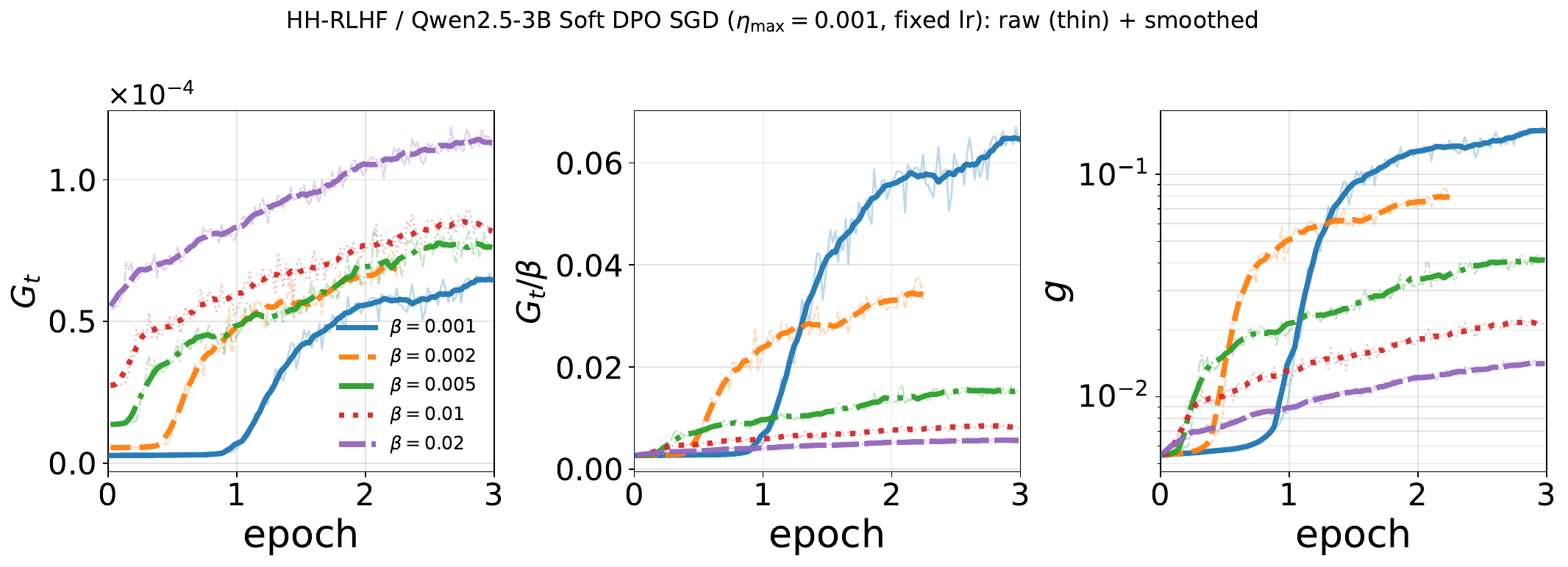}
\caption{HH-RLHF / Qwen2.5-3B, our DPO implementation, SGD, $\eta_{\max}=10^{-3}$ (fixed across $\beta$).
Same three-factor layout as Figure~\ref{fig:app-I-hsteer}.
Thin curves: raw; thick: moving average.
Growth of the reconstructed batch-Jacobian proxy is stronger than on HelpSteer3 across the sweep.}
\label{fig:app-I-hhrlhf}
\end{figure}

\begin{figure}[t]
\centering
\includegraphics[width=\textwidth]{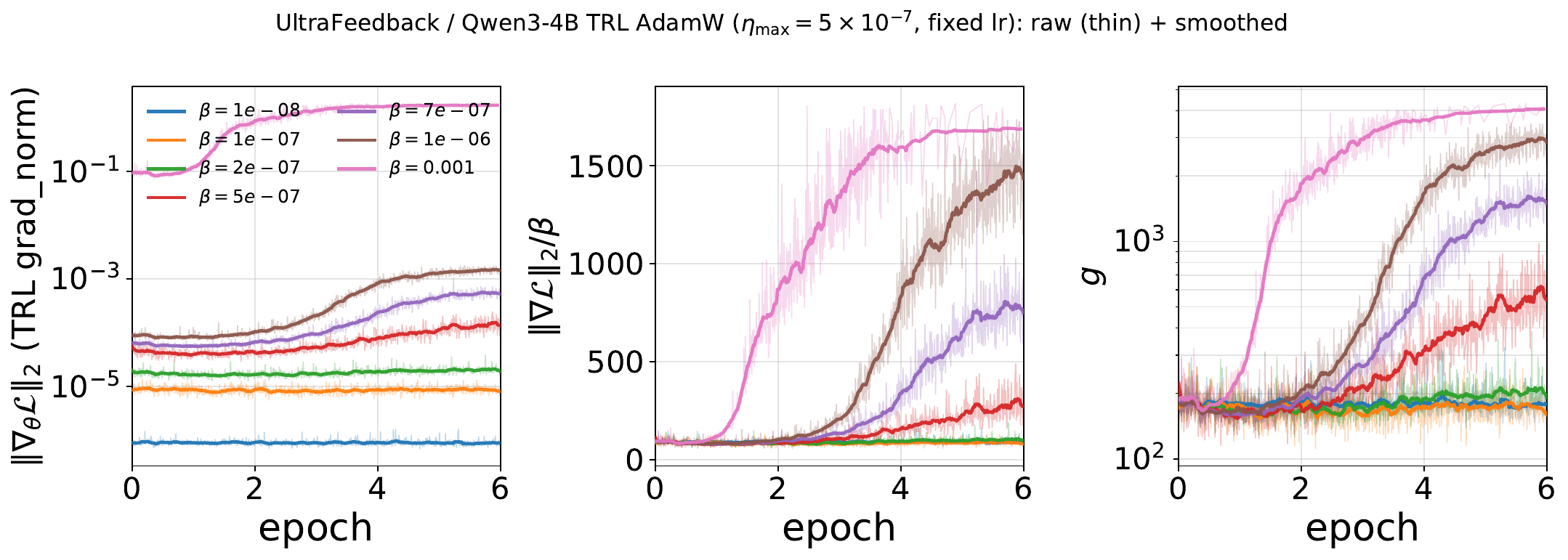}
\caption{UltraFeedback / Qwen3-4B, TRL \texttt{DPOTrainer}, AdamW, $\eta_{\max}=5\times10^{-7}$.
Left: Euclidean $\|\nabla_{\theta}\mathcal{L}\|_{2}$ from \texttt{train/grad\_norm}; center and right apply the same decomposition as Figure~\ref{fig:app-I-hsteer}.
Absolute scale is not comparable to the SGD figures above (different optimizer and $\eta$).}
\label{fig:app-I-ultrafb}
\end{figure}

\paragraph{Three-factor decomposition along training.}
Equation~\eqref{eq:app-I-decomp} splits every loss-gradient magnitude into an explicit $\beta$ scale, a sigmoid/saturation factor, and the Jacobian $g$.
Figures~\ref{fig:app-I-hsteer}--\ref{fig:app-I-ultrafb} isolate those three factors along a run:
\begin{enumerate}
\item left: the logged loss-gradient magnitude ($G$ for our implementation; $\|\nabla_{\theta}\mathcal{L}\|_{2}$ for TRL);
\item center: the same quantity divided by $\beta$, which removes the explicit scale (at $t=0$ the curves collapse);
\item right: division by $\beta\,\sigma(-\beta\Delta_{t})$ as well, leaving a scalar proxy for $g_B(t)$
(using a linear interpolation of the empirical final mean validation margin for runs of our implementation, and the reported mean $|\Delta|$ for TRL).
\end{enumerate}
A drop on the left panel need not mean that $g$ shrinks: under~\eqref{eq:app-I-decomp} the sigmoid factor alone drives the loss gradient down as $\Delta$ grows at fixed $g$.
The right panel is the diagnostic for whether $g$ itself is moving.

\paragraph{HelpSteer3 numerical check (our implementation, SGD, $\eta_{\max}=2\times10^{-4}$).}
Two seed-$43$ runs log both $G$ and $\|\nabla\mathcal{L}\|_{2}$ with $d=33\,030\,144$ LoRA parameters.
Early steps ($100$--$500$):
\begin{center}
\small
\begin{tabular}{lccc}
\toprule
$\beta$ & $G/\beta$ & $r=\|\nabla\mathcal{L}\|_{2}/(\sqrt{d}\,G)$ & $g_B=2\|\nabla\mathcal{L}\|_{2}/\beta$ \\
\midrule
$0.0025$ (peak) & $8.24\times10^{-3}$ & $2.76$ & $261$ \\
$0.075$ (right slope) & $7.17\times10^{-3}$ & $2.94$ & $242$ \\
\bottomrule
\end{tabular}
\end{center}
So $G/\beta$ is essentially the same as in Table~\ref{tab:init-grad-linearity}, while the Euclidean reading gives $r\approx 2.8$--$2.9$ (within the HelpSteer3 / Qwen3-4B range $2.8$--$3.1$) and the batch proxy $g_B\approx 250$, not the isotropic $r=1$ value $g_B\approx 95$.
The isotropic choice $r=1$ is a lower bound on $r$ (hence an upper bound on the scalar horizon proxy $\beta_{\mathrm{r}}\propto 1/g_B^{2}\propto 1/r^{2}$): it would give $g_B\approx 95$ and $\beta_{\mathrm{r}}\approx 0.19$, whereas the measured L2 norm gives $g_B\approx 250$ and $\beta_{\mathrm{r}}\approx 0.027$ (Appendix~\ref{app:right-dead-zone}). Neither value estimates a typical per-pair Jacobian.
By the end of epoch~1, with validation mean $\Delta\approx 20.9$ ($\beta=0.0025$) and $\Delta\approx 13.3$ ($\beta=0.075$),
\[
g_B(t)
=
\frac{\|\nabla\mathcal{L}\|_{2}}{\beta\,\sigma(-\beta\Delta)}
\]
has already risen to $\approx 360$ and $\approx 370$ respectively under the scalar reconstruction: the effective batch proxy grows, while at $\beta=0.075$ the raw $G$ falls mainly because $\sigma(-\beta\Delta)$ shrinks.
This reconstruction additionally substitutes an aggregate validation margin into the sigmoid, so it is diagnostic rather than an exact measurement. Freezing the initial proxy ignores this time variation.

\paragraph{Takeaways for the rest of the paper.}
\begin{itemize}
\item Main-text $G_0$ and the dynamics figures from our implementation use mean-abs $G$, including some panels labelled $\|\nabla L\|$; this mixed labelling is a legacy of those logs, as above.
Euclidean $\|\nabla_{\theta}\mathcal{L}\|_{2}$ differs from $G$ by a near-constant factor $C=\sqrt{d}\,r$ (empirically, for Qwen3-4B, $r\approx 2.8$--$3.1$ on HelpSteer3 and $r\approx 3.7$--$4.1$ on UltraFeedback), so the two are interchangeable for cross-$\beta$ comparisons.
\item The scalar horizon formula uses the logged batch quantity $g_B$ as a proxy for the one-pair $g$ in~\eqref{eq:app-I-decomp}. An exact minibatch calculation instead requires cross-pair Gram products.
\item Because $\beta_{\mathrm{r}}\propto 1/g_B^{2}$ in the scalar closure, time variation and cross-pair cancellation preclude a precise cutoff. The resulting $\beta_{\mathrm{r}}$ is reported only to order of magnitude; factor-of-few shifts occupy less than one decade on the logarithmic $\beta$ axis used in the sweep.
\end{itemize}

\section{Time to Saturation and the Right Cutoff \texorpdfstring{$\beta_{\mathrm{r}}$}{beta\_r}}
\label{app:right-dead-zone}

\setcounter{equation}{0}
\renewcommand{\theequation}{K.\arabic{equation}}
\renewcommand{\theHequation}{\theequation}

Appendix~\ref{app:saturation} gives the one-pair saturation margin $\Delta^{*}_{\mathrm{DPO}}(\beta)$ and the \emph{left} dead zone $\beta\le 2\tau$.
On the right slope, $\Delta^{*}_{\mathrm{DPO}}\sim\log(\beta/\tau)/\beta\to 0$, but that asymptotic alone does not say at which $\beta$ a finite training budget ceases to move the policy.
This appendix seeks only an order-of-magnitude horizon scale $\beta_{\mathrm{r}}$, represented by the broad right-hand band of Figure~\ref{fig:beta-peak-nll-kl-delta}, rather than a calibrated cutoff.
Two approximations dominate its uncertainty. First, the scalar closure freezes the initial Jacobian proxy even though the reconstructed proxy usually grows during training. Second, the global minibatch gradient identifies the norm of a batch-averaged Jacobian, including cross-pair cancellation, rather than a typical per-pair norm. An exact calculation would retain the full cross-pair Gram matrix. We state that correction below, then deliberately keep the scalar closure because the goal is a rough location on a logarithmic $\beta$ axis.
Notation for the loss gradient, the one-pair Jacobian $g=\|\nabla_{\theta}\Delta\|_{2}$, and the effective batch proxy $g_B$ is fixed in Appendix~\ref{app:grad-decomposition}.

\paragraph{Inputs already fixed elsewhere.}
We reuse the one-pair saturation ceiling $\Delta^{*}_{\mathrm{DPO}}(\beta)$ and peak location $\beta_{\mathrm{peak}}\approx 4.59\,\tau$ from Appendix~\ref{app:saturation} (Equation~\eqref{eq:sat-margin-dpo}) without re-deriving them.
For the HelpSteer3 / Qwen3-4B standard-DPO SGD sweep we take the shape+scale estimate $\tau=6.37\times 10^{-4}$ from Appendix~\ref{app:tau-estimators}; the rough $\beta_{\mathrm{r}}$ estimate below additionally folds in the learning-rate budget and the initial proxy $g_B$.

\paragraph{Learning-rate budget.}
Write $\eta_{t}$ for the per-step learning rate, $T_{\mathrm{budget}}$ for the planned number of SGD steps, and
\begin{equation}
\label{eq:rdz-H}
H
:=
\sum_{t=0}^{T_{\mathrm{budget}}-1}\eta_{t},
\qquad
\eta_{\mathrm{eff}}
:=
H/T_{\mathrm{budget}}.
\end{equation}
For a cosine (or triangular) schedule that rises from $0$ to $\eta_{\max}$ and returns to $0$,
$H\approx\eta_{\max}T_{\mathrm{budget}}/2$.

\subsection{From a parameter step to an equation on the margin}
\label{app:rdz-ode}

The DPO loss on a single pair depends on $\theta$ only through $\Delta(\theta)$.
Appendix~\ref{app:grad-decomposition} records the chain rule
\begin{equation}
\label{eq:rdz-chain}
\nabla_{\theta}\mathcal{L}
=
\ell'(\Delta)\,
\nabla_{\theta}\Delta
=
-\beta\,\sigma(-\beta\Delta)\,
\nabla_{\theta}\Delta,
\qquad
g(\theta)
:=
\|\nabla_{\theta}\Delta(\theta)\|_{2},
\end{equation}
so $\|\nabla_{\theta}\mathcal{L}\|_{2}=\beta\,\sigma(-\beta\Delta)\,g$.
An SGD step $\theta_{t+1}=\theta_{t}-\eta_{t}\nabla_{\theta}\mathcal{L}(\theta_{t})$ therefore changes the scalar margin by a first-order Taylor expansion
\begin{equation}
\label{eq:rdz-taylor}
\Delta(\theta_{t+1})-\Delta(\theta_{t})
=
\nabla_{\theta}\Delta(\theta_{t})^{\top}
(\theta_{t+1}-\theta_{t})
+O(\|\theta_{t+1}-\theta_{t}\|^{2}).
\end{equation}
Substituting the update and~\eqref{eq:rdz-chain} gives the discrete drift
\begin{equation}
\label{eq:rdz-disc}
\Delta_{t+1}-\Delta_{t}
=
\eta_{t}\,g(\theta_{t})^{2}\,\beta\,\sigma(-\beta\Delta_{t})
+O(\eta_{t}^{2}),
\end{equation}
where $\Delta_{t}:=\Delta(\theta_{t})$.
The factor $g^{2}$ appears because the SGD step is parallel to $\nabla_{\theta}\Delta$, so the induced change in the scalar margin is $\nabla_{\theta}\Delta^{\top}(\eta|\ell'|\,\nabla_{\theta}\Delta)=\eta|\ell'|\,\|\nabla_{\theta}\Delta\|_{2}^{2}$; it is the squared Jacobian of $\Delta$, not $(\partial\mathcal{L}/\partial\Delta)^{2}$.
Identifying one step with $dt=1$ yields the mean-field ODE
\begin{equation}
\label{eq:rdz-ode}
\dot\Delta(t)
=
\eta(t)\,g\bigl(\theta(t)\bigr)^{2}\,
\beta\,\sigma\bigl(-\beta\Delta(t)\bigr).
\end{equation}
Here $\beta$ enters explicitly through the sigmoid; $g$ enters through $\theta(t)$, and the path $\theta(t)$ itself depends on $\beta$, so logged $g(t)$ need not be the same across a $\beta$-sweep (Appendix~\ref{app:grad-decomposition}, Figures~\ref{fig:app-I-hsteer}--\ref{fig:app-I-ultrafb}).

\paragraph{Exact minibatch correction.}
For a training batch, let $j_i=\nabla_\theta\Delta_i$, $a_i=\sigma(-\beta\Delta_i)$, and
$K_{ki}=j_k^\top j_i$. The first-order drift of pair $k$ is
\begin{equation}
\label{eq:rdz-gram}
\Delta_{k,t+1}-\Delta_{k,t}
=
\frac{\eta_t\beta}{B}\sum_{i=1}^{B}a_iK_{ki}
+O(\eta_t^2).
\end{equation}
Thus the exact dynamics depend on cross-pair alignment, not only on $\|j_k\|_2^2$. For a mean over $M$ probe pairs, the corresponding drift is
\[
\overline{\Delta}_{t+1}-\overline{\Delta}_{t}
=
\frac{\eta_t\beta}{MB}\,\mathbf{1}^{\top}K\mathbf{a}
+O(\eta_t^2),
\]
where $K$ may be the cross-Gram matrix between probe and training pairs. A precise horizon estimate would evaluate or approximate these Gram contractions along the trajectory and integrate them under the actual learning-rate schedule.

\paragraph{Frozen-Jacobian closure.}
For a deliberately simpler order-of-magnitude estimate, we return to the one-dimensional closure and freeze $g(\theta(t))\equiv g:=g(\theta_0)$. Numerically, $g$ is replaced by the initial effective batch proxy $g_B$ from Equation~\eqref{eq:app-I-batch-proxy}; this does not turn it into a typical per-pair Jacobian.
Then~\eqref{eq:rdz-ode} separates:
\begin{equation}
\label{eq:rdz-sep}
\bigl(1+e^{\beta\Delta}\bigr)\,d\Delta
=
\eta\,g^{2}\,\beta\,dt.
\end{equation}
With a variable schedule the right-hand side integrates to $g^{2}\beta\,H$.
If $g$ is not frozen, the same integral becomes $I=\int\eta(t)g(\theta(t))^{2}\,dt$ in place of $g^{2}H$.

\paragraph{Stochasticity.}
Equation~\eqref{eq:rdz-ode} is a scalar mean-field closure of the minibatch recursion~\eqref{eq:rdz-gram}; random batch sampling adds stochastic variation around the corresponding Gram-weighted drift.
Appendix~\ref{app:margin-diffusion} gives an optional stochastic interpretation of this variation and describes how an effective drift--noise crossover could be estimated from windowed margin trajectories. It does not assume convergence to an equilibrium process.

\subsection{Time to saturation and definition of \texorpdfstring{$\beta_{\mathrm{r}}$}{beta\_r}}
\label{app:rdz-beta-r}

Integrating~\eqref{eq:rdz-sep} from $\Delta(0)=0$ to $\Delta(T)=\Delta^{*}_{\mathrm{DPO}}(\beta)$ gives
\begin{equation}
\label{eq:rdz-int}
\Delta^{*}+\frac{e^{\beta\Delta^{*}}-1}{\beta}
=
\eta g^{2}\beta\,T
\end{equation}
for constant $\eta$ (replace $\eta T$ by $H$ under a schedule).
From Equation~\eqref{eq:sat-margin-dpo}, $e^{\beta\Delta^{*}}=\beta/\tau-1$ whenever $\beta>2\tau$, so
\begin{equation}
\label{eq:rdz-T}
T_{\mathrm{DPO}}(\beta)
=
\frac{1}{\eta g^{2}\beta^{2}}
\left[
\log\!\left(\frac{\beta}{\tau}-1\right)
+\frac{\beta}{\tau}-2
\right]
\;\xrightarrow[\beta\gg\tau]{}\;
\frac{1}{\eta g^{2}\beta\tau}.
\end{equation}
Thus $T_{\mathrm{DPO}}\to 0$ at both ends of the admissible interval: $\Delta^{*}$ collapses as $\beta\to(2\tau)^{+}$ and as $\beta\to\infty$.
The leading right-branch condition $T_{\mathrm{DPO}}\lesssim T_{\mathrm{budget}}$ rearranges to
\begin{equation}
\label{eq:rdz-beta-r}
\beta_{\mathrm{r}}
\;:=\;
\frac{1}{\eta_{\mathrm{eff}}\,g^{2}\,\tau\,T_{\mathrm{budget}}}
=
\frac{1}{g^{2}\,\tau\,H}.
\end{equation}
For $\beta\gtrsim\beta_{\mathrm{r}}$ the budget can still formally reach $\Delta^{*}(\beta)$, but that ceiling itself is already small, so policy movement is practically negligible.
Equation~\eqref{eq:rdz-beta-r} is therefore a characteristic scale for the right cutoff, not a sharp threshold on the $\beta$-axis. Under minibatch training it is a scalar proxy for the Gram dynamics in Equation~\eqref{eq:rdz-gram}, so we retain only its order of magnitude.

\subsection{Jacobian input}
\label{app:rdz-read-g}

We take the effective proxy $g_B(\theta_0)$ from Appendix~\ref{app:grad-decomposition}: at initialization,
$g_B(\theta_0)=2\|\nabla_{\theta}\mathcal{L}_{B}(\theta_0)\|_{2}/\beta$, with the Euclidean minibatch-loss gradient (not mean-abs~$G$).
This is the norm of the batch-averaged Jacobian, not a typical $\|\nabla_\theta\Delta_i\|_2$.

\subsection{Numerical estimate on HelpSteer3}
\label{app:rdz-numeric}

\paragraph{Protocol and inputs.}
Standard DPO on HelpSteer3 / Qwen3-4B-Instruct-2507 with SGD, batch size~$24$, six epochs (same sweep as Figure~\ref{fig:beta-peak-nll-kl-delta} and Table~\ref{tab:app-nll-kl-delta}):
$T_{\mathrm{budget}}=9078$, $\eta_{\max}=2\times10^{-4}$, hence $H=\eta_{\max}T_{\mathrm{budget}}/2=0.9078$,
and $\tau=6.37\times10^{-4}$ as above.
Appendix~\ref{app:grad-decomposition} supplies the initial effective batch proxy $g_B(\theta_0)\approx 250$ from the measured Euclidean loss gradient. We substitute it for $g$ in the scalar closure.
Substituting into~\eqref{eq:rdz-beta-r} yields
\begin{equation}
\label{eq:rdz-result-L2}
\beta_{\mathrm{r}}
\approx 0.027.
\end{equation}
The reconstructed proxy reaches roughly $360$--$370$ later in training. Holding the other inputs fixed, substituting that scale would move $\beta_{\mathrm{r}}$ downward by about a factor of two, from $0.027$ to roughly $0.012$--$0.013$. Cross-pair cancellation introduces an additional, unquantified difference between $g_B$ and a typical per-pair Jacobian. These factor-of-few shifts occupy less than one decade on the logarithmic sweep and are why we report a broad order-of-magnitude region rather than a precise boundary.

\begin{table}[t]
\centering
\small
\setlength{\tabcolsep}{4pt}
\caption{HelpSteer3 / Qwen3-4B standard DPO (SGD): right slope of the sweep in Figure~\ref{fig:beta-peak-nll-kl-delta} and Table~\ref{tab:app-nll-kl-delta}.
Min validation pair NLL (with $\beta=1$), per-token KL, and mean validation margin; last column is the one-pair ceiling~$\Delta^{*}_{\mathrm{DPO}}(\beta)$ from Equation~\eqref{eq:sat-margin-dpo} at $\tau=6.37\times10^{-4}$.
Highlighted row: the rough scalar estimate~$\beta_{\mathrm{r}}$ from~\eqref{eq:rdz-result-L2}; NLL, $\overline{\Delta}$, and KL are interpolated log-linearly in~$\beta$ between the neighbouring measured points $0.02$ and $0.04$, while $\Delta^{*}_{\mathrm{DPO}}(0.027)$ is the closed-form ceiling. The highlighted value is an order-of-magnitude guide, not a measured or calibrated cutoff.}
\label{tab:rdz-empirical-right}
\begin{tabular}{lccccc}
\toprule
& $\beta$ & min NLL & KL / token & mean val.\ $\overline{\Delta}$ & $\Delta^{*}_{\mathrm{DPO}}(\beta)$ \\
\midrule
$\beta_{\mathrm{peak}}$ & $0.0025$ & $9.92\pm 0.14$ & $1.00\pm 0.02$ & $152\pm 6$ & $429$ \\
& $0.01$ & $11.55\pm 0.08$ & $0.540\pm 0.008$ & $75.1\pm 0.6$ & $269$ \\
& $0.02$ & $12.61\pm 0.01$ & $0.290\pm 0.024$ & $45.4\pm 0.0$ & $171$ \\
\rowcolor{gray!20}
$\beta_{\mathrm{r}}$ & $0.027$ & $\approx 12.88$ & $\approx 0.221$ & $\approx 38.5$ & $138$ \\
& $0.04$ & $13.23\pm 0.01$ & $0.130\pm 0.007$ & $29.4\pm 0.0$ & $103$ \\
& $0.075$ & $13.36\pm 0.02$ & $0.066\pm 0.001$ & $21.9\pm 0.3$ & $63$ \\
& $0.2$ & $13.67$ & $0.019$ & $14.9$ & $29$ \\
& $0.3$ & $13.80$ & $0.010$ & $12.3$ & $21$ \\
\bottomrule
\end{tabular}
\end{table}

\paragraph{Comparison with the empirical right slope.}
Table~\ref{tab:rdz-empirical-right} places~\eqref{eq:rdz-result-L2} on the same $\beta$-axis as Figure~\ref{fig:beta-peak-nll-kl-delta} (highlighted row).
The empirical entries at $\beta_{\mathrm{r}}$ (NLL~$\approx 12.88$, KL~$\approx 0.221$, $\overline{\Delta}\approx 38.5$) sit smoothly between $\beta=0.02$ and $\beta=0.04$; the $138$ in the last column is $\Delta^{*}_{\mathrm{DPO}}(0.027)$ from Equation~\eqref{eq:sat-margin-dpo}, between the neighbouring ceilings $171$ and $103$.
As throughout Appendix~\ref{app:tau-estimators}, the pair-averaged $\overline{\Delta}$ lies well below the one-pair ceiling.
Training remains clearly active at $\beta=0.075$; by $\beta=0.2$--$0.3$, pair NLL has risen most of the way back toward the dead-zone floor and per-token KL has collapsed, placing these runs on the empirical right-hand saturation slope.
Equations~\eqref{eq:rdz-T}--\eqref{eq:rdz-beta-r} provide only a rough horizon scale under the scalar frozen-proxy closure; they do not predict the vertical scale of $\overline{\Delta}(\beta)$.

\paragraph{Takeaway.}
The left dead zone $\beta\le 2\tau$ is a property of the marginal gradient at $\Delta=0$.
The scalar proxy $\beta_{\mathrm{r}}=1/(g_B^{2}\tau H)$ combines the finite horizon with a frozen effective batch-Jacobian scale.
On this HelpSteer3 SGD sweep it gives $\beta_{\mathrm{r}}\approx 0.027$, placing the right-hand crossover within the observed logarithmic range. The precise minibatch value would require the trajectory-dependent Gram contractions in Equation~\eqref{eq:rdz-gram}; therefore neither $0.027$ nor the shaded band should be read as a sharp prediction.


\section{Stochastic Dynamics of the Preference Margin}
\label{app:margin-diffusion}

\setcounter{equation}{0}
\renewcommand{\theequation}{L.\arabic{equation}}
\renewcommand{\theHequation}{\theequation}

This appendix is not required for the main results of the paper.
Its purpose is to provide additional intuition for the effective stopping tolerance $\tau$, the nonzero movement observed inside the nominal dead zone, and the transition from directed motion to noise-dominated training.
The deterministic threshold formulas in Sections~\ref{sec:loss-geometry} and Appendix~\ref{app:saturation} remain phenomenological finite-training descriptions; the stochastic view below suggests how their effective tolerance could be estimated from training dynamics without fitting the cross-$\beta$ curve of Figure~\ref{fig:beta-peak-nll-kl-delta}.

\paragraph{A coarse-grained stochastic observable.}
Let $m_t$ denote a scalar margin observable, ideally the mean margin on a fixed probe set and, in the current implementation, approximately the mean training margin aggregated over a window of steps.
At a fixed model state, minibatch sampling makes its update random.
We write the coarse-grained recursion as
\begin{equation}
\label{eq:diff-recursion}
m_{t+1}-m_t
=
d_t+e_t,
\qquad
\mathbb{E}[e_t\mid\mathcal{F}_t]=0,
\qquad
\operatorname{Var}(e_t\mid\mathcal{F}_t)=q_t^2,
\end{equation}
where $d_t$ is the local directed drift and $q_t$ is the innovation scale in units of margin per observation interval.
For standard DPO, the scalar mean-field closure of Appendices~\ref{app:saturation}--\ref{app:right-dead-zone} writes
\[
d_t
\approx
\eta_t g_{\mathrm{eff}}^2(t)\,a_t,
\qquad
a_t
:=
\beta\,\mathbb{E}_i\!\left[\sigma(-\beta\Delta_i(t))\right].
\]
The exact minibatch drift contains the Gram contractions in Equation~\eqref{eq:rdz-gram}; $g_{\mathrm{eff}}$ is only their scalar proxy.

The corresponding continuous notation is
\begin{equation}
\label{eq:diff-sde}
dm=b(m,t)\,dt+q(t)\,dW.
\end{equation}
This process has no finite equilibrium under the positive DPO drift alone.
What appears as a plateau in finite training is instead produced by a combination of sigmoid attenuation, stochastic fluctuations, learning-rate decay, and the finite horizon.
Local perturbations around the moving mean path can contract, but this local loss of memory should not be confused with convergence of the mean to a fixed point.

\paragraph{Operational noise scale and stopping tolerance.}
We define a noise-dominated crossover at a chosen observation scale by
\[
|d_t|\lesssim c\,q_t,
\]
where $c=O(1)$ specifies the desired signal-to-noise criterion.
The associated margin-update threshold is
\[
\widehat{\varepsilon}_t:=c\,q_t.
\]
In the scalar closure, $\tau=\varepsilon/(\eta g_{\mathrm{eff}}^2)$.
Because the same effective mobility satisfies
$\eta_tg_{\mathrm{eff}}^2(t)\approx d_t/a_t$, it can be eliminated:
\[
\widehat{\tau}_t
=
c\,a_t\,\frac{q_t}{|d_t|}.
\]
At the crossover time $t_*$, where $|d_{t_*}|\approx c q_{t_*}$,
\[
\widehat{\tau}
\approx
a_{t_*}
=
\beta\,\mathbb{E}_i\!\left[\sigma(-\beta\Delta_i(t_*))\right].
\]
Thus $\tau$ can in principle be estimated from one sufficiently resolved trajectory: estimate its local drift and innovation scale, locate the noise-dominated crossover, and read out the marginal-gradient factor there.
Several seeds are nevertheless preferable because they separate a common drift from stochastic variation more reliably.

\paragraph{Practical estimation from windowed logs.}
Let $W$ be the logging window in optimizer steps; in the present experiments $W=100$.
For seed $s$ and window $k$, let $X_{s,k}$ be the mean training margin aggregated over that window.
The following procedure gives a window-scale estimate:
\begin{enumerate}
\item Fit a smooth trajectory $\widehat m_{\beta}(k)$ to the aligned windows, preferably jointly across seeds.
\item Estimate the directed window drift by
$D_{\beta,k}=\widehat m_{\beta}(k+1)-\widehat m_{\beta}(k)$.
\item Form residuals $r_{s,k}=X_{s,k}-\widehat m_{\beta}(k)$ and fit locally
$r_{s,k+1}=\phi_{\beta,k}r_{s,k}+u_{s,k}$.
Use $Q_{\beta,k}=\operatorname{Std}(u_{s,k})$ as the innovation scale.
\item Locate the first sustained region, for example three consecutive windows, in which
$|D_{\beta,k}|\lesssim cQ_{\beta,k}$.
\item Compute $a_{\beta,k}$ from the per-example margins before averaging:
$a_{\beta,k}=\beta\,\operatorname{mean}_i\sigma(-\beta\Delta_i)$.
If this quantity is unavailable, $\beta\sigma(-\beta X_{s,k})$ is only a scalar approximation because sigmoid and averaging do not commute.
\item Report $\widehat{\tau}_{\beta}=a_{\beta,k_*}$ and combine the independently obtained estimates across $\beta$ and seeds, for example by a median and a bootstrap interval.
\end{enumerate}

The within-window standard deviation of the training margin is not the innovation scale $Q_{\beta,k}$ above: it mixes stochastic policy motion, deterministic trend, and changing minibatch composition.
It can still be used to weight the window means.
A cleaner estimate follows if one also records the window-endpoint margins, the mean and variance of per-step increments, their lag-one covariance, and the per-example factor $\beta\,\operatorname{mean}_i\sigma(-\beta\Delta_i)$.
Evaluating a small fixed probe set on the same grid would additionally remove minibatch-composition noise.

\paragraph{Left and right noise-dominated regimes.}
The same operational tolerance can act on both sides of the sweep.
At initialization, $a_0=\beta/2$; hence the left dead-zone condition is
$\beta/2\lesssim\tau$, so noise dominates before a directed transit develops.
For larger $\beta$, $a_t$ starts above tolerance and decreases as the margin grows, eventually crossing the same level on the right.
This does not make the dead zone absolute: stochastic updates and finite-sample effects can still produce nonzero displacement, consistent with the measured KL at $\beta=5\times10^{-4}$ in Table~\ref{tab:app-nll-kl-delta}.

The $\tau$ inferred in this way is specifically a noise-scale tolerance at the chosen temporal resolution.
The effective $\tau$ fitted in Figure~\ref{fig:beta-peak-nll-kl-delta} also absorbs finite-horizon and learning-rate-decay effects.
Agreement between the two would indicate a noise-limited regime; a systematic difference would diagnose an additional horizon-limited contribution.
If the apparent plateau occurs only after $\eta_t$ has become negligible, $\tau$ is not identifiable from that plateau alone, because any trajectory freezes when the learning rate vanishes.

\paragraph{Takeaway.}
This optional appendix is intended to clarify, rather than support, the principal analytical and empirical claims of the paper.
Its practical proposal is to treat $\tau$ as an operational drift--noise crossover: estimate the smooth margin drift and stochastic innovations from aligned trajectories, identify where noise becomes comparable to directed motion, and evaluate the DPO marginal-gradient factor at that point.
No equilibrium distribution is assumed.
The procedure can be applied to one well-resolved trajectory, while multiple seeds and a fixed probe set provide a cleaner and independently checkable estimate.

\section{Training Details and Compute Resources}
\label{app:training-details}

\paragraph{Datasets and preprocessing.}
For HelpSteer3-Preference, we keep non-zero pairwise labels and exclude indifference cases ($\mathrm{overall\_preference}=0$). For UltraFeedback Binarized, we use the provided chosen/rejected pairs directly.
For HH-RLHF, we use the PKU-Alignment processed release with explicit context/chosen/rejected fields \citep{PKUProcessedHHRLHF} and cite the original Anthropic corpus \citep{AnthropicHHRLHF,HHRLHF2022}.
Prior data-centric and reward-modeling analyses report that HH-RLHF is comparatively noisy, less informative, and prone to weak or ambiguous pairwise margins \citep{Shen2024DataCentricRLHF,Wang2024SecretsRLHFPartII}; we discuss the resulting training dynamics in Appendix~\ref{app:hh-rlhf-dynamics}.
This follows common preference-learning practice that emphasizes unambiguous winner/loser supervision, including strong-preference filtering protocols such as SHP \citep{Ethayarajh2022SHP}.

\paragraph{Splits, sequence lengths, and training times.}
Dataset splits are as follows: HelpSteer3-Preference has 36,299/1,920 training/validation pairs, UltraFeedback has 61,054/1,997, and HH-RLHF has 159,700/8,492. For all our experiments, the input context is constrained to 768 prompt tokens, with a 1,536-token cap on the full prompt-plus-response sequence. Typical one-epoch training times are 12.4 h for HelpSteer3 with Qwen3-4B-Instruct-2507, 15.5 h for UltraFeedback Binarized with Qwen3-4B-Instruct-2507, and 19.9 h for HH-RLHF with Qwen2.5-3B-Instruct. Depending on the experiment, runs are trained for three or six epochs (see figure captions and per-run configurations); for UltraFeedback and HH-RLHF, validation is also performed at mid-epoch.

\paragraph{Mamba-2 backbone.}
For HH-RLHF experiments with a Mamba-2 backbone (including the TRL baseline runs in Appendix~\ref{app:hh-rlhf-dynamics} and the $\beta_{\mathrm{crit}}$ analysis in Appendix~\ref{app:adam-threshold}), we use the public Hugging Face checkpoint \texttt{AntonV/mamba2-2.7b-hf}~\citep{Mamba227BHf}. The full model has $2{,}716{,}050{,}944$ parameters. We fine-tune with LoRA ($r=16$, $\alpha=32$, dropout $0.05$, target modules \texttt{in\_proj}, \texttt{x\_proj}, \texttt{dt\_proj}), leaving $13{,}451{,}264$ trainable parameters ($\approx 0.50\%$ of the total).

\paragraph{Additional backbones and evaluation models.}
Appendix~\ref{app:loss-invariance} additionally uses the community checkpoint \texttt{ministral/Ministral-3b-instruct}~\citep{Ministral3BInstruct} (not an official Mistral AI release) on UltraFeedback Binarized.
For downstream checks in Tables~\ref{tab:hsteer-benchmarks} and~\ref{tab:ultrafb-alpaca}, AlpacaEval~2 pairwise scoring uses \texttt{Qwen/Qwen2.5-14B-Instruct}~\citep{Qwen2514BInstruct} as a fixed judge model; IFEval uses the official verifiable benchmark without a judge \citep{Zhou2023IFEval}.

\paragraph{Optimization.}
Optimizer choices and key hyperparameters are reported separately for each run, including learning rate, $\beta$, batch size, gradient accumulation, random seed, and gradient clipping. In the TRL baseline (\texttt{DPOTrainer}), we use per-device batch size $4$ with gradient accumulation $4$ (effective batch size $16$). In our code runs, we use dataset-specific batch sizes: UltraFeedback $26$, HelpSteer3-Preference $24$, and HH-RLHF $32$. We use a cosine learning-rate schedule with warmup and optimize with either AdamW or SGD, depending on the experiment (see per-run settings).

\paragraph{Monte Carlo KL estimation.}
We compute the reported KL ourselves as a Monte Carlo estimate of forward KL: for each selected validation prompt, the policy generates multiple responses, and we average $\log \pi_{\theta}(y \mid x)-\log \pi_{\mathrm{ref}}(y \mid x)$ over samples $y \sim \pi_{\theta}(\cdot\mid x)$. In this experiment, KL is estimated on the first 256 validation prompts with 8 samples per prompt and reported both per sequence and per token.

\paragraph{Hardware and software.}
All reported experiments were run in the cloud on NVIDIA A100-SXM4-80GB GPUs with driver 560.35.03. The software environment used Python 3.10.12, PyTorch 2.5.1+cu121 with CUDA 12.1 and cuDNN 90100, TRL 0.18.2, Transformers 4.53.2, Accelerate 1.7.0, and PEFT 0.18.1.

\end{document}